\documentclass[journal,twoside,web]{ieeecolor}
\usepackage{tmi}
\usepackage{cite}
\usepackage{amsmath,amssymb,amsfonts}
\usepackage{algorithmic}
\usepackage{graphicx}
\usepackage{textcomp}

\usepackage{amsmath, amsfonts, graphicx, booktabs, tabularx, multirow, siunitx, soul, bm, tikz, mathtools, subcaption, colortbl, nicematrix, xurl}

\makeatletter
\let\NAT@parse\undefined
\makeatother

\usepackage{hyperref}

\graphicspath{{./figures/}}

\begin{document}

\title{HNOSeg-XS: Extremely Small Hartley Neural Operator for Efficient and Resolution-Robust 3D Image Segmentation}
\author{
Ken C. L. Wong, \IEEEmembership{Senior Member, IEEE}, Hongzhi Wang \IEEEmembership{Senior Member, IEEE}, and Tanveer Syeda-Mahmood, \IEEEmembership{Fellow, IEEE}
\thanks{Ken C. L. Wong,  Hongzhi Wang, and Tanveer Syeda-Mahmood are with the IBM Research -- Almaden Research Center, San Jose, CA, USA (e-mail: \{clwong, hongzhiw, stf\}@us.ibm.com). This is an invited submission from ISBI 2023 best papers. This paper was accepted by IEEE TMI 2025. \copyright 2025 IEEE. Personal use of this material is permitted. Permission from IEEE must be obtained for all other uses, in any current or future media, including reprinting/republishing this material for advertising or promotional purposes, creating new collective works, for resale or redistribution to servers or lists, or reuse of any copyrighted component of this work in other works.}
}

\maketitle

\begin{abstract}
In medical image segmentation, convolutional neural networks (CNNs) and transformers are dominant. For CNNs, given the local receptive fields of convolutional layers, long-range spatial correlations are captured through consecutive convolutions and pooling. However, as the computational cost and memory footprint can be prohibitively large, 3D models can only afford fewer layers than 2D models with reduced receptive fields and abstract levels. For transformers, although long-range correlations can be captured by multi-head attention, its quadratic complexity with respect to input size is computationally demanding. Therefore, either model may require input size reduction to allow more filters and layers for better segmentation. Nevertheless, given their discrete nature, models trained with patch-wise training or image downsampling may produce suboptimal results when applied on higher resolutions. To address this issue, here we propose the resolution-robust HNOSeg-XS architecture. We model image segmentation by learnable partial differential equations through the Fourier neural operator which has the zero-shot super-resolution property. By replacing the Fourier transform by the Hartley transform and reformulating the problem in the frequency domain, we created the HNOSeg-XS model, which is resolution robust, fast, memory efficient, and extremely parameter efficient. When tested on the BraTS'23, KiTS'23, and MVSeg'23 datasets with a Tesla V100 GPU, HNOSeg-XS showed its superior resolution robustness with fewer than 34.7k model parameters. It also achieved the overall best inference time ($<$ 0.24 s) and memory efficiency ($<$ 1.8 GiB) compared to the tested CNN and transformer models\footnote{The code repository is available at \url{https://github.com/IBM/multimodal-3d-image-segmentation}.}.
\end{abstract}

\begin{IEEEkeywords}
Fourier neural operator, Hartley transform, image segmentation, implicit neural representation, super-resolution.
\end{IEEEkeywords}

\section{Introduction}
\label{sec:introduction}

With the introduction of convolutional neural networks (CNNs) and U-Net, the use of encoder-decoder architectures has become dominant in medical image segmentation \cite{Journal:Hesamian:DI2019:deep,Journal:Liu:Sustainability2021:review,Journal:Wang:IET2022:medical}. Combining with GPU computation, deep learning models can provide fast and accurate segmentation that is unachievable by traditional approaches. These models use convolutions with shared weights to capture local features, and use pooling or downsampling to increase the effective receptive field during encoding. For computationally demanding 3D segmentation, however, the receptive fields and abstract levels are more limited than in 2D as fewer layers can be used. As a result, input size reduction is usually required to allow more layers to be used for better accuracy.

\begin{figure}[t]
	\centering
  \begin{minipage}[t]{0.4\linewidth}
    \includegraphics[width=\linewidth]{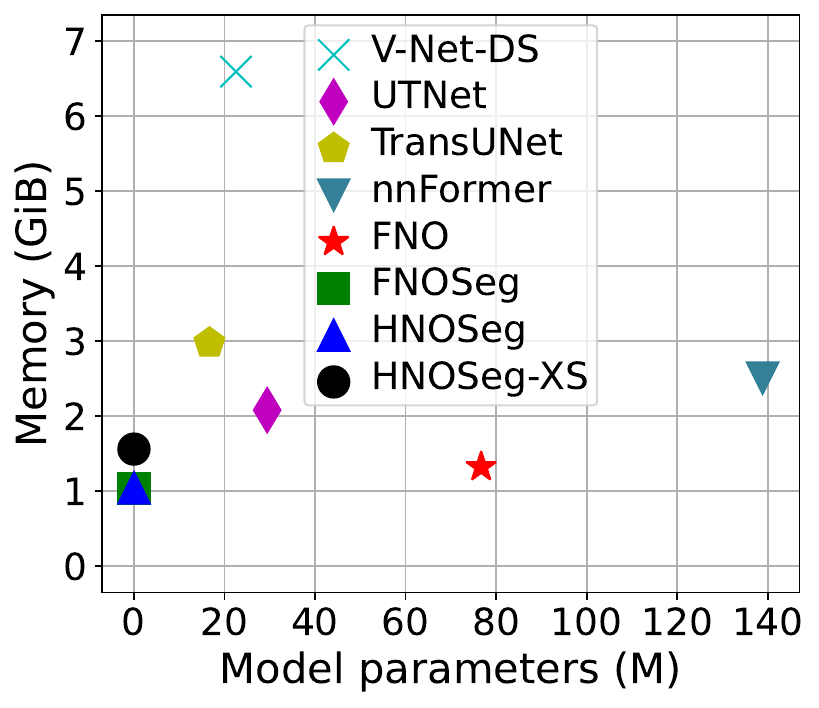}
  \end{minipage}
  \hspace{1em}
  \begin{minipage}[t]{0.42\linewidth}
    \includegraphics[width=\linewidth]{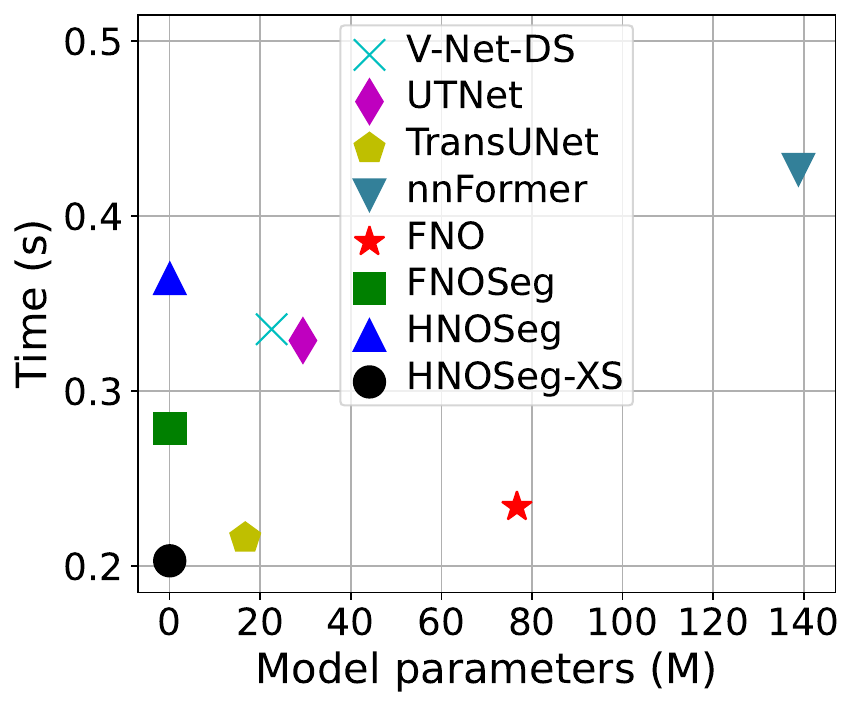}
  \end{minipage}
    \caption{Computational requirements of tested models on a single image, with average values from three datasets.}
    \label{fig:param_time_mem}
\end{figure}

To address the limitation of local receptive fields in CNNs, the multi-head attention (MHA) mechanism of transformers has been introduced to image segmentation \cite{Conference:Vaswani:NIPS2017:attention,Conference:Dosovitskiy:ICLR2021:an,Conference:Liu:ICCV2021:swin,Journal:Xiao:BSPC2023:transformers}. To utilize MHA, the Vision Transformer (ViT) and its alternatives \cite{Conference:Dosovitskiy:ICLR2021:an,Conference:Liu:ICCV2021:swin} divide an image into a sequence of patches, and each patch is treated as a token. Using MHA, nonlinear dependencies among tokens are calculated to capture long-range spatial correlations. For medical image segmentation, some methods use similar patch tokenization approaches \cite{Conference:Hatamizadeh:WACV2022:unetr,Workshop:Cao:ECCV2023:swin,Journal:Zhou:TIP2023:nnformer}, while some use the multi-resolution features from the U-Net as tokens \cite{Conference:Gao:MICCAI2021:utnet,Conference:Xie:MICCAI2021:cotr,Journal:Chen:MIA2024:transunet}. Regardless of its capability, MHA has quadratic computational complexity with respect to the number of tokens which is proportional to the image size. Thus, input size reduction is usually needed for large images. 

To reduce input size, patch-wise training and image downsampling are common approaches. Patch-wise training allows preservation of local details but the effective receptive field can be largely reduced depending on the patch size, and pre- and post-processing are required for patch partitioning and stitching. In contrast, image downsampling reduces image details, but it is more computationally efficient and allows better capturing of long-range dependencies. Therefore, the choice depends on imaging modalities, computational resources, and object structures. In this paper, we focus on the downsampling approach, especially the loss in accuracy when a trained model is applied to different resolutions. The findings are also applicable to each patch in patch-wise training.

To address the issues of using downsampled images, it is desirable to design an architecture that is robust to training resolutions so that the trained model can be applied on higher-resolution images with decent accuracy. Moreover, having a global receptive field in each layer is advantageous, and it is ideal if the model is computationally efficient. To achieve these goals, here we propose HNOSeg-XS, an e\textbf{X}tremely \textbf{S}mall \textbf{H}artley \textbf{N}eural \textbf{O}perator for \textbf{Seg}mentation. This model is a significant improvement of our previously proposed FNOSeg model \cite{Conference:Wong:ISBI2023:fnoseg3d} inspired by the Fourier neural operator (FNO) \cite{Conference:Li:ICLR2021:fourier}. FNO is a deep learning model that learns mappings between functions in partial differential equations (PDEs), with the majority of its model parameters in the Fourier domain. FNO has the appealing properties of zero-shot super-resolution and global receptive field. Our contributions include:
\begin{enumerate}
  \item We replace the Fourier transform by the Hartley transform which produces real numbers in the frequency domain. As the Fourier transform produces complex numbers, operations in the frequency domain are more complicated, less flexible, and require more memory. On the contrary, the Hartley transform allows most deep learning operations to be used in the frequency domain for better accuracy and efficiency.
  \item Using the real-valued advantages of the Hartley transform, we modify the FNO formulations in the frequency domain with consecutive nonlinear operations. As the frequency components required for image segmentation are relatively few, this further reduces the model parameters, computation time, and memory use (Fig. \ref{fig:param_time_mem}).
  \item As the simple structures of neural operators allow flexible adoption of different network architectures, here we adopt the idea of the self-normalizing neural network (SNN) \cite{Conference:Klambauer:NIPS2017:self} to remove normalization layers for even better efficiency. We also use U-Net skip connections with HNOSeg-XS to improve its accuracy.
\end{enumerate}
Experimental results on the BraTS'23 \cite{Journal:Menze:TMI2015:multimodal,Journal:Bakas:SD2017:advancing,Journal:Baid:Arxiv2021:rsna}, KiTS'23 \cite{Journal:Heller:arxiv2023:kits21}, and MVSeg'23 \cite{Conference:Carnahan:MICCAI2021:deepmitral} datasets show that HNOSeg-XS has superior robustness to training image resolution than other tested CNN and transformer models with fewer than 0.2\% of their model parameters. HNOSeg-XS is also memory efficient and has the shortest inference time among the tested models.

\section{Related Work}

\subsection{Medical Image Segmentation}

Deep learning has become the most popular approach for medical image segmentation in recent years because of its superior accuracy and the elimination of manual feature engineering. With the introduction of U-Net \cite{Conference:Ronneberger:MICCAI2015}, the U-shape encoder-decoder architecture and skip connections have become dominant even after the introduction of MHA in transformers. In \cite{Conference:Milletari:3DV2016}, V-Net was proposed for 3D image segmentation with the consideration of memory footprint, and the Dice loss was introduced. In \cite{Conference:Zhou:DLMIA2018:unet++}, UNet++ was introduced as a deeply-supervised encoder-decoder network based on nested and dense skip connections, with the hypothesis that fine-grained details are better captured when high-resolution feature maps from the encoder network are gradually enriched. In \cite{Journal:Isensee:Nature2021:nnu}, the framework nnU-Net was proposed to automatically configure preprocessing, network architecture, training, and post-processing for any new task. 

Given the capability of capturing long-range dependencies, MHA has become popular for medical image segmentation. In UTNet \cite{Conference:Gao:MICCAI2021:utnet}, CNN features at each resolution are sent to a transformer layer, in which the key and value tensors are downsampled in spatial shape for more efficient MHA. In CoTr \cite{Conference:Xie:MICCAI2021:cotr}, multi-scale feature maps from the CNN-encoder are combined to form a sequence of tokens, which are processed by transformer layers before feeding to the CNN-decoder. In TransUNet \cite{Journal:Chen:MIA2024:transunet}, a transformer decoder with masked cross-attention is used to refine localized multi-scale CNN features for 3D segmentation. In UNETR \cite{Conference:Hatamizadeh:WACV2022:unetr}, image patch tokens are processed by the transformer encoder. Similar to U-Net, transformer features at different abstract levels are upsampled to various resolutions by transposed convolutions, which are concatenated in the decoding path. In Swin-Unet \cite{Workshop:Cao:ECCV2023:swin}, the U-Net encoder-decoder architecture is used with Swin Transformer blocks, and feature downsampling and upsampling are achieved by patch merging and expanding layers. In nnFormer \cite{Journal:Zhou:TIP2023:nnformer}, a combination of interleaved convolution and self-attention operations was proposed with the introduction of the local and global volume-based self-attention mechanisms.

\subsection{Image Analysis in Frequency Domain}

Image analysis in frequency domain is an active research area as it allows customized learning strategies on low- and high-frequency components depending on the problems. In \cite{Conference:Yang:CVPR2020:fda}, it was found that replacing the low-frequency components of the source images by those of the target images can reduce the domain gap, and the Fourier domain adaptation framework was proposed for semantic segmentation. In \cite{Conference:Rao:NIPS2021:global}, by replacing MHA with global filter layers based on the Fourier transform, long-range dependencies can be learned with reduced computational cost for image classification. In \cite{Conference:Wang:WACV2024:fremim}, based on the fact that different structural and semantic information lies in low- and high-frequency components, FreMIM was proposed as a masked image model for medical image segmentation with a frequency-domain loss function.

\subsection{Implicit Neural Representations}

Recently, implicit neural representations (INRs) have gained prominence in medical imaging \cite{Workshop:Molaei:ICCV2023:implicit}. INRs represent signals as continuous functions that map the coordinates of a point to the values at that point, with the functions implicitly defined by neural networks. As learning continuous functions provides advantages such as super-resolution and memory/data efficiency, this idea has been adopted for image segmentation. In \cite{Conference:Khan:MICCAI2022:implicit}, a CNN encoder is used to provide local and global features at each selected normalized coordinates, which are then concatenated to provide the segmentation features using a multilayer perceptron. In \cite{Conference:You:MICCAI2023:implicit}, the decoder features from a segmentation backbone are used to provide a coarse segmentation, and the sampling points with high uncertainties in this coarse prediction are selected for adaptive refinements. In \cite{Conference:Stolt:MICCAI2023:nisf}, given the coordinates of points and a subject-specific latent vector, a network is trained to predict the image intensity and segmentation value pairs at those points for image reconstruction and segmentation. At inference, segmentation of an unseen image is achieved by learning a latent vector that minimizes the image reconstruction error.

\subsection{Relation to Our Approach}

Different from CNNs, HNOSeg-XS provides global receptive fields based on the kernel integral operator realized by the Hartley transform. Different from transformers, we do not use patch-wise tokenization which may limit the ability of applying a trained model on different image sizes. Furthermore, we utilize the real-valued characteristic of the Hartley transform in the frequency domain to improve computational efficiency. As we model segmentation by learnable PDEs in continuous space with neural operators, our approach is similar to INRs. Different from other INR-inspired segmentation methods \cite{Workshop:Molaei:ICCV2023:implicit}, we do not use a backbone network for feature extraction. In contrast, the neural operators define model structures that produce image features. Our method does not require specialized sampling methods, training algorithms, or loss functions. We simply propose novel block structures that can be flexibly incorporated into different architectures, with the goals of improving efficiency and resolution robustness.

\begin{figure}[t]
	\footnotesize
    \begin{minipage}[t]{0.19\linewidth}
      \includegraphics[width=\linewidth]{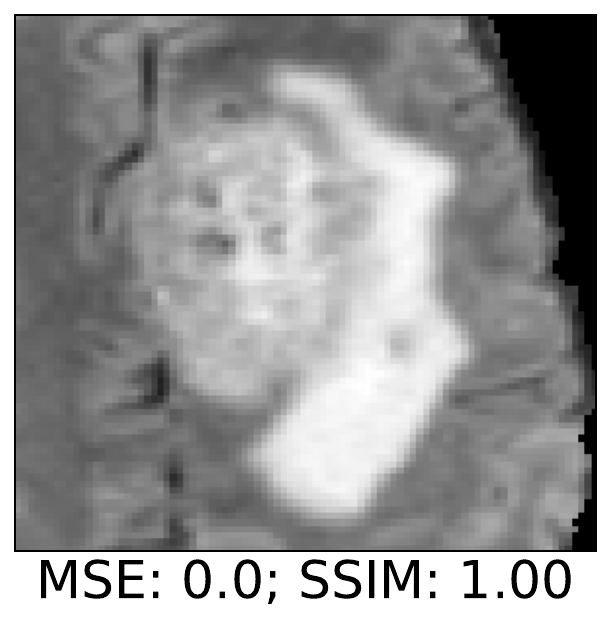}
      \centering{Original}
    \end{minipage}
    \begin{minipage}[t]{0.19\linewidth}
      \includegraphics[width=\linewidth]{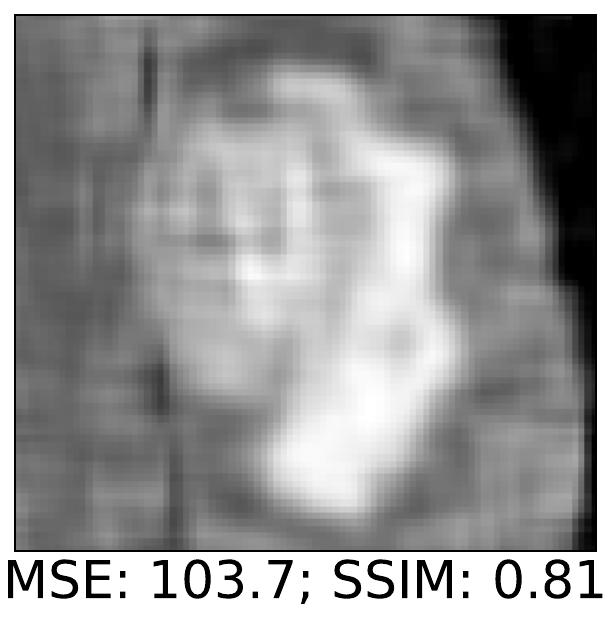}
      \centering{10\%}
    \end{minipage}
    \begin{minipage}[t]{0.19\linewidth}
      \includegraphics[width=\linewidth]{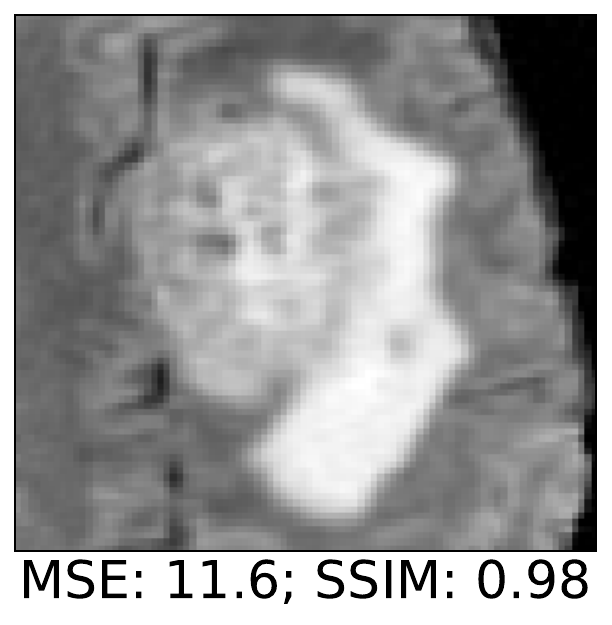}
      \centering{30\%}
    \end{minipage}
    \begin{minipage}[t]{0.19\linewidth}
      \includegraphics[width=\linewidth]{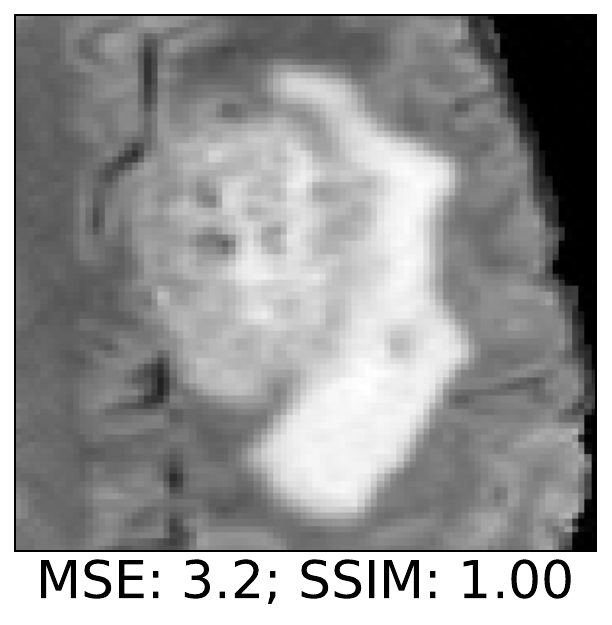}
      \centering{50\%}
    \end{minipage}
    \begin{minipage}[t]{0.19\linewidth}
      \includegraphics[width=\linewidth]{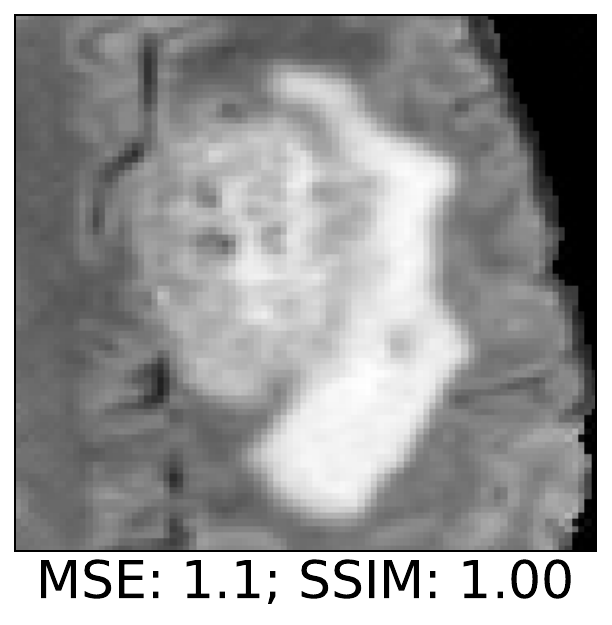}
      \centering{70\%}
    \end{minipage}
    \caption{Images reconstructed with different percentages of frequency modes in each dimension. The total number of modes is half the length of each side. Image intensity $\in$ [0, 255]. MSE: mean squared error; SSIM: structural similarity.}
    \label{fig:modes_illustration}
\end{figure}

\begin{figure*}[t]
  \centering
  \begin{minipage}[t]{1\linewidth}
    \includegraphics[width=0.33\linewidth]{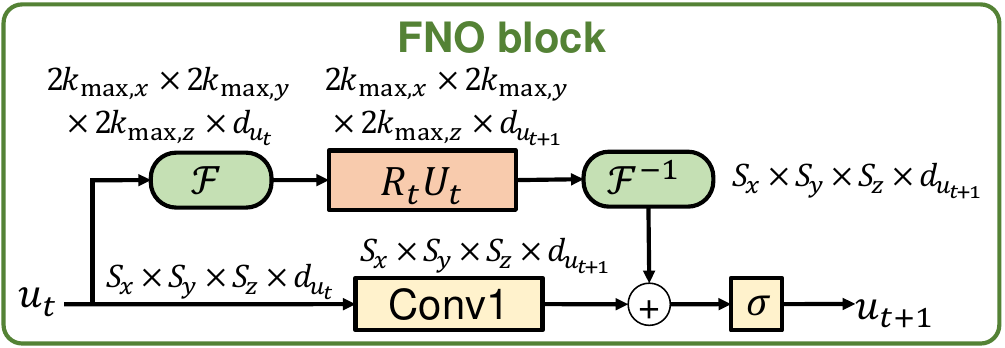}
    \includegraphics[width=0.33\linewidth]{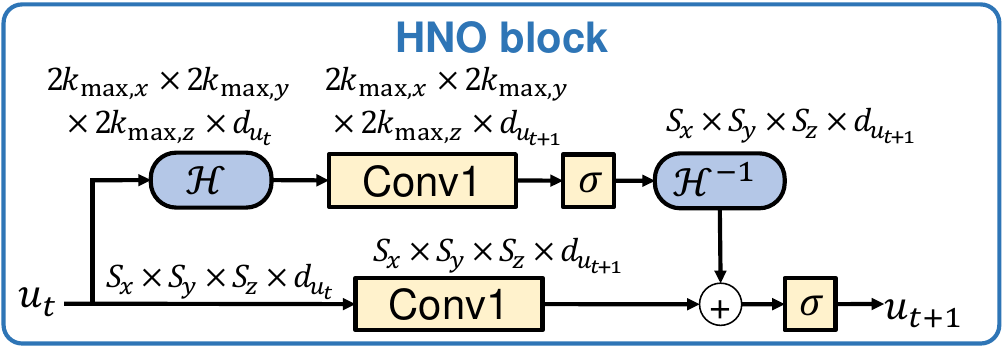}
    \includegraphics[width=0.33\linewidth]{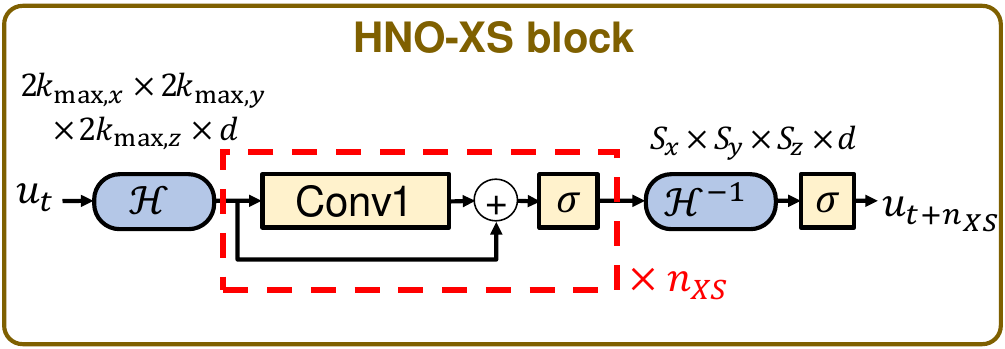}
  \end{minipage}
  \\
  \centering{(a) Different types of neural operator blocks}
  \\
  \medskip
  \begin{minipage}[t]{1\linewidth}
    \includegraphics[width=\linewidth]{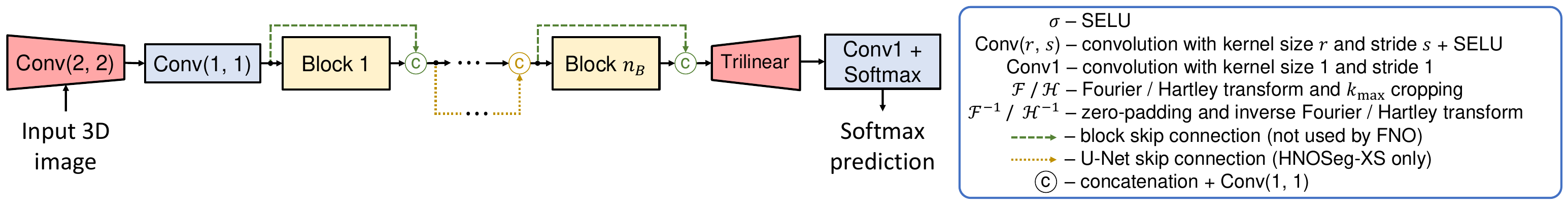}
  \end{minipage}
  \\
  \centering{(b) Network architectures}
  \\
  \caption{Network architectures. (a) The FNO block and HNO block are (\ref{eq:fno_update}) with the kernel integral operator implemented by the Fourier transform and the Hartley transform, respectively. The HNO-XS block is the implementation of (\ref{eq:hno-xs}). (b) The network architectures used by the FNO, FNOSeg, HNOSeg, and HNOSeg-XS models which have different types of neural operator blocks (yellow). $S_x \times S_y \times S_z$ represents the spatial size of a tensor. $k_\mathrm{max} = (k_{\mathrm{max},x}, k_{\mathrm{max},y}, k_{\mathrm{max},z})$ are the numbers of frequency modes, which correspond to a frequency domain of size $2k_{\mathrm{max},x} \times 2k_{\mathrm{max},y} \times 2k_{\mathrm{max},z}$ to cover both positive and negative frequency terms. $n_B$ is the number of blocks and $n_{XS}$ is the number of frequency-domain convolutions in each HNO-XS block. The red blocks are for input and output resampling. We used $d = d_{u_{t}} = d_{u_{t+1}}$ in our experiments.}
  \label{fig:network}
\end{figure*}

\section{Methodology}

In this section, we give a brief description of FNO from which the super-resolution property is inherited, and describe the Hartley transform and frequency-domain weight sharing for model simplification. The reformulation of the frequency-domain equation and the network architectures are also introduced with the loss function and training strategy.

\subsection{Fourier Neural Operator}\label{sec:FNO}

FNO is a deep learning model for learning mappings between functions in PDEs without the PDEs provided \cite{Conference:Li:ICLR2021:fourier}. In computer vision, PDEs have been used to extract features such as gradient, texture, curvature, and other abstract features for image analysis, and have been successfully applied on tasks including image denoising, restoration, and segmentation \cite{Book:Aubert2006:mathematical,Book:Sapiro2006:geometric}. Hence, image segmentation can be modeled by learnable parametric PDEs, and FNO can be used to learn the complicated nonlinear relations that are analytically intractable.

The neural operator is composed of linear integral operators and nonlinear activation functions in continuous space. It is resolution agnostic, which means that the same set of model parameters is shared by different discretization approaches. Therefore, FNO can be trained on lower resolution images and applied on higher resolution images without re-training, i.e., zero-shot super-resolution. For computationally expensive 3D segmentation, such zero-shot super-resolution capability is desirable as a model can be trained on lower-resolution images with reduced computational cost. The neural operator is formulated as iterative updates:
\begin{equation}\label{eq:fno_update}
    \begin{gathered}
      u_{t+1}(x) \coloneqq \sigma \left(W_t u_t(x) + \left(\mathcal{K}_t u_t\right)(x)\right) \\
      \textrm{with} \ \ \left(\mathcal{K}_t u_t\right)(x) \coloneqq \int_{D} \kappa_t(x - y)u_t(y) \,dy, \ \ \forall x \in D
    \end{gathered}
\end{equation}
where $u_t(x) \in \mathbb{R}^{d_{u_t}}$ is a function of $x$. $W_t \in \mathbb{R}^{d_{u_{t+1}} \times d_{u_t}}$ is a learnable linear transformation and $\sigma$ is an activation function. In our work, $D \subset \mathbb{R}^3$ represents the 3D imaging space, and $u_t(x)$ are the outputs of hidden layers with $d_{u_t}$ channels. $\mathcal{K}_t$ is the kernel integral operator with $\kappa_t \in \mathbb{R}^{d_{u_{t+1}} \times d_{u_t}}$ as a learnable kernel function. In fact, $\kappa_t$ can be viewed as a learnable Green's function and $\mathcal{K}_t$ is a solution to a linear differential equation \cite{Journal:Li:arXiv2020:neural}. Using (\ref{eq:fno_update}), the low-level image features $u_t(x)$ can be iteratively transformed to high-level segmentation features. As $\left(\mathcal{K}_t u_t\right)$ is a convolution, it can be efficiently computed by the convolution theorem which states that the Fourier transform ($\mathcal{F}$) of a convolution of two functions is the pointwise product of their Fourier transforms:
\begin{equation}\label{eq:fourier_conv}
    \begin{aligned}
      \left(\mathcal{K}_t u_t\right)(x) &= \mathcal{F}^{-1}\left(\mathcal{F}(\kappa_t) \mathcal{F}(u_t)\right)(x) \\
      &= \mathcal{F}^{-1}\left(R_t U_t\right)(x), \ \ \forall x \in D
    \end{aligned}
\end{equation}
$R_t(k) = (\mathcal{F}\kappa_t)(k) \in \mathbb{C}^{d_{u_{t+1}} \times d_{u_t}}$ is a learnable function in the frequency domain and $U_t(k) = \left(\mathcal{F}u_t\right)(k) \in \mathbb{C}^{d_{u_t}}$. Therefore, each pointwise product at $k$ is realized as a matrix multiplication of complex numbers. When the fast Fourier transform (FFT) is used in implementation, $k \in \mathbb{N}^3$ are non-negative integer coordinates, and each $k$ has a learnable $R_t(k)$. As mainly low-frequency components (modes) are required for image segmentation, only $k_i \leq k_{\mathrm{max},i}$ correspond to the lower frequencies in each dimension $i$ are used to reduce model parameters and computational cost. Fig. \ref{fig:modes_illustration} shows that with only 30\% of frequency modes in each dimension, the original and reconstructed images are already visually indistinguishable, and the structural similarity between them becomes one with 50\% of modes. Therefore, relatively small numbers of frequency modes contain sufficient information for image segmentation. The structural representation of (\ref{eq:fno_update}) with the Fourier transform is shown in Fig. \ref{fig:network}(a) as the FNO block. 

As the kernel integral is over the entire spatial domain, the neural operator in (\ref{eq:fno_update}) has a global receptive field, and this property is ensured in implementation by using FFT in (\ref{eq:fourier_conv}). 

In FNO, as each frequency mode has its own $R_t(k)$, the number of model parameters can be very large. Therefore, we proposed FNOSeg in \cite{Conference:Wong:ISBI2023:fnoseg3d} which uses shared $R_t$ to reduce model parameters by several thousand times. We also modified the architecture to improve the segmentation accuracy. In the following sections, we provide new insights that can further improve segmentation accuracy and efficiency.

\subsection{Hartley Neural Operator (HNO)}

FNO requires complex number operations in the frequency domain, thus the computational requirements such as floating point operations and memory usage are higher than using real number operations. Furthermore, some desirable operations may be undefined for complex numbers and are unavailable in standard libraries. To address these issues, here we use the Hartley transform instead, which is an integral transform alternative to the Fourier transform \cite{Journal:Hartley:IRE1942:more}. The Hartley transform ($\mathcal{H}$) converts real-valued functions to real-valued functions, which is related to the Fourier transform as $(\mathcal{H}f) = \mathrm{Real}(\mathcal{F}f) - \mathrm{Imag}(\mathcal{F}f)$. The convolution theorem of the discrete Hartley transform is more complicated \cite{Journal:Bracewell:JOSA1983:discrete}, and the kernel integration in (\ref{eq:fno_update}) becomes:
\begin{equation}\label{eq:hartley_conv}
    \begin{aligned}
      \mathcal{H}\left(\mathcal{K}_t u_t\right)(k) =& \frac{1}{2} \left[ \hat{R}_t(k) \left(\hat{U}_t(k) + \hat{U}_t(N - k)\right) \right. \\ 
      &+ \left. \hat{R}_t(N - k) \left(\hat{U}_t(k) - \hat{U}_t(N - k)\right) \right]
    \end{aligned}
\end{equation}
with $\hat{R}_t(k) = (\mathcal{H}\kappa_t)(k) \in \mathbb{R}^{d_{u_{t+1}} \times d_{u_t}}$ and $\hat{U}_t(k) = \left(\mathcal{H}u_t\right)(k) \in \mathbb{R}^{d_{u_t}}$. $N \in \mathbb{N}^3$ is the size of the frequency domain. $\hat{R}_t$ and $\hat{U}_t$ are $N$-periodic in each dimension\footnote{In Python, this means $\hat{U}_t[N_x, :, :] = \hat{U}_t[0, : , :]$, etc.}.

In fact, the more complicated formulation in (\ref{eq:hartley_conv}) may offset the benefits of using real numbers. Moreover, similar to (\ref{eq:fourier_conv}), models built using (\ref{eq:hartley_conv}) can have tens of million parameters depending on the numbers of modes ($k_\mathrm{max}$). To improve computational efficiency, inspired by the parameter sharing strategy in convolutional layers, we use the same (shared) $\hat{R}_t$ for all $k$ instead and (\ref{eq:hartley_conv}) becomes:
\begin{equation}\label{eq:hartley_conv_shared}
    \begin{aligned}
      \mathcal{H}\left(\mathcal{K}_t u_t\right)(k) &= \hat{R}_t \hat{U}_t(k) \\
      \textrm{thus} \ \ \ \ \ \left(\mathcal{K}_t u_t\right)(x) &= \mathcal{H}^{-1}\left(\hat{R}_t \hat{U}_t\right)(x)
    \end{aligned}
\end{equation}
which resembles the form in (\ref{eq:fourier_conv}). With shared $\hat{R}_t$, (\ref{eq:hartley_conv_shared}) is equivalent to applying a convolutional layer with the kernel size of one in the frequency domain (HNO block in Fig. \ref{fig:network}(a)). 

Although using (\ref{eq:hartley_conv_shared}) reduces computational complexity and model parameters, the accuracy was unsatisfactory in our initial experiments. In (\ref{eq:fourier_conv}), as the real and imaginary parts of $R_t$ are different, its effect on the magnitudes and phases of $U_t$ is nonlinear and thus more sophisticated features can be learned \cite{Journal:Oppenheim:IEEEProc1981:importance}. With this observation, we introduce nonlinearity to (\ref{eq:hartley_conv_shared}) using a nonlinear activation function ($\sigma$):
\begin{equation}\label{eq:hartley_conv_shared_nonlinear}
      \left(\mathcal{K}_t u_t\right)(x) = \mathcal{H}^{-1}\left(\sigma\left(\hat{R}_t \hat{U}_t\right)\right)(x)
\end{equation}
This makes the HNO models outperform the FNO models as shown in our experiments in Section \ref{sec:experiments}.

\subsection{HNO-XS: Resource-Efficient Extra Small HNO} \label{sec:HNO-XS}

Although HNO with shared parameters is more efficient than FNO, its computational complexity can be further reduced. With our proposed architectures in Section \ref{sec:network}, $W_t$ in (\ref{eq:fno_update}) becomes redundant. Furthermore, as nonlinearity is already included in (\ref{eq:hartley_conv_shared_nonlinear}), $\sigma$ in (\ref{eq:fno_update}) is unnecessary and thus only $\mathcal{K}_t$ remains. In consequence, (\ref{eq:fno_update}) can be rewritten using (\ref{eq:hartley_conv_shared_nonlinear}) as:
\begin{equation}
    \begin{aligned}
     u_{t+1}(x) &= \mathcal{H}^{-1}\left(\sigma\left(\hat{R}_t \mathcal{H}u_t\right)\right)(x) \\
    &= \mathcal{H}^{-1}\left(\sigma\left(\hat{R}_t \mathcal{H}\left(
     \mathcal{H}^{-1}\left(\sigma\left(\hat{R}_{t-1} \mathcal{H}u_{t-1}\right)\right)
     \right)\right)\right)(x) \\
     &= \mathcal{H}^{-1}\left(\sigma\left(\hat{R}_t \sigma\left( \hat{R}_{t-1} \mathcal{H}u_{t-1}\right)\right)\right)(x)
     \end{aligned}
\end{equation}
For $n_{XS}$ frequency-domain updates, we have:
\begin{equation}\label{eq:hno-xs}
    \begin{aligned}
     u_{t + n_{XS}}(x) = \sigma\left(\mathcal{H}^{-1}\left(\sigma\left(\hat{R}_{t + n_{XS} - 1} \ldots \sigma\left(\hat{R}_t \mathcal{H}u_t\right)\right)\right)(x)\right)
     \end{aligned}
\end{equation}
with the final activation ($\sigma$) added for the consistency with FNO and HNO. Eq (\ref{eq:hno-xs}) shows that instead of $n_{XS}$ pairs of forward and inverse Hartley transforms, only one pair is needed. However, $n_{XS}$ cannot be large as linear transformations in the spatial domain are necessary for accurate segmentation (Section \ref{sec:nxs}). We also use residual connections (i.e., replace $ \hat{R}_t$ by $(I + \hat{R}_t)$) to further improve accuracy (Fig. \ref{fig:network}(a)).

Using (\ref{eq:hno-xs}), we can perform multi-layer learning in the frequency domain. This reduces not only the computation time but also the memory requirement as the tensors in the frequency domain are much smaller in size especially in 3D.

\subsection{Network Architectures}
\label{sec:network}

Our proposed network architectures are shown in Fig. \ref{fig:network}(b), which can adopt different types of neural operator blocks in Fig. \ref{fig:network}(a) to produce different models. Different from our work in \cite{Conference:Wong:ISBI2023:fnoseg3d}, here we adopt the idea of the self-normalizing neural network \cite{Conference:Klambauer:NIPS2017:self}. In SNN, when using the scaled exponential linear unit (SELU) with the weights of each layer initialized from a Gaussian distribution of zero mean and  (1/fan-in) variance, the network becomes self-normalizing. Using also block skip connections with concatenation and convolution (green arrows in Fig. \ref{fig:network}(b)), normalization layers become unnecessary. To further reduce computation, we use a convolutional layer with the kernel size and stride of two to downsample the input, and apply trilinear upsampling before the output layer (red blocks in Fig. \ref{fig:network}(b)). Moreover, as the neural operator blocks have intrinsic global receptive fields (Section \ref{sec:FNO}), pooling is not required. The hyperparameters $n_{XS}$ and $k_\mathrm{max}$ can be obtained empirically (Section \ref{sec:nxs} and \ref{sec:modes}).

Different neural operator blocks in Fig. \ref{fig:network}(a) can be used as the yellow blocks in Fig. \ref{fig:network}(b) to produce different models. We name the models as FNOSeg, HNOSeg, and HNOSeg-XS with the FNO block (shared parameters), HNO block, and HNO-XS block, respectively. The FNO model \cite{Conference:Li:ICLR2021:fourier} can also be created using the FNO block with frequency-dependent parameters and the absence of block skip connections.

For HNOSeg-XS, the HNO-XS block resembles the convolution block in U-Net with multiple ($n_{XS}$) convolutions. As numerous models, including transformer models, achieve state-of-the-art results using the U-Net style \cite{Workshop:Cao:ECCV2023:swin,Journal:Zhou:TIP2023:nnformer}, we adopt U-Net skip connections in HNOSeg-XS. We simply treat the first and second halves of the blocks as encoding and decoding blocks, respectively, and connect block $n_i$ with block $n_{B-i+1}$. This idea is less suitable for the other neural operator models as they require more blocks ($n_B$) and thus memory.

\subsection{Loss Function with Pearson's Correlation Coefficient (PCC)}

The PCC loss ($L_{PCC} \in [0, 1]$) is used as it is robust to learning rate and accurate for image segmentation \cite{Workshop:Wong:MLMI2022:3d}. It also consistently outperformed the Dice loss and weighted cross-entropy in our experiments. $L_{PCC}$ is computed as:
\begin{gather}
\label{eq:pcc_loss_1}
L_{PCC} = \mathbf{E}[1 - PCC_l], \\
\label{eq:pcc_loss_2}
PCC_l = 0.5 \left(\tfrac{\sum_{i=1}^{n_v}(p_{li} - \bar{p}_l)(y_{li} - \bar{y}_l)}{\sqrt{\left(\sum_{i=1}^{n_v}(p_{li} - \bar{p}_l)^2\right)\left(\sum_{i=1}^{n_v}(y_{li} - \bar{y}_l)^2\right)  + \epsilon}} + 1 \right)
\end{gather}
with $\mathbf{E}[\bullet]$ the mean value across semantic labels $l$. $p_{li} \in [0, 1]$ are the prediction scores, $y_{li} \in \{0, 1\}$ are the ground-truth values, and $n_v$ is the number of voxels in an image. $\epsilon$ is a small positive number (e.g., $10^{-7}$) to avoid divide-by-zero situations, for example, when label $l$ is missing in an image and all $y_{li} = 0$. $L_{PCC}$ = 0, 0.5, and 1 represent perfect prediction, random prediction, and total disagreement, respectively. As the mean values ($\bar{p}_l$ and $\bar{y}_l$) are subtracted from the samples in (\ref{eq:pcc_loss_2}), both foreground and background voxels of each label contribute to $L_{PCC}$, and a low $L_{PCC}$ is achievable only if both foreground and background are well classified.

\subsection{Training Strategy}

For multimodal imaging datasets, the images of different modalities are stacked along the channel axis to provide a multi-channel input. As the intensity ranges across modalities can be quite different, intensity normalization is performed on each image of each modality. Image augmentation with rotation (axial, $\pm$\ang{30}), shifting ($\pm$20\%), and scaling ([0.8, 1.2]) is used and each image has an 80\% chance to be transformed. The Adamax optimizer \cite{Journal:Kingma:arXiv2014} is used with the cosine annealing learning rate scheduler \cite{Conference:Loshchilov:ICLR2017:SGDR}.

\section{Experiments}
\label{sec:experiments}

To study the characteristics of the proposed models, we performed experiments on three open datasets of 3D images, covering three imaging modalities and three organ types. We studied the number of frequency-domain convolutions of HNO-XS ($n_{XS}$), and the effect of the numbers of modes ($k_\mathrm{max}$) on different neural operator models and datasets. We also compared HNOSeg-XS with other models in terms of resolution robustness and computational requirements. 

\subsection{Data and Experimental Setups}

\subsubsection{Multimodal Brain Tumor Segmentation Challenge 2023}

The BraTS'23 adult glioma dataset has 1251 cases of gliomas, each has four modalities of T1, post-contrast T1, T2, and T2-FLAIR images with 240$\times$240$\times$155 voxels \cite{Journal:Baid:Arxiv2021:rsna,Misc:BraTS23}. We split the data into 60\% for training, 10\% for validation, and 30\% for testing. Regardless of the training image sizes, the image size of 240$\times$240$\times$155 was used in testing. The Dice coefficients and 95\% Hausdorff distances (HD95) in the ``whole tumor'' (WT), ``tumor core'' (TC), and ``enhancing tumor'' (ET) regions are reported.

\subsubsection{The 2023 Kidney and Kidney Tumor Segmentation Challenge}

The KiTS'23 dataset has 489 contrast-enhanced computed tomography (CT) scans of renal tumors and cysts \cite{Journal:Heller:arxiv2023:kits21,Misc:KiTS23}. The image size in the transverse plane is 512$\times$512, while the length along the longitudinal axis varies from 29 to 1059 pixels with the average around 195. We resized all images to 256$\times$256$\times$128 for the experiments. We split the data into 70\% for training, 10\% for validation, and 20\% for testing. Regardless of the training image sizes, the image size of 256$\times$256$\times$128 was used in testing. Complying with the challenge, the Dice coefficients and surface Dice coefficients in the ``kidney and mass'' (KM), ``mass'' (M), and ``tumor'' (T) regions are reported.

\subsubsection{Segmentation of the Mitral Valve from 3D Transesophageal Echocardiography}
The MVSeg'23 dataset consists of 3D transesophageal echocardiography volumes, with 105, 30, and 40 images for training, validation, and testing, respectively \cite{Conference:Carnahan:MICCAI2021:deepmitral,Misc:MVSeg23}. All images have 208 pixels along the longitudinal axis, but their lengths vary from 144 to 336 pixels along the frontal axis, and from 112 to 256 pixels along the sagittal axis. We resized all images to 224$\times$160$\times$208 for the experiments. Regardless of the training image sizes, the image size of 224$\times$160$\times$208 was used in testing. The Dice coefficients and 95\% Hausdorff distances in the ``posterior leaflet'' (PL) and ``anterior leaflet'' (AL) regions are reported.

\subsection{Tested Models}

We compared HNOSeg-XS with the following models:
\begin{enumerate}
  \item \textbf{V-Net-DS} \cite{Conference:Wong:MICCAI2018}: a V-Net with deep supervision representing the commonly-used encoder-decoder architectures.
  \item \textbf{UTNet} \cite{Conference:Gao:MICCAI2021:utnet}: a U-Net enhanced by MHA with improved efficiency through spatial downsampling of the key and value tensors.
  \item \textbf{TransUNet} \cite{Journal:Chen:MIA2024:transunet}: a transformer decoder with masked cross-attention is used to refine localized multi-scale CNN features for 3D segmentation.
  \item \textbf{nnFormer} \cite{Journal:Zhou:TIP2023:nnformer}: it uses interleaved convolution and self-attention with local and global volume-based self-attention mechanisms for 3D segmentation.
  \item \textbf{FNO} \cite{Conference:Li:ICLR2021:fourier}: the original FNO without shared parameters and skip connections.
  \item \textbf{FNOSeg} \cite{Conference:Wong:ISBI2023:fnoseg3d}: FNO with shared parameters and skip connections.
  \item \textbf{HNOSeg}: FNOSeg with the Fourier transform replaced by the Hartley transform, and an additional nonlinear activation function in the frequency domain.
\end{enumerate}
The architectures in Fig. \ref{fig:network}(b) were used by the neural operator models. The resampling approach in Section \ref{sec:network} (red blocks in Fig. \ref{fig:network}(b)) was applied to all models, except nnFormer which has its own resampling approach. For hyperparameters, FNO, FNOSeg, and HNOSeg had $d=24$ and $n_B=24$. HNOSeg-XS had $d=24$ and $n_{XS} \times n_{B}$ = 24, where $n_{XS}$ is dataset dependent (Section \ref{sec:nxs}). For V-Net-DS, the hyperparameters in \cite{Conference:Wong:MICCAI2018} were used but the number of filters was doubled. UTNet shared the common hyperparameters with V-Net-DS, while the transformer-related hyperparameters were empirically obtained for decent results. For TransUNet and nnFormer, the hyperparameters from the corresponding publications were used \cite{Journal:Chen:MIA2024:transunet,Journal:Zhou:TIP2023:nnformer}. The numbers of epochs for BraTS'23, KiT'23, and MVSeg'23 were 300, 400, and 400, respectively. Except TransUNet and nnFormer, the maximum and minimum learning rates were $5\times10^{-3}$ and $10^{-3}$, respectively. For TransUNet and nnFormer which involve embeddings learning, smaller maximum and minimum learning rates of $10^{-3}$ and $10^{-4}$ were used. An NVIDIA Tesla V100 GPU with 32 GB memory was used with a batch size of one, and PyTorch 2.5.1 was used for implementation. Note that as our goal is to compare the architectural difference but not the training algorithm, nnU-Net \cite{Journal:Isensee:Nature2021:nnu} was not included as it is a self-configuring framework. Similarly, the resolution-robust models in \cite{Conference:Khan:MICCAI2022:implicit} and \cite{Conference:Stolt:MICCAI2023:nisf} were not included as they require specialized training and inference algorithms.

\begin{figure}[t]
  \centering
  \begin{minipage}[t]{0.8\linewidth}
  \includegraphics[width=\linewidth]{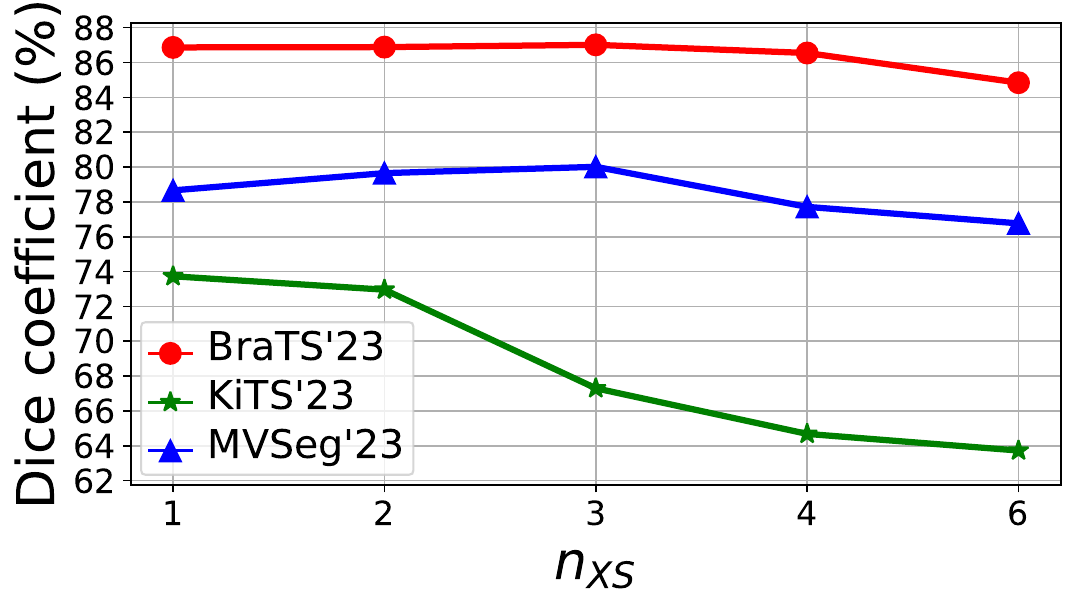}
  \end{minipage}
  \caption{Segmentation accuracy versus the number of frequency-domain convolutions in the HNO-XS block ($n_{XS}$ in (\ref{eq:hno-xs})). Each point represents the average value of different regions of the testing data. The number of blocks $n_B$ (Fig. \ref{fig:network}(b)) changed with $n_{XS}$ to keep $n_{XS} \times n_B = 24$. BraTS'23: training 120$\times$120$\times$78, testing 240$\times$240$\times$155. KiTS'23: training 128$\times$128$\times$64, testing 256$\times$256$\times$128. MVSeg'23: training 112$\times$80$\times$104, testing 224$\times$160$\times$208.}
  \label{fig:vs_block}
\end{figure}

\begin{figure*}[t]
    \centering
    \begin{minipage}[t]{0.3\linewidth}
    \includegraphics[width=\linewidth]{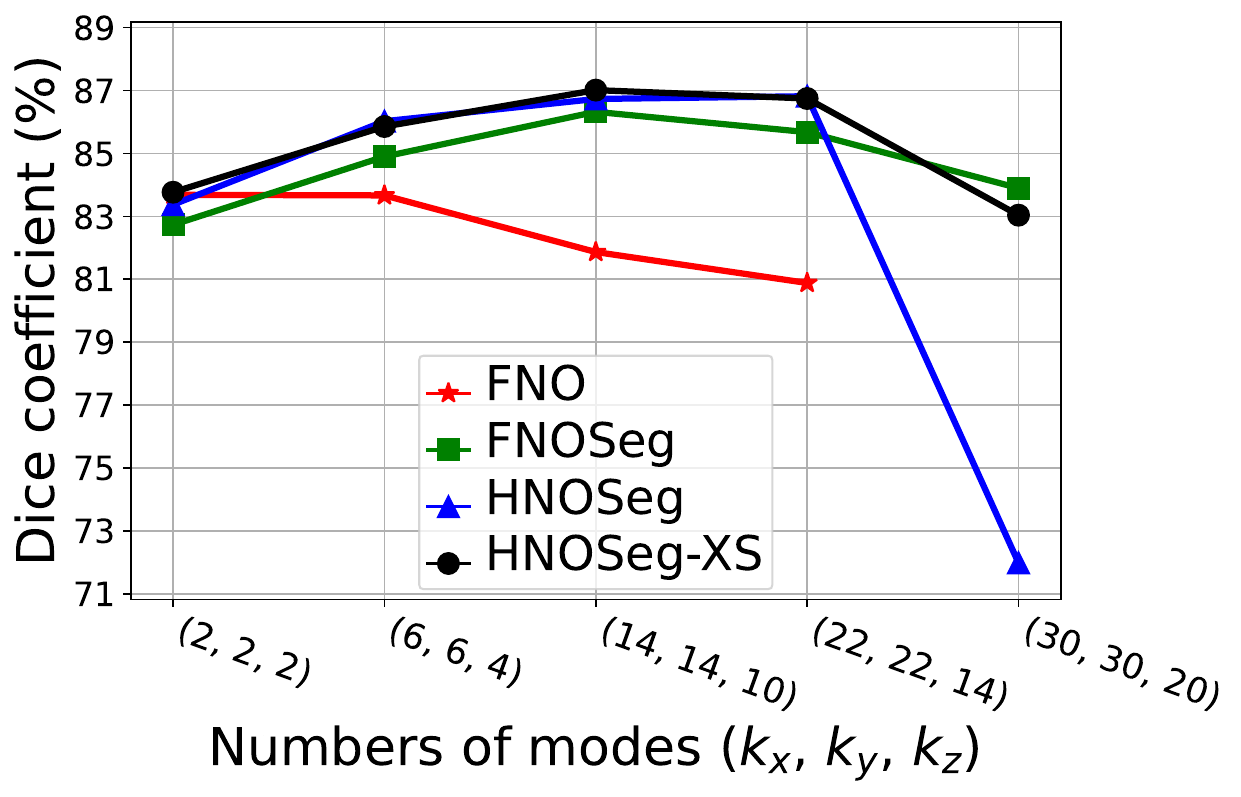}
    \centering{(a) BraTS'23}
    \end{minipage}
    \begin{minipage}[t]{0.3\linewidth}
    \includegraphics[width=\linewidth]{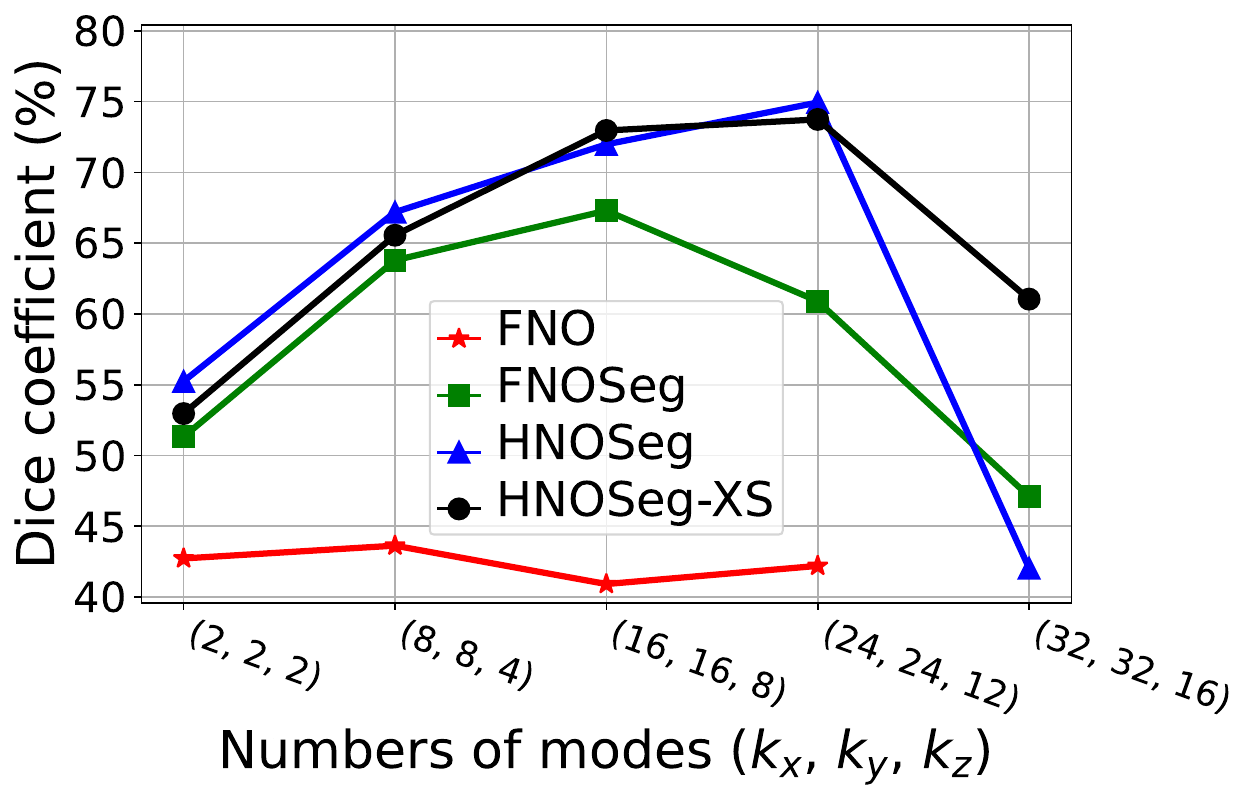}
    \centering{(b) KiTS'23}
    \end{minipage}
    \begin{minipage}[t]{0.3\linewidth}
    \includegraphics[width=\linewidth]{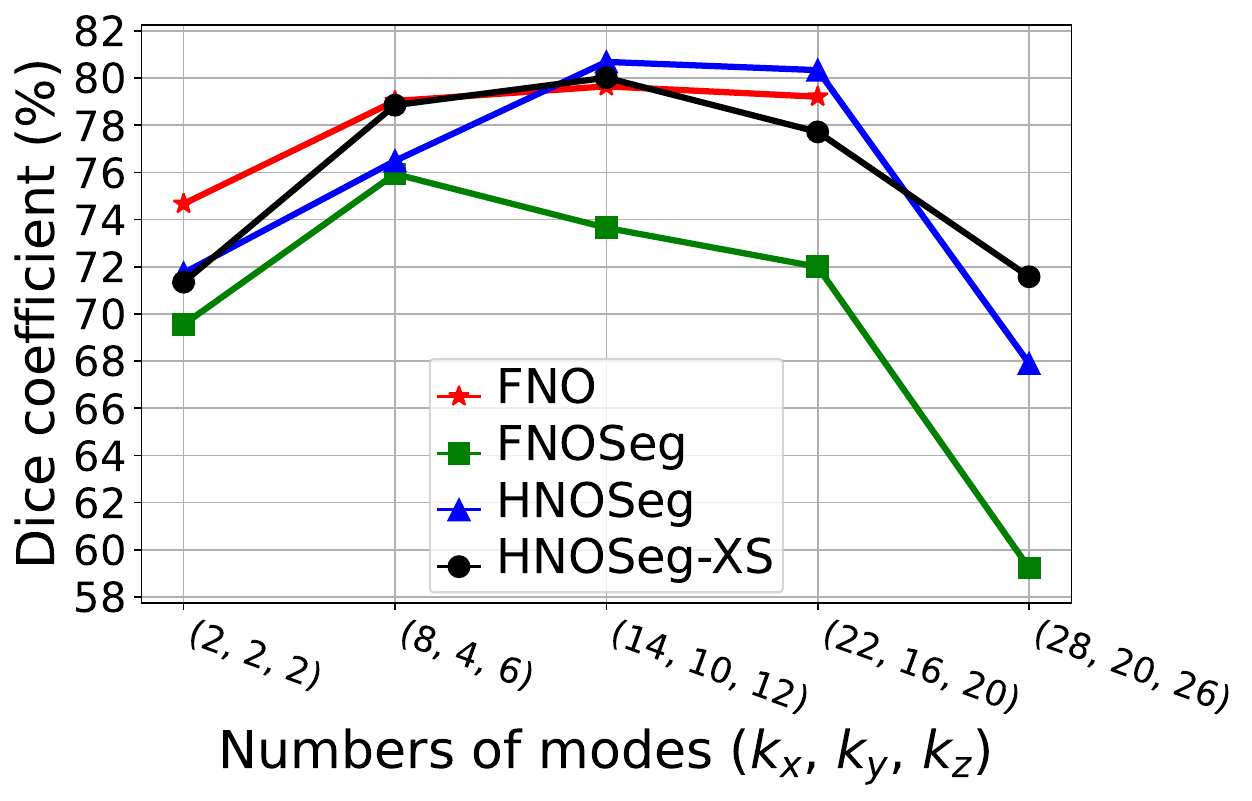}
    \centering{(c) MVSeg'23}
    \end{minipage}
    \caption{The effect of the numbers of modes ($k_\mathrm{max}$) on segmentation accuracy. Each point represents the average value of different regions of the testing data. BraTS'23: training 120$\times$120$\times$78, testing 240$\times$240$\times$155. KiTS'23: training 128$\times$128$\times$64, testing 256$\times$256$\times$128. MVSeg'23: training 112$\times$80$\times$104, testing 224$\times$160$\times$208.}
    \label{fig:vs_modes}
\end{figure*}

\begin{figure*}[t]
  \centering
  \begin{minipage}[t]{0.3\linewidth}
  \includegraphics[width=\linewidth]{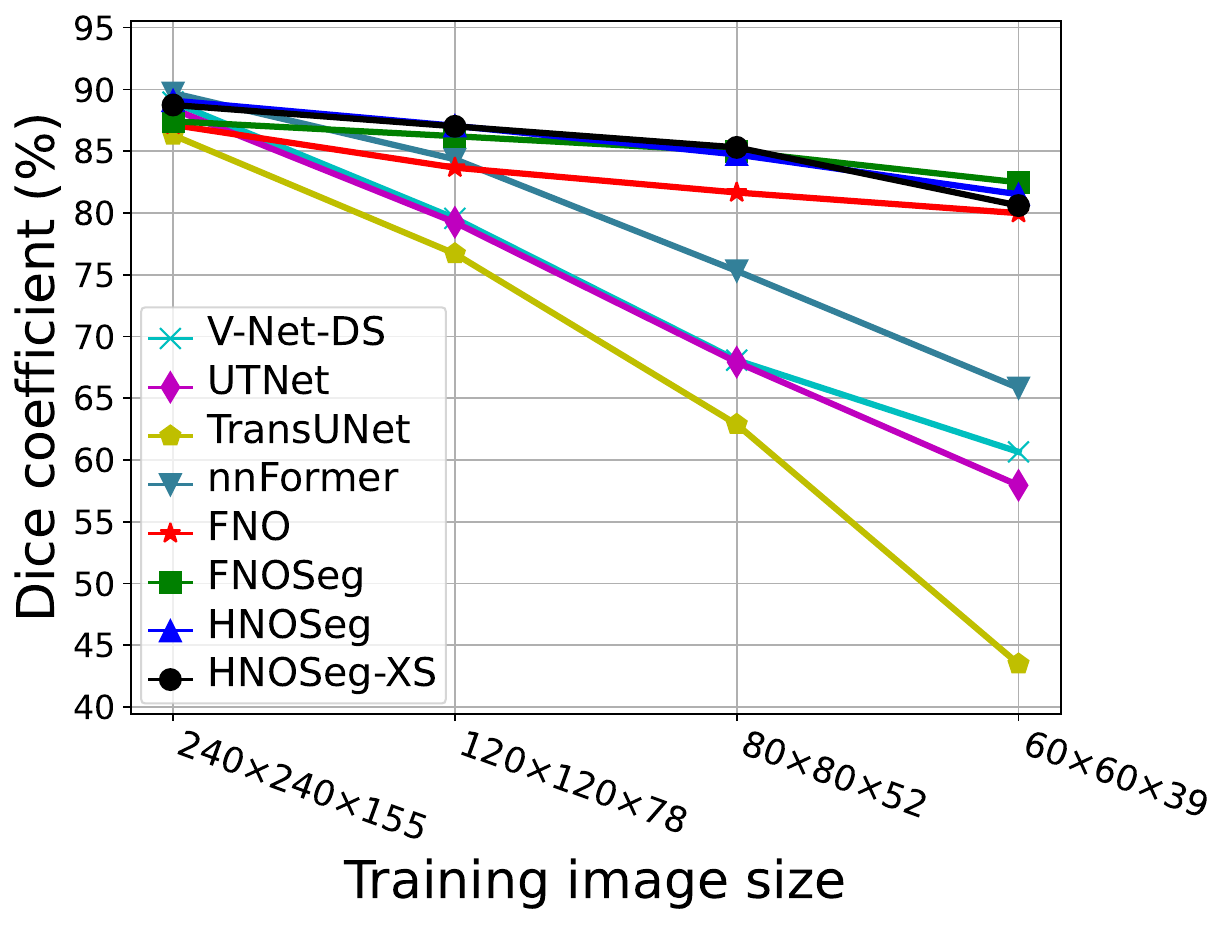}
  \\
  \includegraphics[width=\linewidth]{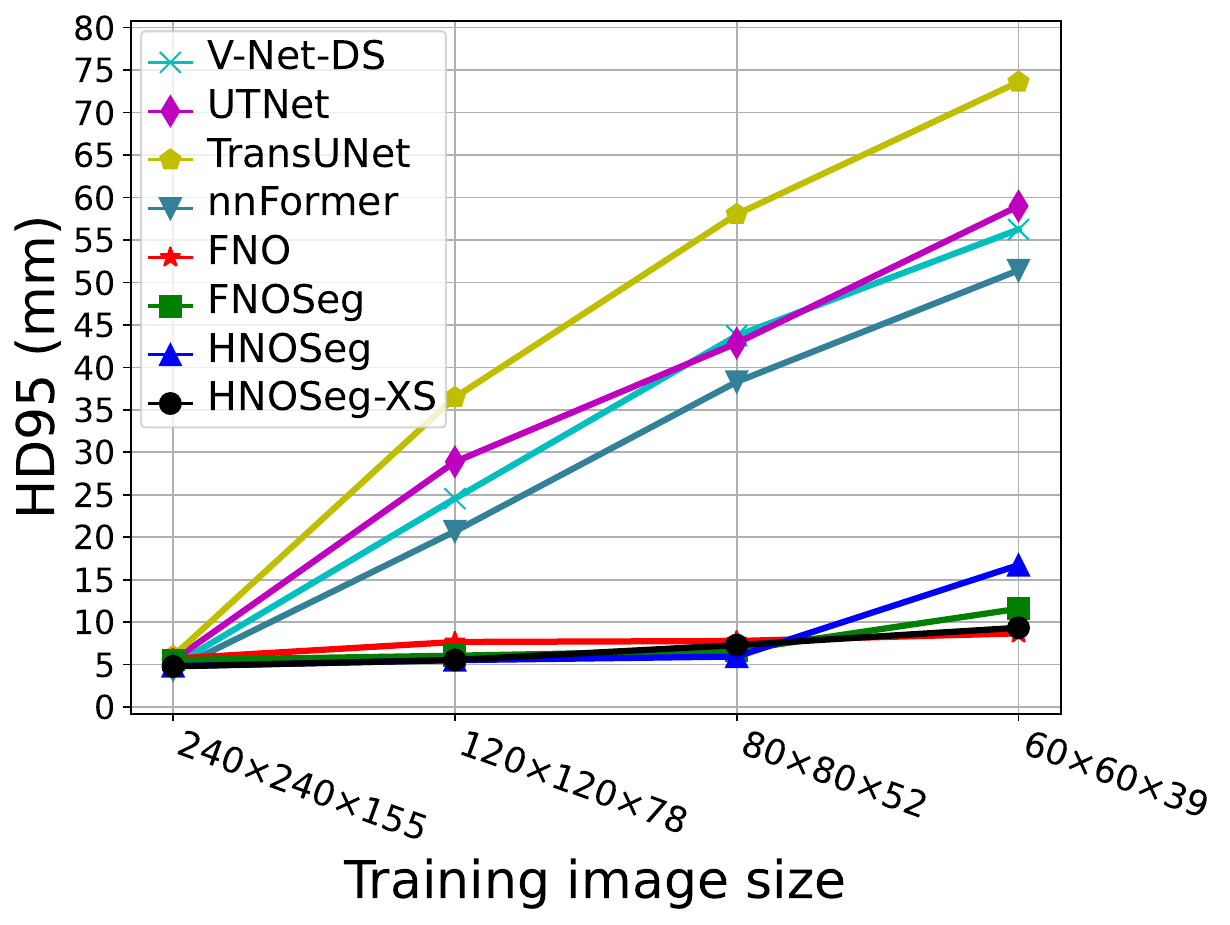}
  \\
  \centering{(a) BraTS'23}
  \end{minipage}
  \vrule
  \hspace{1em}
  \begin{minipage}[t]{0.3\linewidth}
  \includegraphics[width=\linewidth]{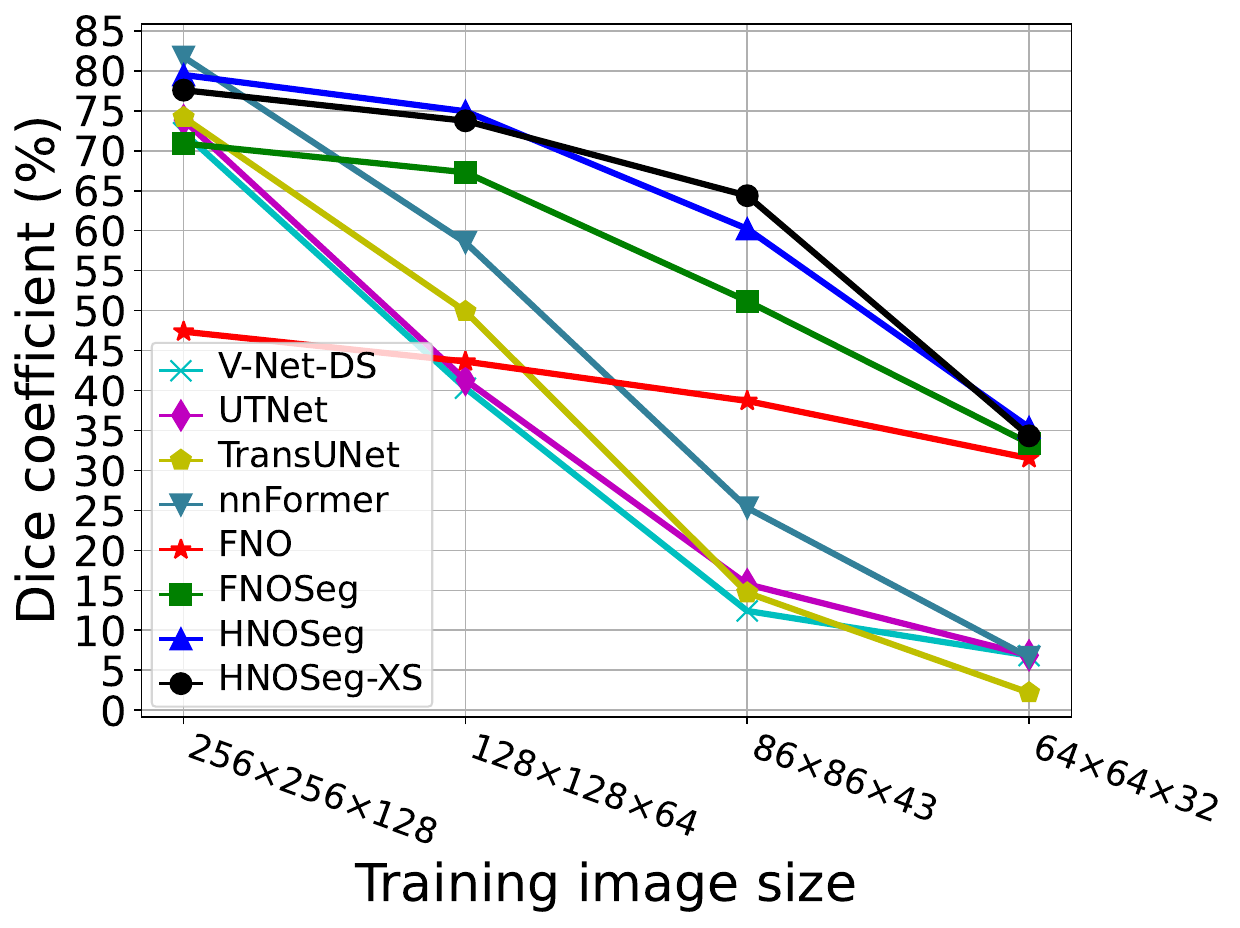}
  \\
  \includegraphics[width=\linewidth]{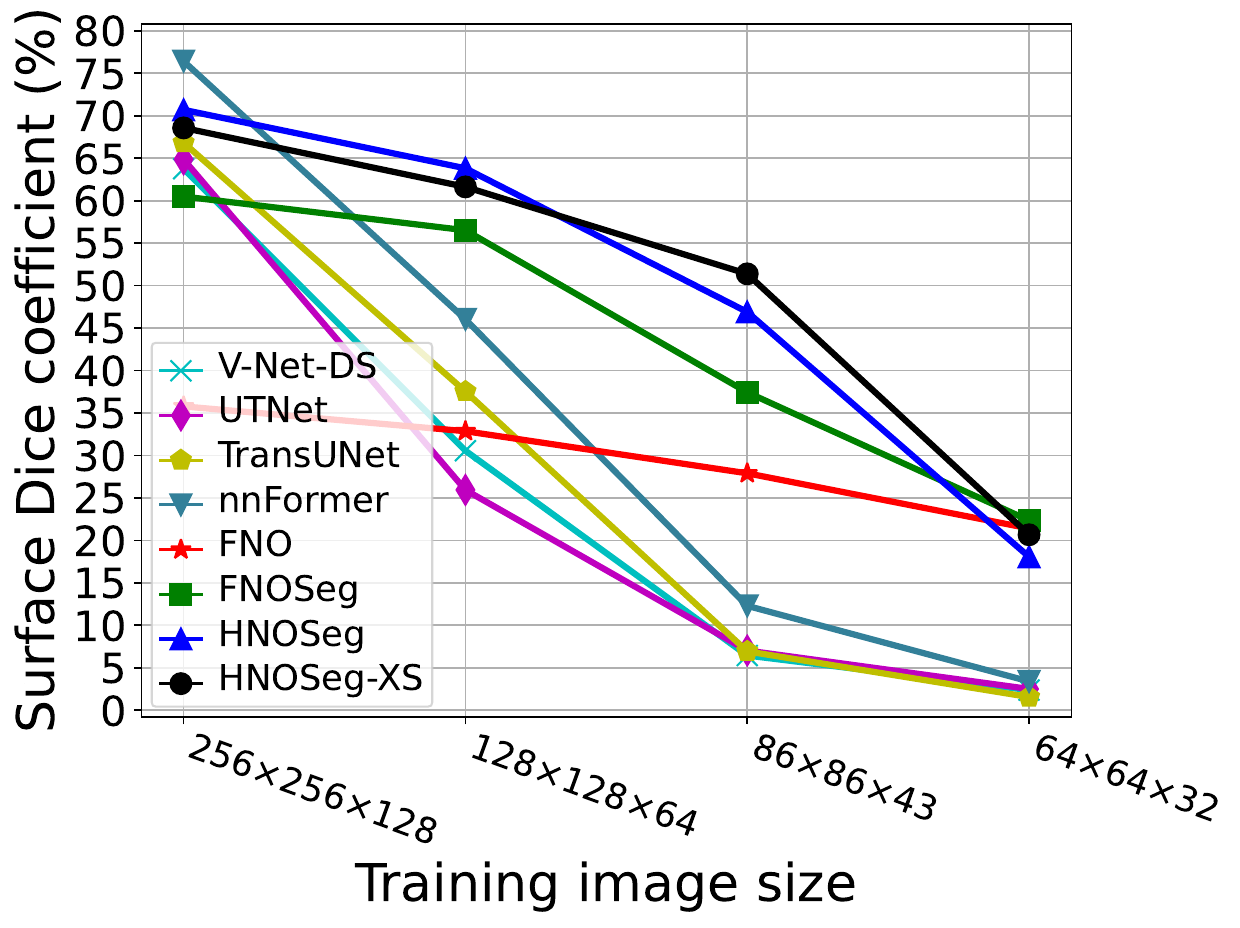}
  \\
  \centering{(b) KiTS'23}
  \end{minipage}
  \vrule
  \hspace{1em}
  \begin{minipage}[t]{0.3\linewidth}
  \includegraphics[width=\linewidth]{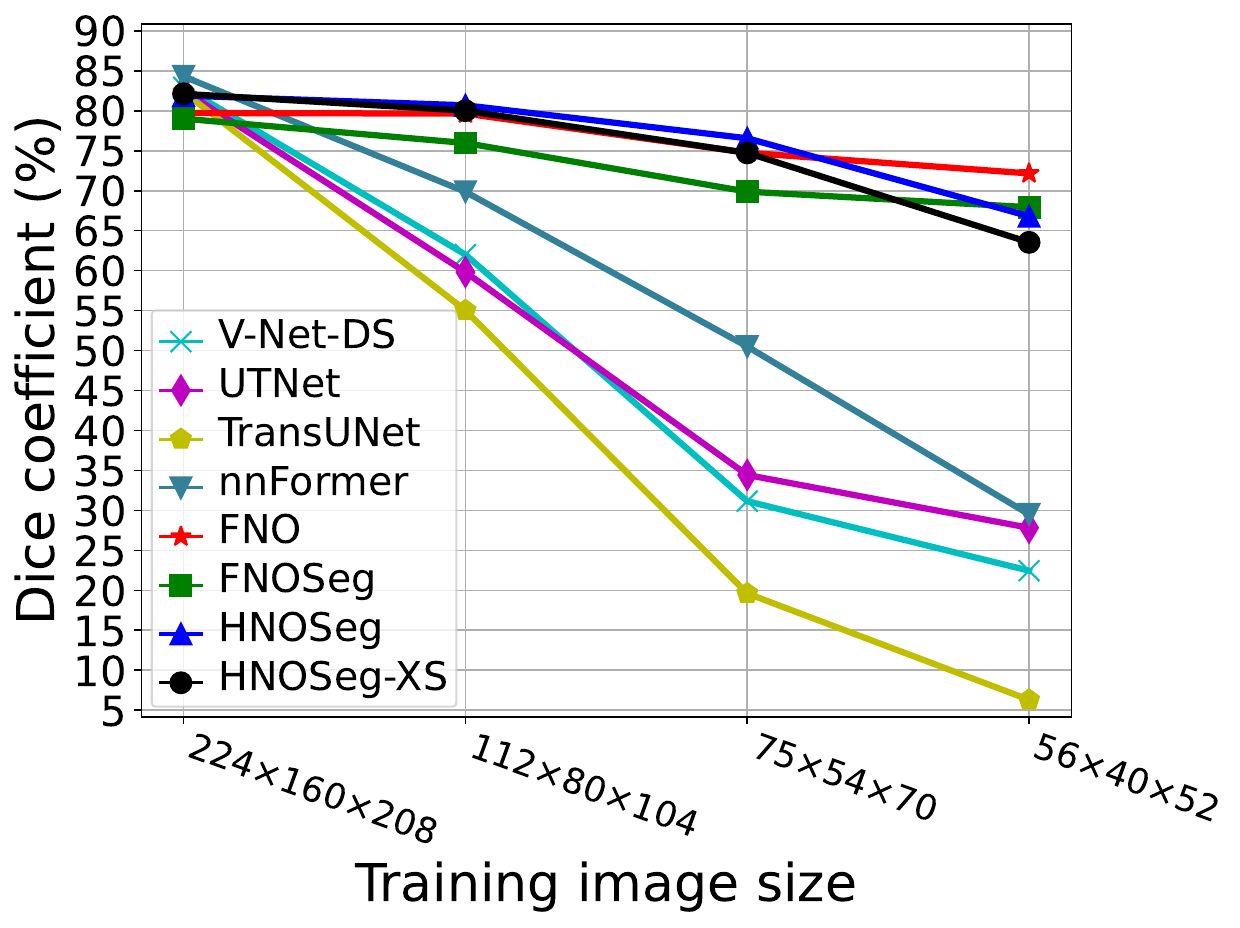}
  \\
  \includegraphics[width=\linewidth]{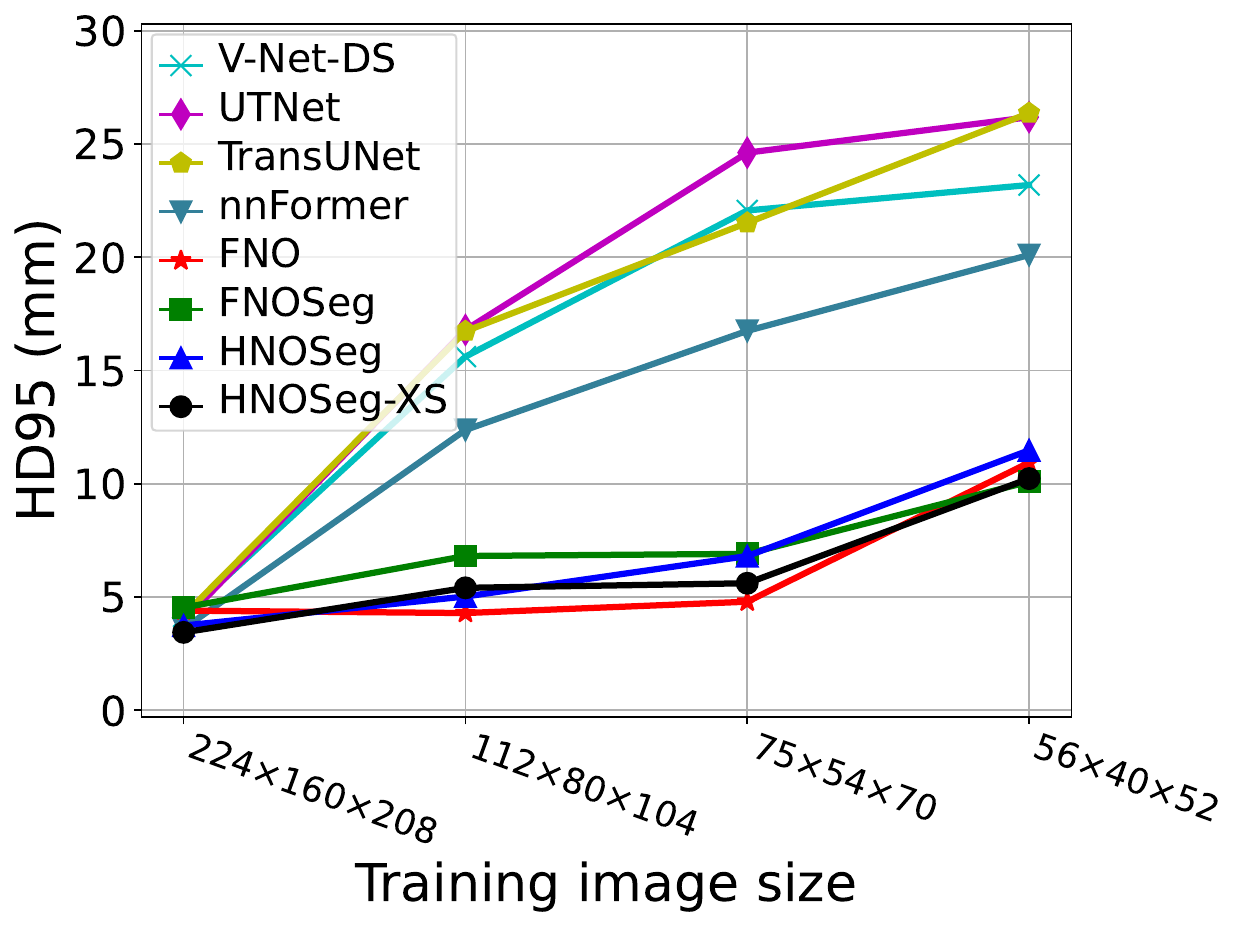}
  \\
  \centering{(c) MVSeg'23}
  \end{minipage}
  \caption{Comparisons of robustness to training image resolutions. Each point represents the average value of different regions of the testing data. Regardless of the training image sizes, the largest image sizes were used in testing (BraTS'23: 240$\times$240$\times$155; KiTS'23: 256$\times$256$\times$128; MVSeg'23: 224$\times$160$\times$208).}
  \label{fig:vs_imgsize}
\end{figure*}

\begin{table*}[t]
\caption{Numerical comparisons of models trained with different image resolutions. Regardless of the training image sizes, the largest image sizes were used in testing (BraTS'23: 240$\times$240$\times$155; KiTS'23: 256$\times$256$\times$128; MVSeg'23: 224$\times$160$\times$208). In BraTS'23, WT, TC, and ET represent ``whole tumor'', ``tumor core'', and ``enhancing tumor'', respectively. In KiTS'23, KM, M, and T represent ``kidney and mass'', ``mass'', and ``tumor', respectively. In MVSeg'23, PL and AL represent ``posterior leaflet'' and ``anterior leaflet'', respectively. The best two values in each column are highlighted.}
\label{table:results}

\scriptsize
\centering

\definecolor{Gray}{gray}{0.85}

\newcolumntype{G}{>{\raggedleft\arraybackslash\columncolor{Gray}}X@{\extracolsep{10pt}}}
\newcolumntype{W}{>{\raggedleft\arraybackslash}X@{\extracolsep{10pt}}}
\newcolumntype{L}{l@{\extracolsep{10pt}}}
\newcommand{\boldblue}[1]{\textcolor{blue}{\textbf{#1}}}
\newcommand{\myrule}{\specialrule{0.5pt}{0pt}{0pt}}

\begin{tabularx}{\linewidth}{LGGGWWWGGGWWWGGGWWW}
\toprule
\textbf{BraTS'23} & \multicolumn{6}{c}{240$\times$240$\times$155 (training)} & \multicolumn{6}{c}{120$\times$120$\times$78 (training)} & \multicolumn{6}{c}{80$\times$80$\times$52 (training)} \\
\cline{1-1} \cline{2-7} \cline{8-13} \cline{14-19} \noalign{\smallskip}
Metric & \multicolumn{3}{c}{Dice (\%)} & \multicolumn{3}{c}{HD95 (mm)} & \multicolumn{3}{c}{Dice (\%)} & \multicolumn{3}{c}{HD95 (mm)} & \multicolumn{3}{c}{Dice (\%)} & \multicolumn{3}{c}{HD95 (mm)} \\
\cline{1-1} \cline{2-4} \cline{5-7} \cline{8-10} \cline{11-13} \cline{14-16} \cline{17-19} \noalign{\smallskip}
Region & WT & TC & ET & WT & TC & ET & WT & TC & ET & WT & TC & ET & WT & TC & ET & WT & TC & ET \\
\myrule
V-Net-DS
& \boldblue{92.4} & 89.0 & \boldblue{85.5} & 5.9 & 4.7 & 3.9
& 82.4 & 79.3 & 76.9 & 28.5 & 25.1 & 20.0
& 72.6 & 63.3 & 68.3 & 46.9 & 49.8 & 34.5
\\
UTNet
& 92.1 & 88.3 & 84.8 & 6.3 & 5.7 & 4.5
& 82.8 & 79.2 & 75.7 & 33.2 & 28.9 & 24.5
& 73.2 & 62.2 & 68.3 & 45.4 & 47.0 & 36.2
\\
TransUNet
& 89.3 & 87.2 & 82.4 & 8.1 & 5.3 & 4.5
& 77.6 & 77.5 & 75.0 & 43.2 & 37.7 & 28.6
& 62.5 & 61.7 & 64.4 & 59.2 & 61.5 & 53.4
\\
nnFormer
& \boldblue{92.4} & \boldblue{90.1} & \boldblue{86.6} & \boldblue{5.8} & \boldblue{3.9} & \boldblue{3.4}
& 87.2 & 84.5 & 81.4 & 26.4 & 21.2 & 14.5
& 76.9 & 74.1 & 74.9 & 46.2 & 40.1 & 28.5
\\
FNO
& 91.2 & 87.3 & 82.9 & 6.8 & 5.6 & 4.7
& 88.8 & 83.6 & 78.6 & 8.2 & 8.0 & 6.8
& 87.6 & 82.2 & 75.2 & 8.3 & 8.1 & 6.9
\\
FNOSeg
& 91.6 & 87.3 & 83.4 & 6.5 & 5.5 & 4.6
& 90.4 & 86.9 & 81.4 & 7.5 & \boldblue{5.7} & 4.9
& \boldblue{89.1} & \boldblue{85.8} & \boldblue{80.0} & \boldblue{7.7} & 6.8 & \boldblue{5.5}
\\
HNOSeg
& 92.2 & \boldblue{89.9} & 85.2 & 6.4 & \boldblue{4.4} & \boldblue{3.7}
& \boldblue{90.9} & \boldblue{88.4} & \boldblue{81.9} & \boldblue{7.1} & \boldblue{5.1} & \boldblue{4.4}
& \boldblue{89.8} & 85.7 & 78.6 & \boldblue{6.7} & \boldblue{6.1} & \boldblue{5.1}
\\
HNOSeg-XS
& \boldblue{92.5} & 89.3 & 84.4 & \boldblue{5.5} & 4.7 & 4.1
& \boldblue{90.7} & \boldblue{88.3} & \boldblue{82.0} & \boldblue{7.3} & \boldblue{5.1} & \boldblue{4.2}
& 88.6 & \boldblue{86.7} & \boldblue{80.5} & 9.8 & \boldblue{6.4} & 5.7
\\
\myrule
\end{tabularx}

\medskip

\begin{tabularx}{\linewidth}{LGGGWWWGGGWWWGGGWWW}
\toprule
\textbf{KiTS'23} & \multicolumn{6}{c}{256$\times$256$\times$128 (training)} & \multicolumn{6}{c}{128$\times$128$\times$64 (training)} & \multicolumn{6}{c}{86$\times$86$\times$43 (training)} \\
\cline{1-1} \cline{2-7} \cline{8-13} \cline{14-19} \noalign{\smallskip}
Metric & \multicolumn{3}{c}{Dice (\%)} & \multicolumn{3}{c}{Surface Dice (\%)} & \multicolumn{3}{c}{Dice (\%)} & \multicolumn{3}{c}{Surface Dice (\%)} & \multicolumn{3}{c}{Dice (\%)} & \multicolumn{3}{c}{Surface Dice (\%)} \\
\cline{1-1} \cline{2-4} \cline{5-7} \cline{8-10} \cline{11-13} \cline{14-16} \cline{17-19} \noalign{\smallskip}
Region & KM & M & T & KM & M & T & KM & M & T & KM & M & T & KM & M & T & KM & M & T  \\
\myrule
V-Net-DS
& \boldblue{94.9} & 65.7 & 56.2 & \boldblue{93.1} & 53.8 & 44.4
& 68.6 & 26.9 & 25.4 & 58.7 & 17.4 & 15.6
& 19.2 & 9.3 & 8.8 & 12.4 & 3.6 & 3.3
\\
UTNet
& 94.8 & 67.5 & 59.2 & 92.1 & 55.0 & 47.4
& 69.6 & 28.2 & 26.0 & 50.8 & 14.4 & 12.6
& 25.9 & 11.2 & 10.2 & 13.4 & 4.1 & 3.6
\\
TransUNet
& 94.6 & 68.6 & 59.3 & 92.5 & 58.3 & 49.7
& 77.6 & 37.2 & 34.9 & 67.6 & 23.3 & 21.7
& 20.8 & 12.1 & 11.1 & 11.8 & 4.8 & 4.2
\\
nnFormer
& \boldblue{95.9} & \boldblue{77.9} & \boldblue{71.4} & \boldblue{94.5} & \boldblue{70.5} & \boldblue{64.2}
& 82.9 & 47.9 & 44.7 & 72.2 & 34.1 & 31.8
& 42.4 & 17.2 & 16.3 & 23.0 & 7.2 & 6.6
\\
FNO
& 89.1 & 27.6 & 25.4 & 80.1 & 14.1 & 13.1
& 87.4 & 22.1 & 21.4 & 75.7 & 11.6 & 11.3
& 84.3 & 16.0 & 15.7 & 69.0 & 7.5 & 7.3
\\
FNOSeg
& 92.5 & 63.2 & 57.1 & 87.4 & 49.7 & 44.3
& 91.0 & 59.6 & 51.3 & 85.4 & 45.3 & 38.8
& 86.0 & 36.0 & 31.6 & 72.8 & 20.9 & 18.5
\\
HNOSeg
& \boldblue{94.9} & \boldblue{75.1} & \boldblue{68.5} & 91.4 & \boldblue{63.4} & \boldblue{57.4}
& \boldblue{93.8} & \boldblue{68.9} & \boldblue{62.1} & \boldblue{88.3} & \boldblue{54.2} & \boldblue{48.9}
& \boldblue{89.5} & \boldblue{49.5} & \boldblue{41.6} & \boldblue{78.9} & \boldblue{33.8} & \boldblue{27.9}
\\
HNOSeg-XS
& 94.8 & 71.8 & 66.2 & 91.4 & 59.6 & 54.7
& \boldblue{92.6} & \boldblue{67.7} & \boldblue{60.9} & \boldblue{86.9} & \boldblue{51.7} & \boldblue{46.3}
& \boldblue{91.1} & \boldblue{54.8} & \boldblue{47.3} & \boldblue{82.6} & \boldblue{38.7} & \boldblue{32.8}
\\
\myrule
\end{tabularx}

\medskip

\begin{tabularx}{\linewidth}{LGGWWGGWWGGWW}
\toprule
\textbf{MVSeg'23} & \multicolumn{4}{c}{224$\times$160$\times$208 (training)} & \multicolumn{4}{c}{112$\times$80$\times$104 (training)} & \multicolumn{4}{c}{75$\times$54$\times$70 (training)} \\
\cline{1-1} \cline{2-5} \cline{6-9} \cline{10-13} \noalign{\smallskip}
Metric & \multicolumn{2}{c}{Dice (\%)} & \multicolumn{2}{c}{HD95 (mm)} & \multicolumn{2}{c}{Dice (\%)} & \multicolumn{2}{c}{HD95 (mm)} & \multicolumn{2}{c}{Dice (\%)} & \multicolumn{2}{c}{HD95 (mm)} \\
\cline{1-1} \cline{2-3} \cline{4-5} \cline{6-7} \cline{8-9} \cline{10-11} \cline{12-13} \noalign{\smallskip}
Region & PL & AL & PL & AL & PL & AL & PL & AL & PL & AL & PL & AL  \\
\myrule
V-Net-DS
& \boldblue{81.3} & \boldblue{84.6} & 4.7 & 3.6
& 62.1 & 61.9 & 14.9 & 16.4
& 30.9 & 31.4 & 24.9 & 19.3
\\
UTNet
& 81.0 & 84.0 & 4.7 & \boldblue{3.2}
& 60.2 & 59.4 & 18.6 & 15.0
& 37.6 & 31.3 & 25.6 & 23.7
\\
TransUNet
& 81.0 & 83.7 & 4.5 & 3.9
& 55.6 & 54.5 & 20.2 & 13.2
& 16.7 & 22.5 & 23.5 & 19.5
\\
nnFormer
& \boldblue{82.7} & \boldblue{85.9} & 4.3 & \boldblue{2.9}
& 68.2 & 71.5 & 8.9 & 15.8
& 45.6 & 55.4 & 15.8 & 17.7
\\
FNO
& 77.9 & 81.5 & 5.4 & 3.4
& 77.6 & \boldblue{81.7} & \boldblue{5.1} & \boldblue{3.5}
& \boldblue{71.0} & \boldblue{78.5} & \boldblue{5.5} & \boldblue{4.0}
\\
FNOSeg
& 77.7 & 80.3 & 5.2 & 3.8
& 75.8 & 76.2 & 8.6 & 5.0
& 70.0 & 69.7 & 8.1 & 5.7
\\
HNOSeg
& 80.7 & 83.1 & \boldblue{3.9} & 3.5
& \boldblue{80.1} & \boldblue{81.3} & 6.9 & \boldblue{3.1}
& \boldblue{73.3} & \boldblue{79.8} & 9.9 & \boldblue{3.7}
\\
HNOSeg-XS
& 80.7 & 83.5 & \boldblue{4.0} & \boldblue{2.9}
& \boldblue{78.7} & \boldblue{81.3} & \boldblue{5.9} & 4.9
& \boldblue{71.0} & \boldblue{78.5} & \boldblue{6.9} & 4.3
\\
\myrule
\end{tabularx}

\end{table*}

\subsection{Number of Frequency-Domain Convolutions ($\mathit{n_{XS}}$) in the HNO-XS Block} \label{sec:nxs}

Fig. \ref{fig:vs_block} shows the effect of $n_{XS}$ in (\ref{eq:hno-xs}) on accuracy. We fixed the total number of frequency-domain convolutions ($n_{XS} \times n_B$) of HNOSeg-XS to 24, and used $k_\mathrm{max} = (14, 14, 10)$ for BraTS'23, $(16, 16, 8)$ for KiTS'23, and $(14, 10, 12)$ for MVSeg'23. We want $n_{XS}$ to be large to reduce computational requirements, but large $n_{XS}$ may reduce  accuracy. To trade off between efficiency and accuracy, we used $n_{XS} = 3$ for BraTS'23, $n_{XS} = 2$ for KiTS'23, and $n_{XS} = 3$ for MVSeg'23. In consequence, comparing with HNOSeg with $n_B = 24$, only one-third or half of the Hartley transforms, and thus the tensors of full spatial sizes, were required. This reduced both memory use and computation time. 

\subsection{Numbers of Frequency Modes ($\mathit{k_\mathrm{max}}$)} \label{sec:modes}

The $k_\mathrm{max}$ is another important hyperparameter that affects the computational requirements and segmentation accuracy. Fig. \ref{fig:vs_modes} shows the effect of using different numbers of modes, from $k_\mathrm{max} = (2, 2, 2)$ to all modes\footnote{Because of the resampling approach and the frequency-domain size of $2k_{\mathrm{max},x} \times 2k_{\mathrm{max},y} \times 2k_{\mathrm{max},z}$, the total number of modes is around one-fourth of the training image size in each dimension.}. For FNO, using all modes could not be tested as this required over 1 billion parameters with out-of-memory errors.

For all datasets and models, using all modes did not provide the best results. HNOSeg had the largest drops in accuracies when using all modes. This may be explained by the Fourier transform properties. Applying convolution with the kernel size of one (Conv1) in the frequency domain on all modes, without other operations before the inverse transform, is equivalent to applying the same convolution in the spatial domain because of the linearity property. Hence, no spatial correlations can be captured. In contrast, when the frequency components larger than $k_\mathrm{max}$ are cropped, Conv1 is similar to multiplying a scaled rectangular function to each input channel in the frequency domain before summation. This is equivalent to applying a sinc function convolution to each input channel in the spatial domain by the convolution theorem, and thus better features can be learned. The effect of using all modes on FNOSeg was less severe as the real and imaginary parts had different weights. In fact, for HNOSeg, the effect of using all modes was already reduced by the nonlinearity introduced in (\ref{eq:hartley_conv_shared_nonlinear}). HNOSeg-XS had consecutive nonlinear operations in the frequency domain to further reduce such effect.

Fig. \ref{fig:vs_modes} was used to set $k_{max}$ for each model and dataset. For example, for BraTS'23, $k_{max} = (14, 14, 10)$ was chosen for FNOSeg, HNOSeg, and HNOSeg-XS, and $k_{max} = (6, 6, 4)$ for FNO. In case when the training image size was smaller than $2k_{max}$, the next smaller $k_{max}$ in Fig. \ref{fig:vs_modes} was used.

\begin{figure}[t]
\fontsize{6}{7}\selectfont
    \centering
    \begin{minipage}[t]{0.19\linewidth}
      \vspace{24.5em}
      \centering{Ground truth} \\
      \includegraphics[width=\linewidth]{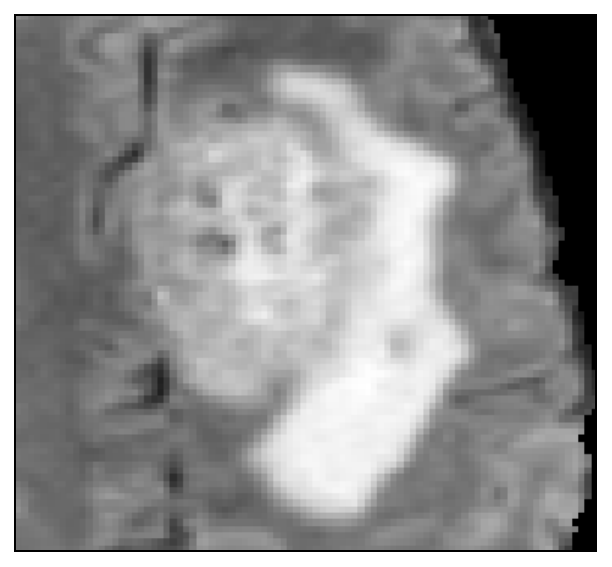}\\
      \includegraphics[width=\linewidth]{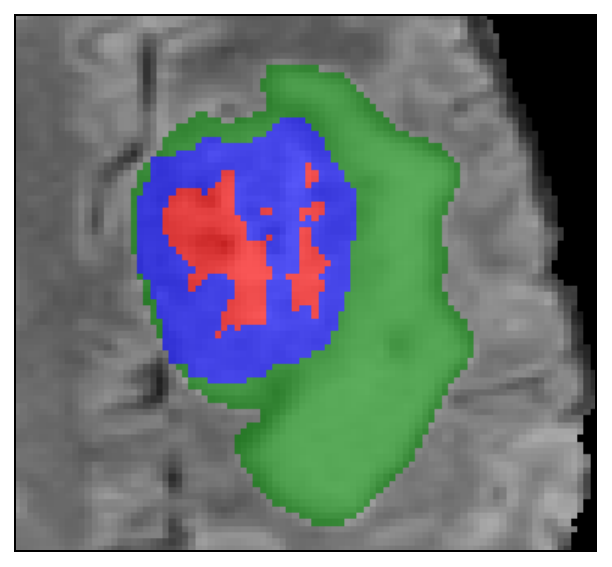}\\
    \end{minipage}
    \begin{minipage}[t]{0.79\linewidth}
        \hrule
        \vspace{0.5em}
        \begin{minipage}[t]{\linewidth}
          \centering{V-Net-DS}
        \end{minipage}
        \begin{minipage}[t]{0.24\linewidth}
          \includegraphics[width=\linewidth]{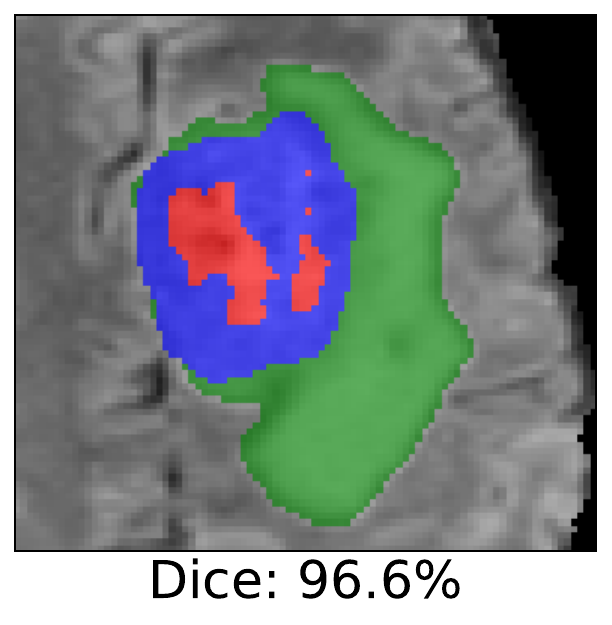}
        \end{minipage}
        \begin{minipage}[t]{0.24\linewidth}
          \includegraphics[width=\linewidth]{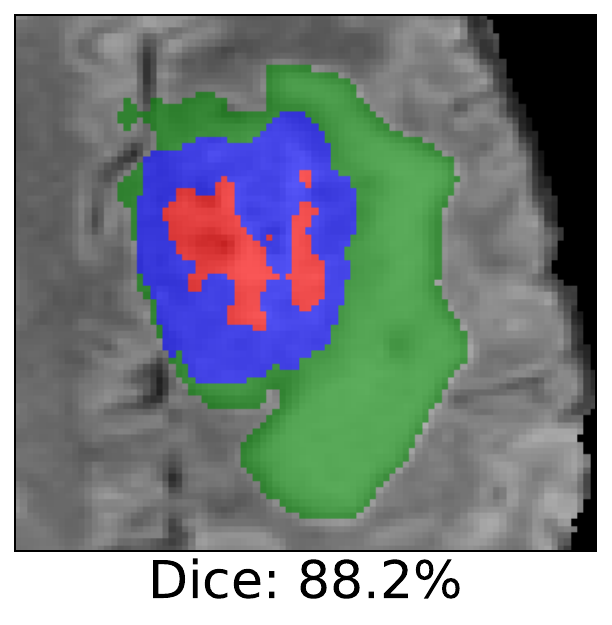}
        \end{minipage}
        \begin{minipage}[t]{0.24\linewidth}
          \includegraphics[width=\linewidth]{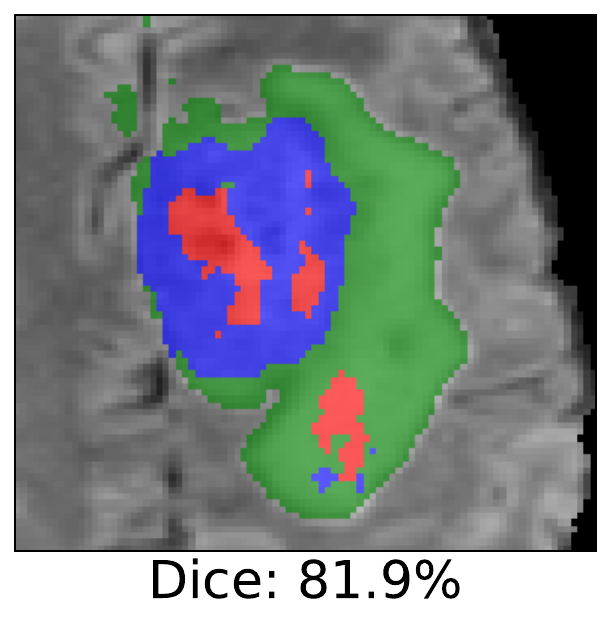}
        \end{minipage}
        \begin{minipage}[t]{0.24\linewidth}
          \includegraphics[width=\linewidth]{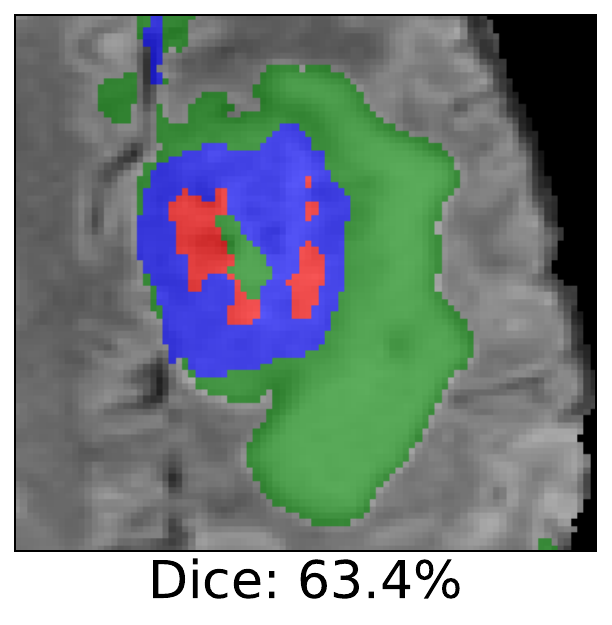}
        \end{minipage}
        \hrule
        \vspace{0.5em}
        \begin{minipage}[t]{\linewidth}
          \centering{UTNet}
        \end{minipage}
        \begin{minipage}[t]{0.24\linewidth}
          \includegraphics[width=\linewidth]{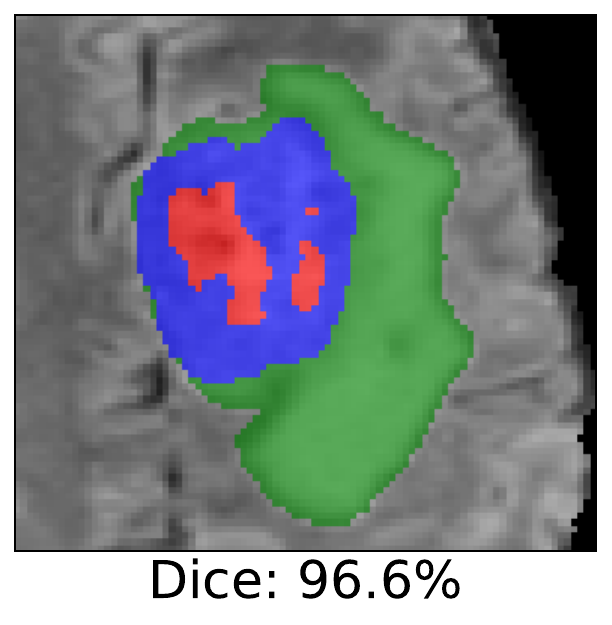}
        \end{minipage}
        \begin{minipage}[t]{0.24\linewidth}
          \includegraphics[width=\linewidth]{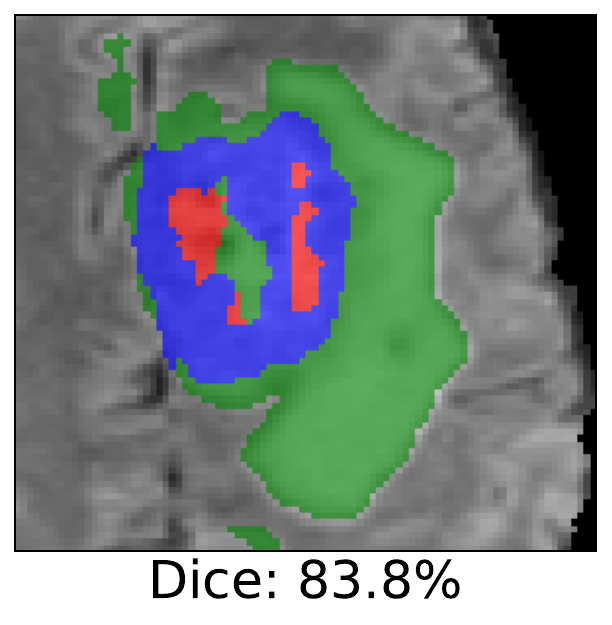}
        \end{minipage}
        \begin{minipage}[t]{0.24\linewidth}
          \includegraphics[width=\linewidth]{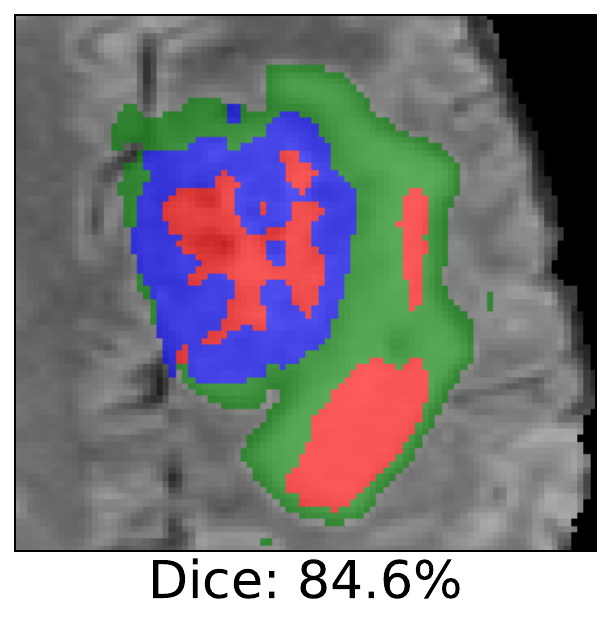}
        \end{minipage}
        \begin{minipage}[t]{0.24\linewidth}
          \includegraphics[width=\linewidth]{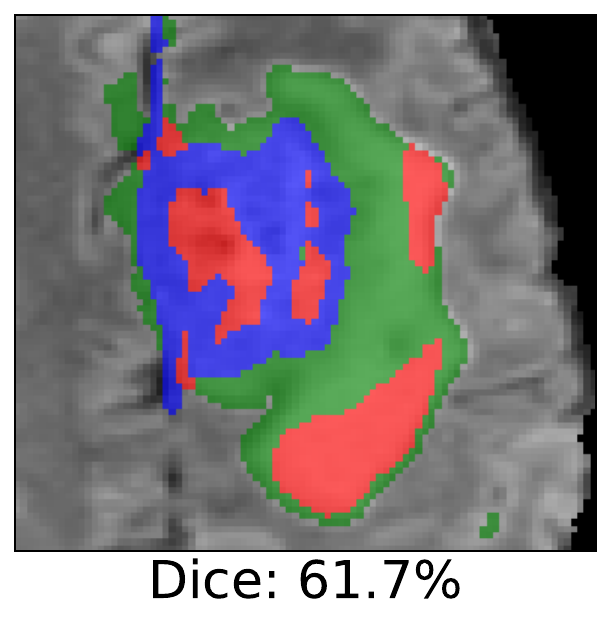}
        \end{minipage}
        \hrule
        \vspace{0.5em}
        \begin{minipage}[t]{\linewidth}
          \centering{TransUNet}
        \end{minipage}
        \begin{minipage}[t]{0.24\linewidth}
          \includegraphics[width=\linewidth]{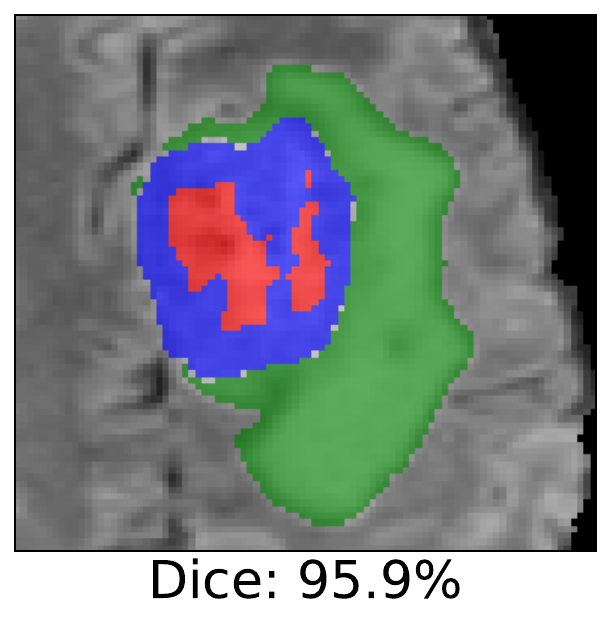}
        \end{minipage}
        \begin{minipage}[t]{0.24\linewidth}
          \includegraphics[width=\linewidth]{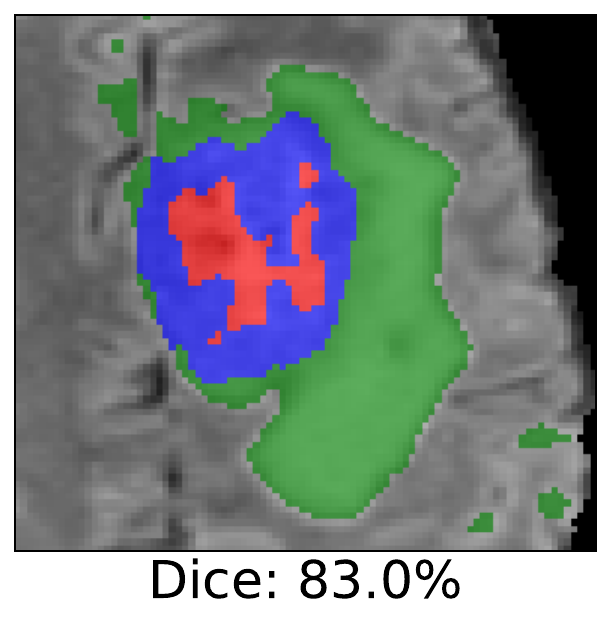}
        \end{minipage}
        \begin{minipage}[t]{0.24\linewidth}
          \includegraphics[width=\linewidth]{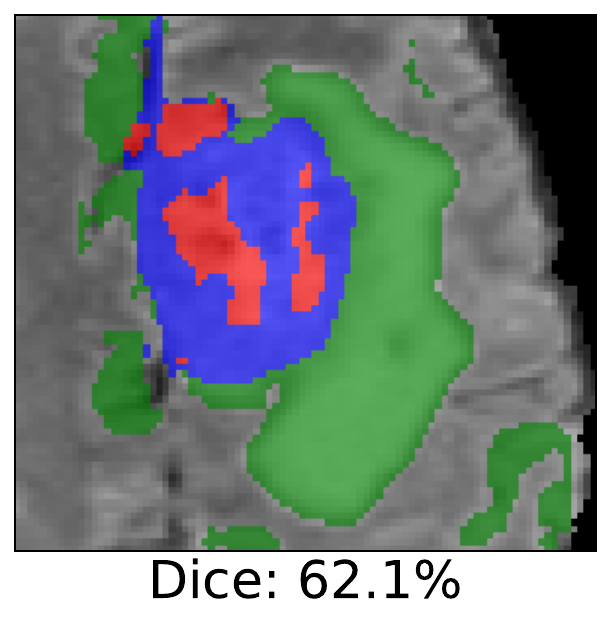}
        \end{minipage}
        \begin{minipage}[t]{0.24\linewidth}
          \includegraphics[width=\linewidth]{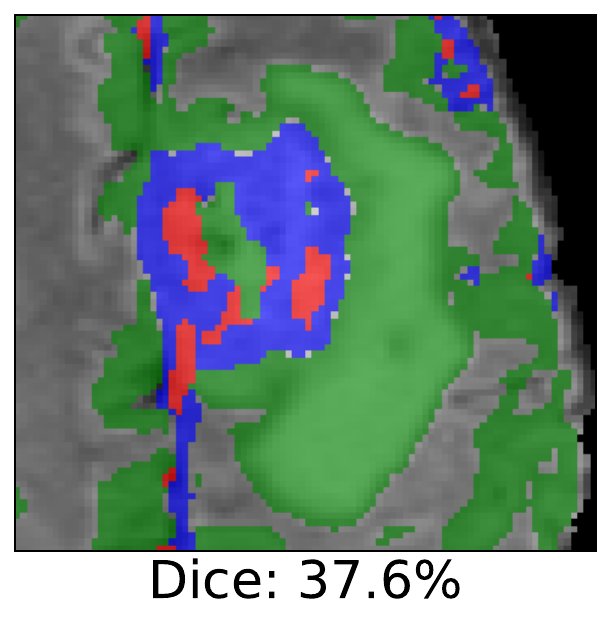}
        \end{minipage}
        \hrule
        \vspace{0.5em}
        \begin{minipage}[t]{\linewidth}
          \centering{nnFormer}
        \end{minipage}
        \begin{minipage}[t]{0.24\linewidth}
          \includegraphics[width=\linewidth]{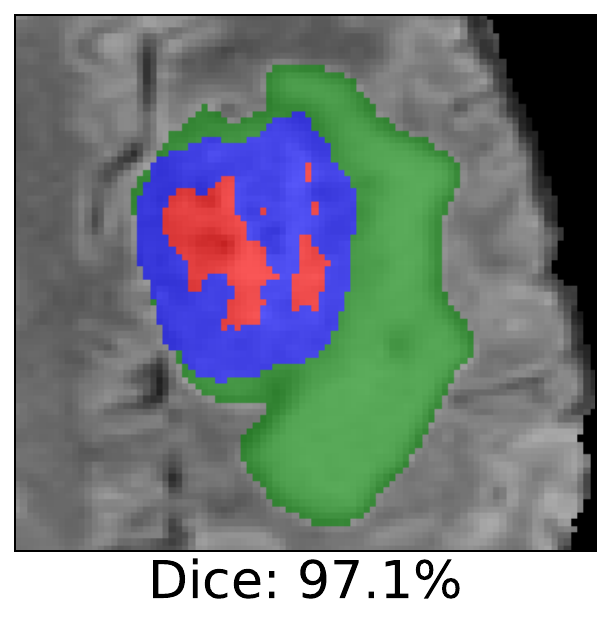}
        \end{minipage}
        \begin{minipage}[t]{0.24\linewidth}
          \includegraphics[width=\linewidth]{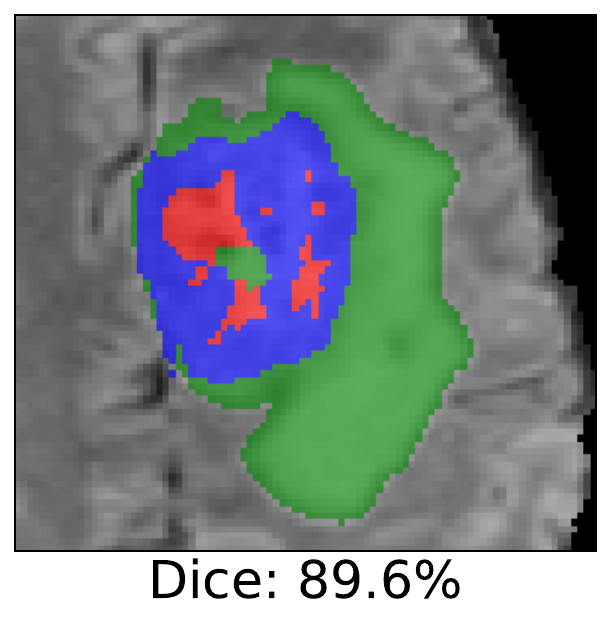}
        \end{minipage}
        \begin{minipage}[t]{0.24\linewidth}
          \includegraphics[width=\linewidth]{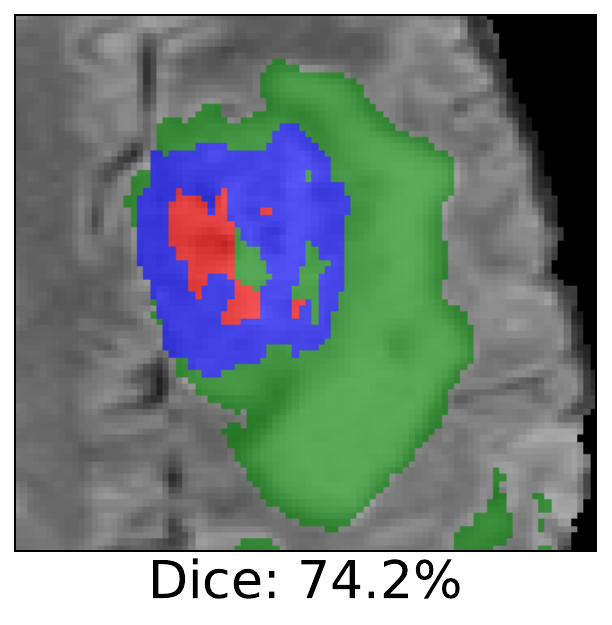}
        \end{minipage}
        \begin{minipage}[t]{0.24\linewidth}
          \includegraphics[width=\linewidth]{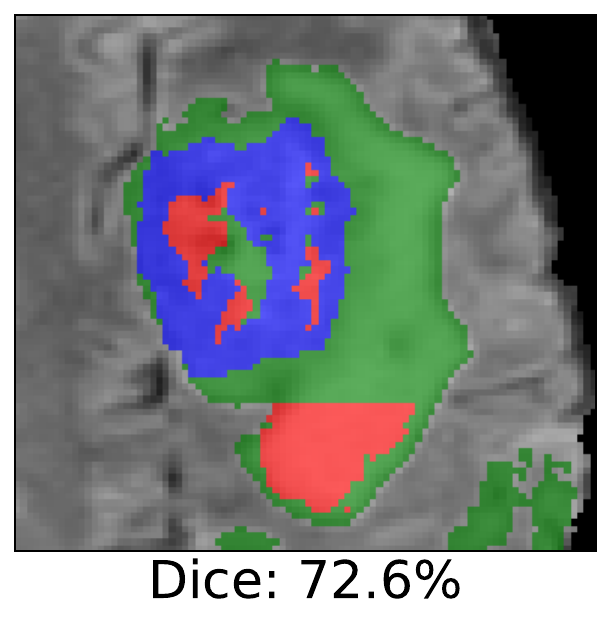}
        \end{minipage}
        \hrule
        \vspace{0.5em}
        \begin{minipage}[t]{\linewidth}
          \centering{FNO}
        \end{minipage}
        \begin{minipage}[t]{0.24\linewidth}
          \includegraphics[width=\linewidth]{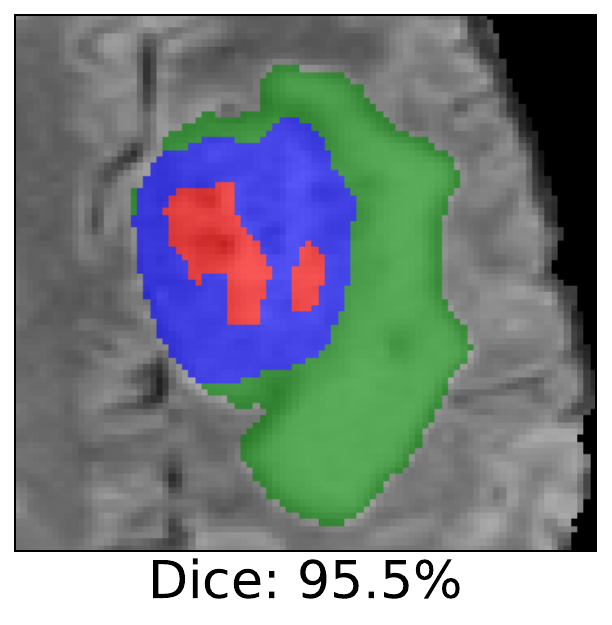}
        \end{minipage}
        \begin{minipage}[t]{0.24\linewidth}
          \includegraphics[width=\linewidth]{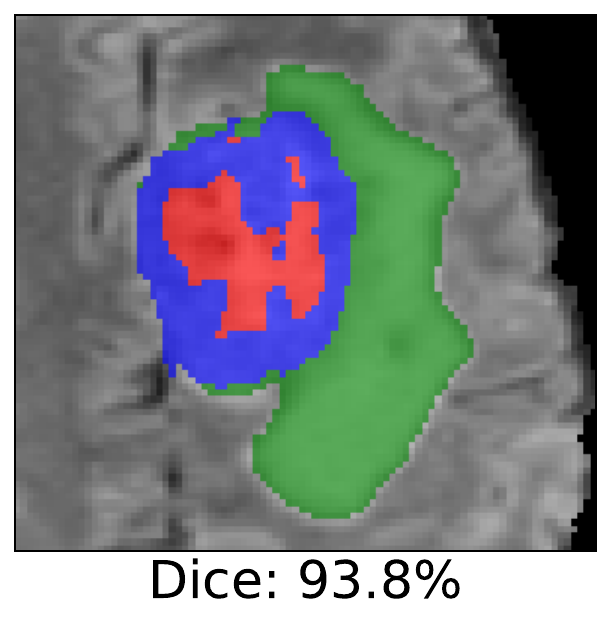}
        \end{minipage}
        \begin{minipage}[t]{0.24\linewidth}
          \includegraphics[width=\linewidth]{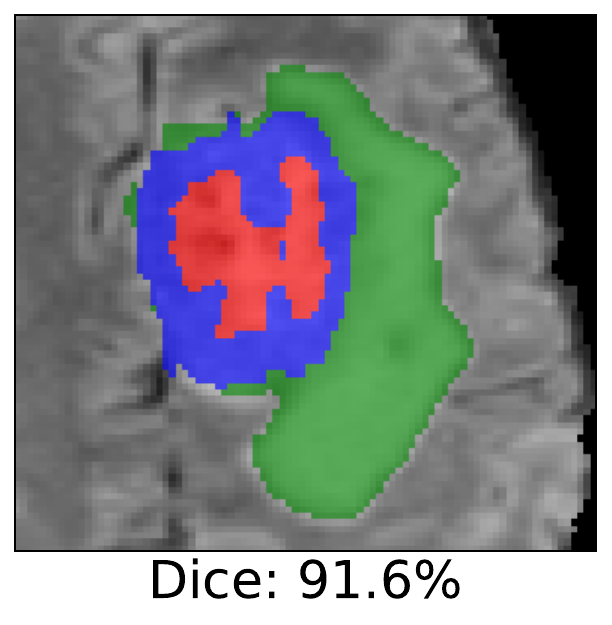}
        \end{minipage}
        \begin{minipage}[t]{0.24\linewidth}
          \includegraphics[width=\linewidth]{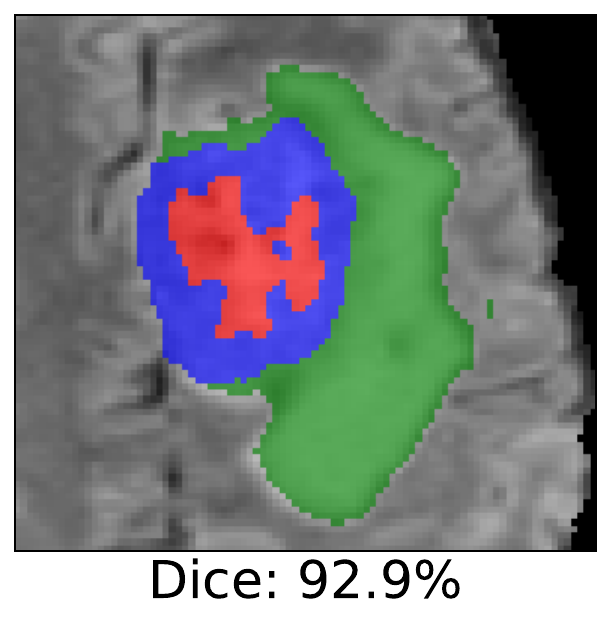}
        \end{minipage}
        \hrule
        \vspace{0.5em}
        \begin{minipage}[t]{\linewidth}
          \centering{FNOSeg}
        \end{minipage}
        \begin{minipage}[t]{0.24\linewidth}
          \includegraphics[width=\linewidth]{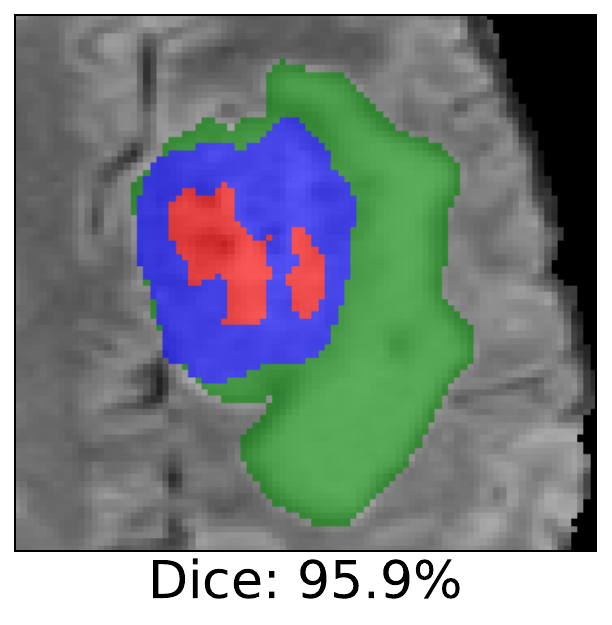}
        \end{minipage}
        \begin{minipage}[t]{0.24\linewidth}
          \includegraphics[width=\linewidth]{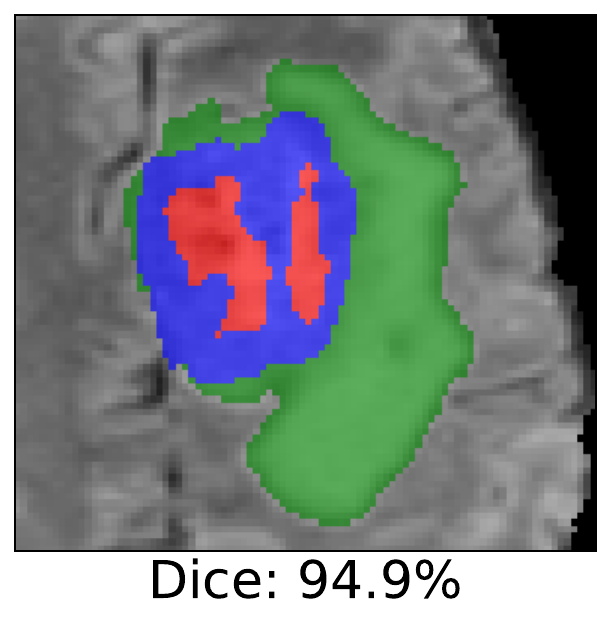}
        \end{minipage}
        \begin{minipage}[t]{0.24\linewidth}
          \includegraphics[width=\linewidth]{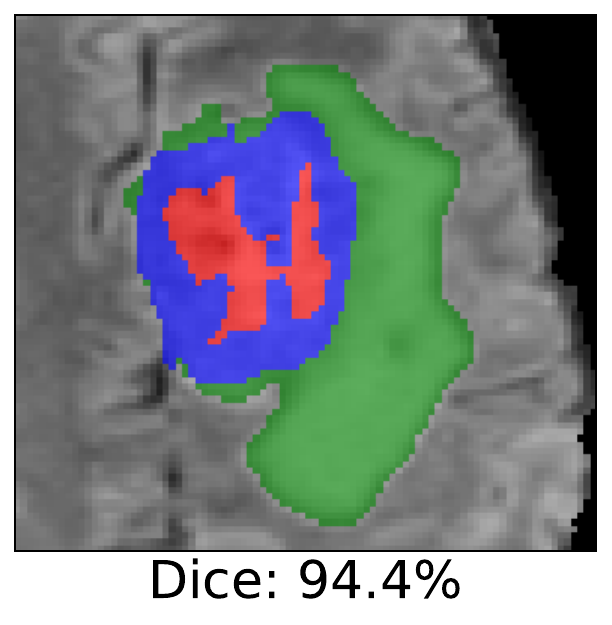}
        \end{minipage}
        \begin{minipage}[t]{0.24\linewidth}
          \includegraphics[width=\linewidth]{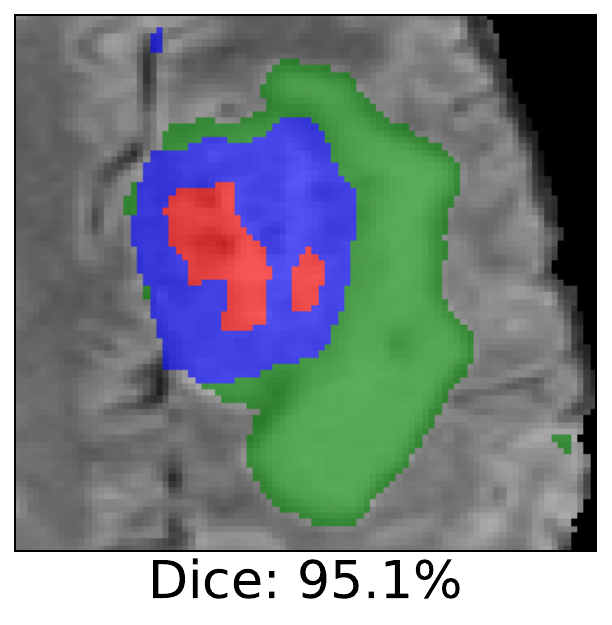}
        \end{minipage}
        \hrule
        \vspace{0.5em}
        \begin{minipage}[t]{\linewidth}
          \centering{HNOSeg}
        \end{minipage}
        \begin{minipage}[t]{0.24\linewidth}
          \includegraphics[width=\linewidth]{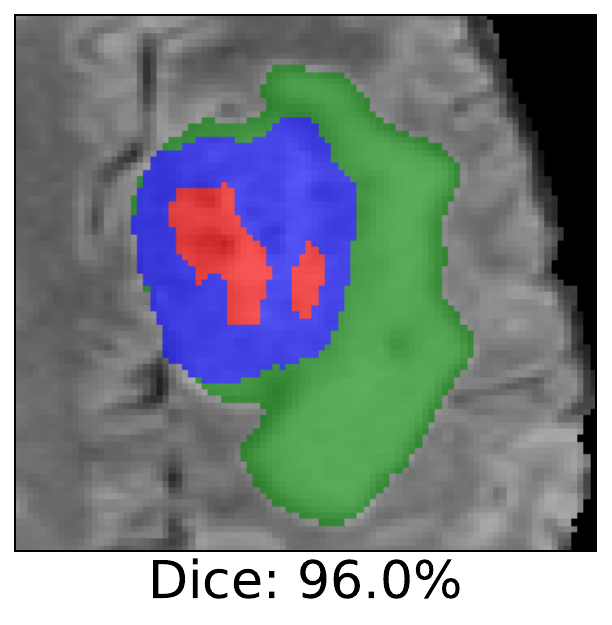}
        \end{minipage}
        \begin{minipage}[t]{0.24\linewidth}
          \includegraphics[width=\linewidth]{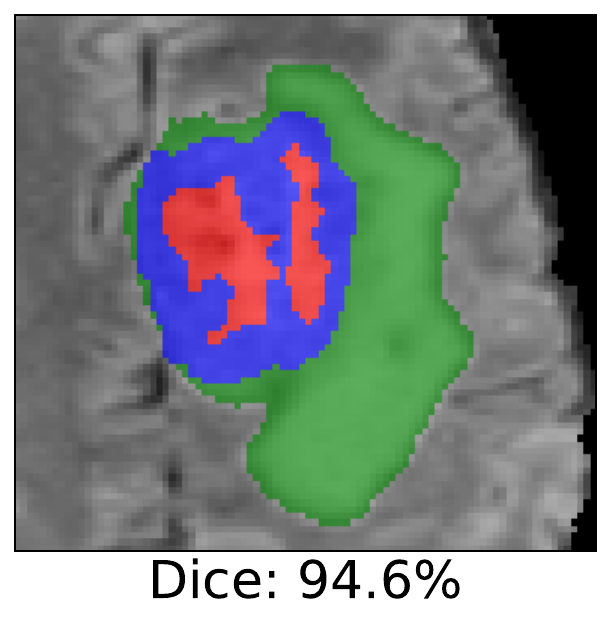}
        \end{minipage}
        \begin{minipage}[t]{0.24\linewidth}
          \includegraphics[width=\linewidth]{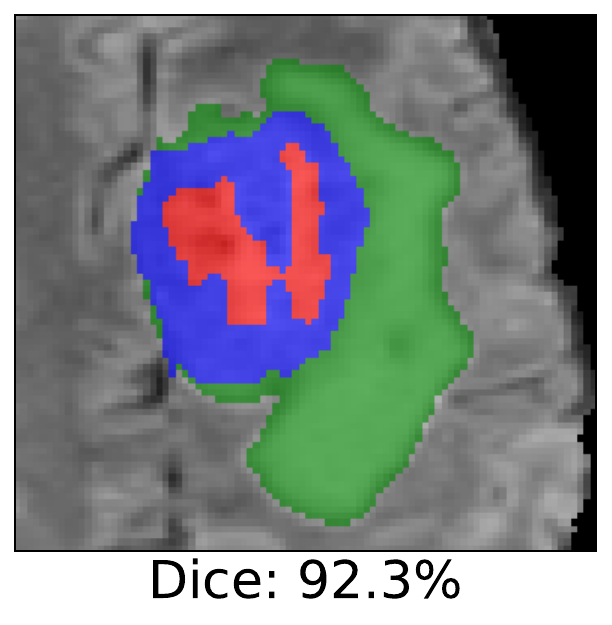}
        \end{minipage}
        \begin{minipage}[t]{0.24\linewidth}
          \includegraphics[width=\linewidth]{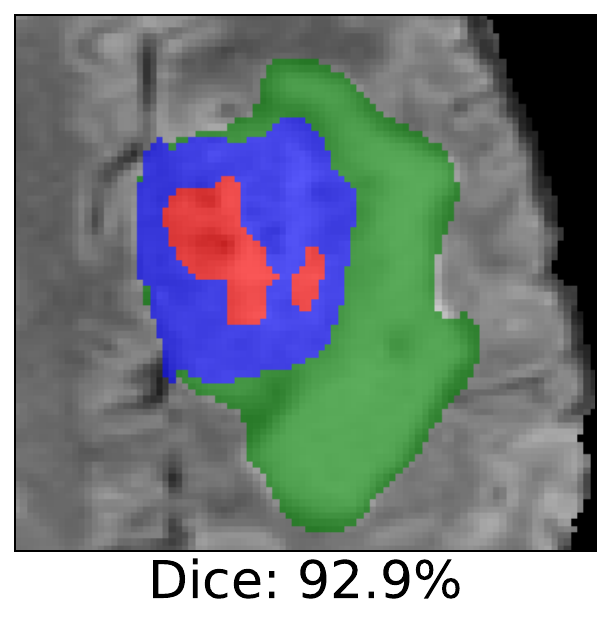}
        \end{minipage}
        \hrule
        \vspace{0.5em}
        \begin{minipage}[t]{\linewidth}
          \centering{HNOSeg-XS}
        \end{minipage}
        \begin{minipage}[t]{0.24\linewidth}
          \includegraphics[width=\linewidth]{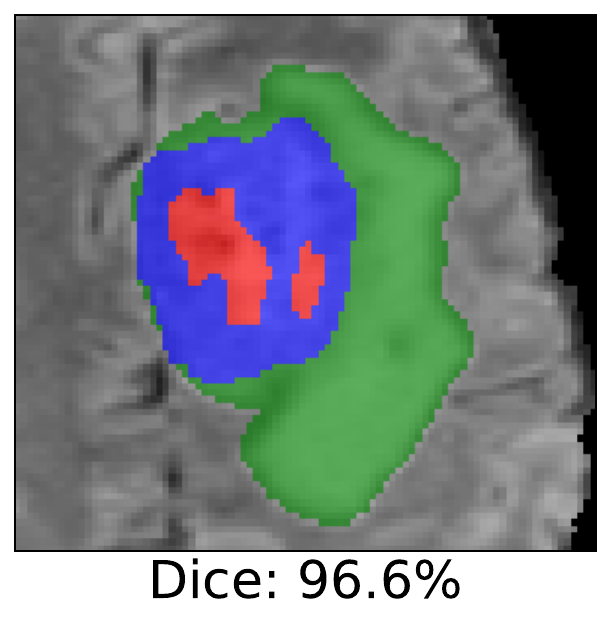}
        \end{minipage}
        \begin{minipage}[t]{0.24\linewidth}
          \includegraphics[width=\linewidth]{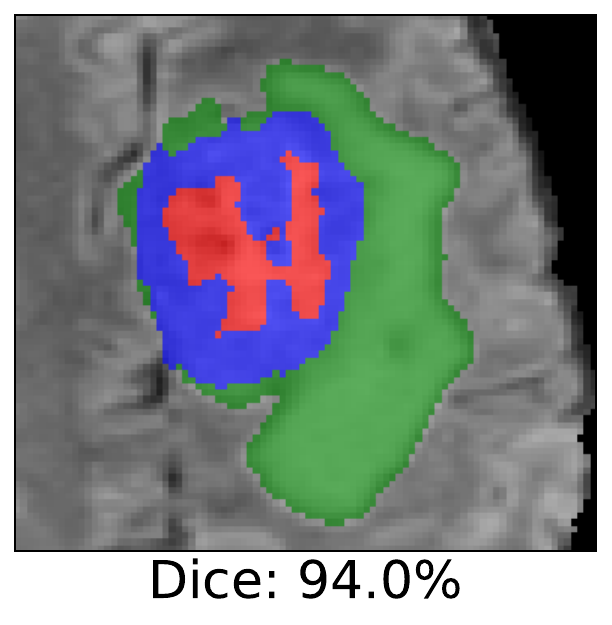}
        \end{minipage}
        \begin{minipage}[t]{0.24\linewidth}
          \includegraphics[width=\linewidth]{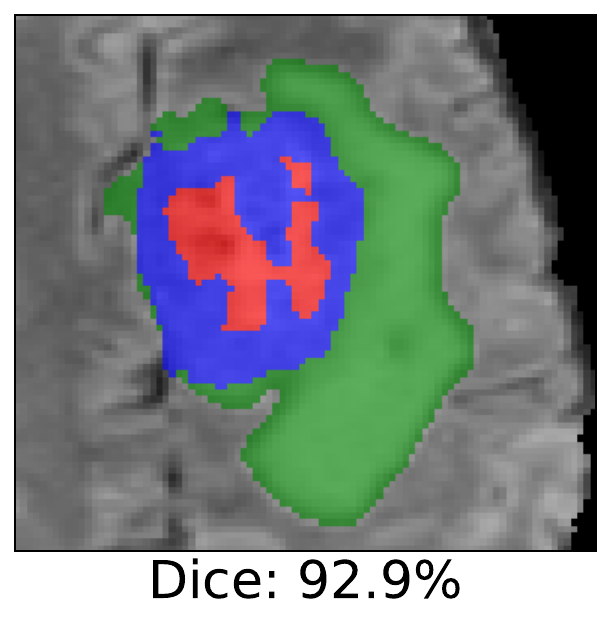}
        \end{minipage}
        \begin{minipage}[t]{0.24\linewidth}
          \includegraphics[width=\linewidth]{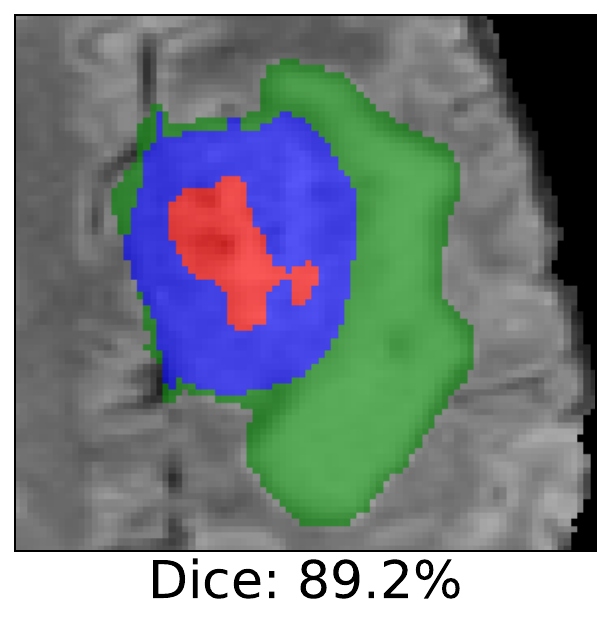}
        \end{minipage}
        \hrule
        \vspace{0.5em}
        \begin{minipage}[t]{0.24\linewidth}
          \centering{240$\times$240$\times$155}
        \end{minipage}
        \begin{minipage}[t]{0.24\linewidth}
          \centering{120$\times$120$\times$78}
        \end{minipage}
        \begin{minipage}[t]{0.24\linewidth}
          \centering{80$\times$80$\times$52}
        \end{minipage}
        \begin{minipage}[t]{0.24\linewidth}
          \centering{60$\times$60$\times$39}
        \end{minipage}
    \end{minipage}
    \caption{Visual comparisons of models trained with different image resolutions (BraTS'23), tested on an unseen sample of size 240$\times$240$\times$155. The Dice coefficients were averaged from the WT, TC, and ET regions. Note that the 2D visualizations may not reflect the Dice coefficients between 3D volumes.}
    \label{fig:brats_visual}
\end{figure}

\begin{figure}[t]
\fontsize{6}{7}\selectfont
    \centering
    \begin{minipage}[t]{1\linewidth}
      \centering{Ground truth} \\
      \centering{
      \begin{minipage}[t]{0.245\linewidth}
        \includegraphics[width=\linewidth]{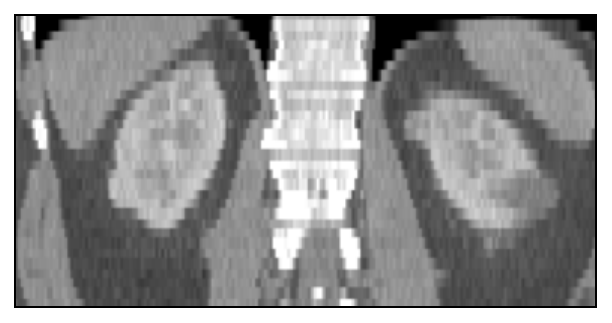}
      \end{minipage}
      \begin{minipage}[t]{0.245\linewidth}
      \includegraphics[width=\linewidth]{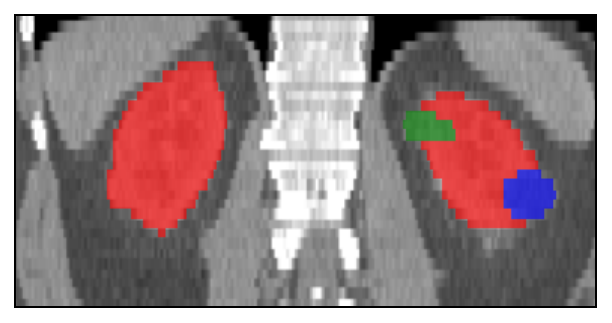}
      \end{minipage}
      }
    \end{minipage}
    \\
    \smallskip
    \begin{minipage}[t]{1\linewidth}
        \hrule
        \vspace{0.5em}
        \begin{minipage}[t]{\linewidth}
          \centering{V-Net-DS}
        \end{minipage}
        \begin{minipage}[t]{0.245\linewidth}
          \includegraphics[width=\linewidth]{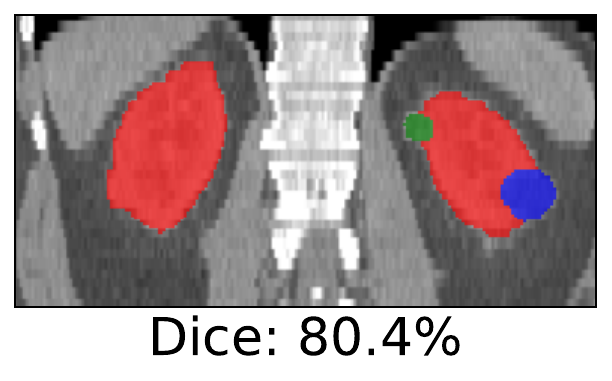}
        \end{minipage}
        \begin{minipage}[t]{0.245\linewidth}
          \includegraphics[width=\linewidth]{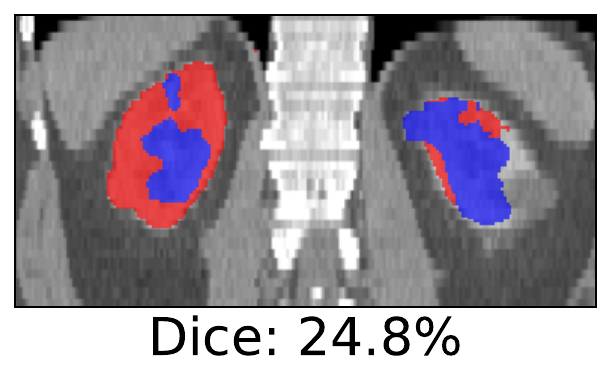}
        \end{minipage}
        \begin{minipage}[t]{0.245\linewidth}
          \includegraphics[width=\linewidth]{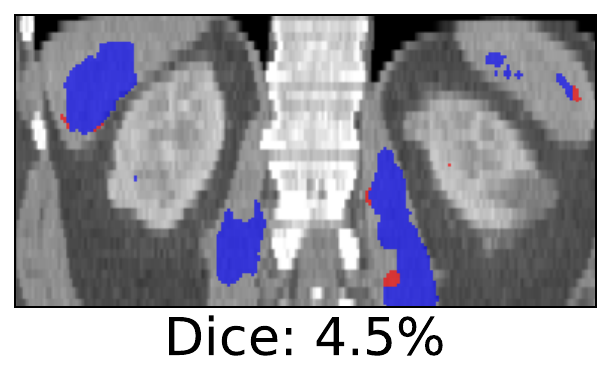}
        \end{minipage}
        \begin{minipage}[t]{0.245\linewidth}
          \includegraphics[width=\linewidth]{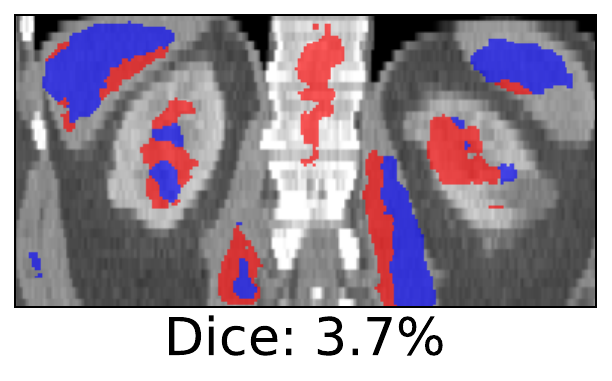}
        \end{minipage}
        \hrule
        \vspace{0.5em}
        \begin{minipage}[t]{\linewidth}
          \centering{UTNet}
        \end{minipage}
        \begin{minipage}[t]{0.245\linewidth}
          \includegraphics[width=\linewidth]{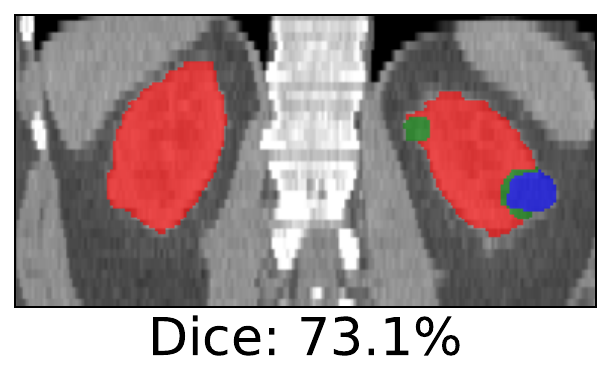}
        \end{minipage}
        \begin{minipage}[t]{0.245\linewidth}
          \includegraphics[width=\linewidth]{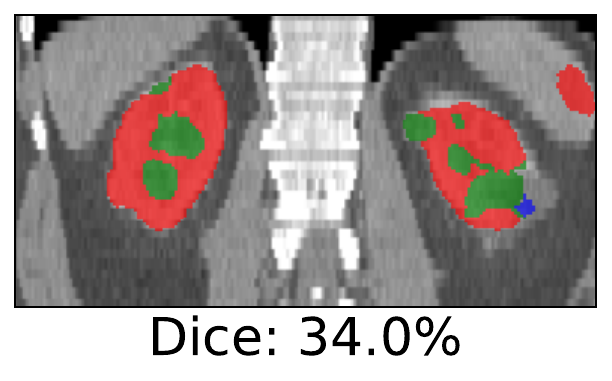}
        \end{minipage}
        \begin{minipage}[t]{0.245\linewidth}
          \includegraphics[width=\linewidth]{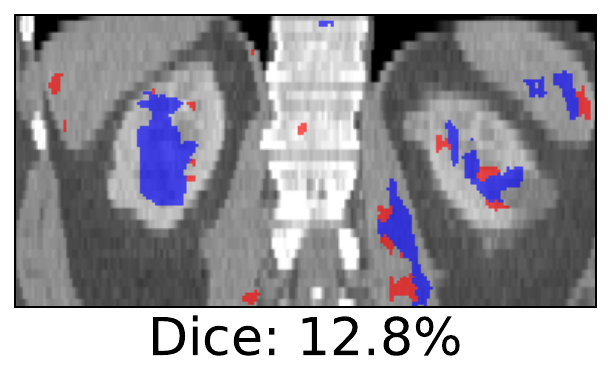}
        \end{minipage}
        \begin{minipage}[t]{0.245\linewidth}
          \includegraphics[width=\linewidth]{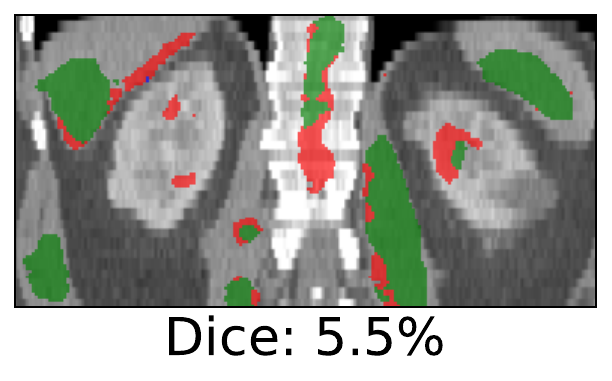}
        \end{minipage}
        \hrule
        \vspace{0.5em}
        \begin{minipage}[t]{\linewidth}
          \centering{TransUNet}
        \end{minipage}
        \begin{minipage}[t]{0.245\linewidth}
          \includegraphics[width=\linewidth]{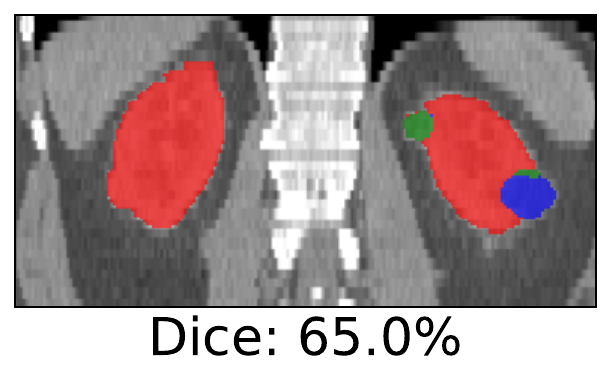}
        \end{minipage}
        \begin{minipage}[t]{0.245\linewidth}
          \includegraphics[width=\linewidth]{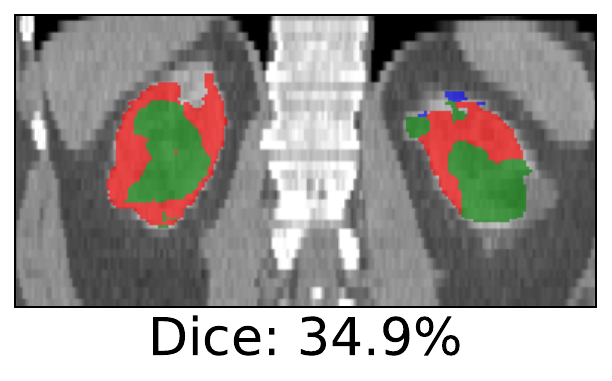}
        \end{minipage}
        \begin{minipage}[t]{0.245\linewidth}
          \includegraphics[width=\linewidth]{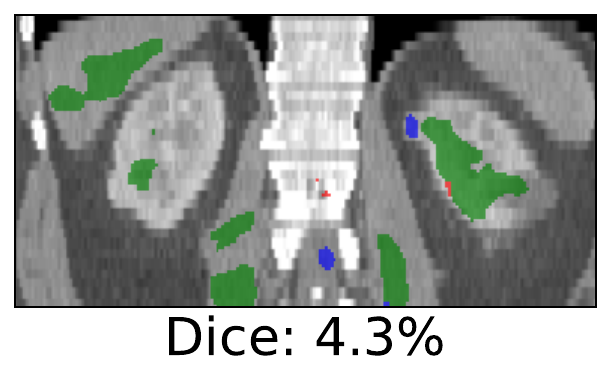}
        \end{minipage}
        \begin{minipage}[t]{0.245\linewidth}
          \includegraphics[width=\linewidth]{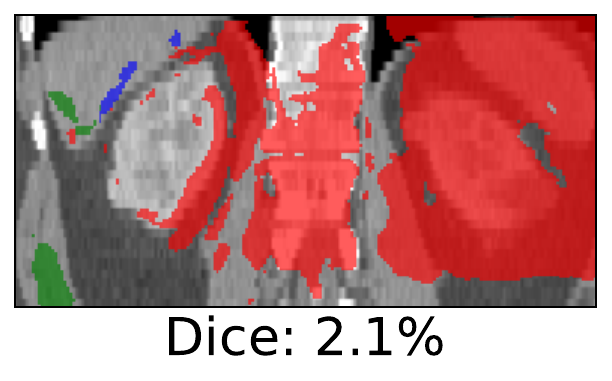}
        \end{minipage}
        \hrule
        \vspace{0.5em}
        \begin{minipage}[t]{\linewidth}
          \centering{nnFormer}
        \end{minipage}
        \begin{minipage}[t]{0.245\linewidth}
          \includegraphics[width=\linewidth]{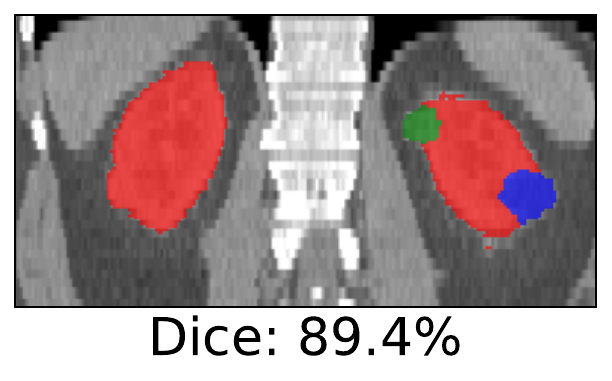}
        \end{minipage}
        \begin{minipage}[t]{0.245\linewidth}
          \includegraphics[width=\linewidth]{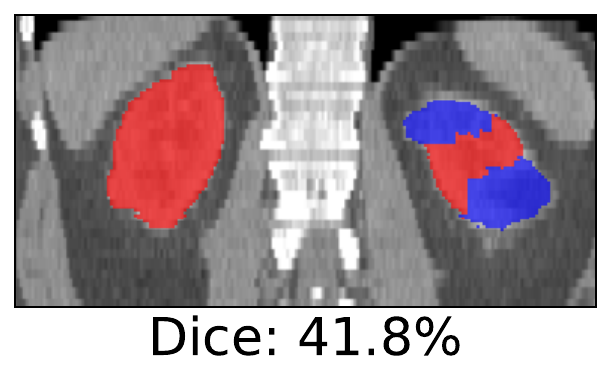}
        \end{minipage}
        \begin{minipage}[t]{0.245\linewidth}
          \includegraphics[width=\linewidth]{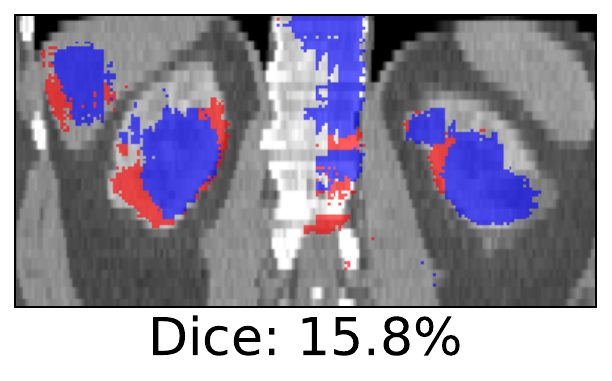}
        \end{minipage}
        \begin{minipage}[t]{0.245\linewidth}
          \includegraphics[width=\linewidth]{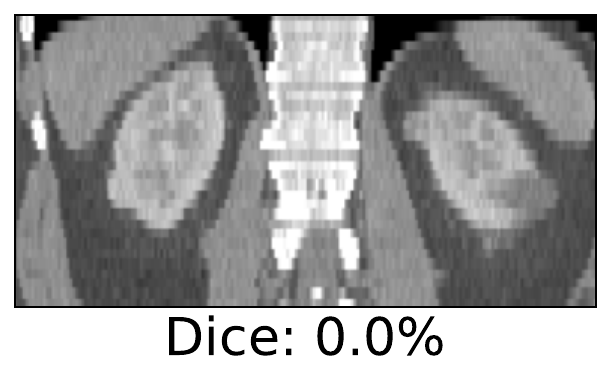}
        \end{minipage}
        \hrule
        \vspace{0.5em}
        \begin{minipage}[t]{\linewidth}
          \centering{FNO}
        \end{minipage}
        \begin{minipage}[t]{0.245\linewidth}
          \includegraphics[width=\linewidth]{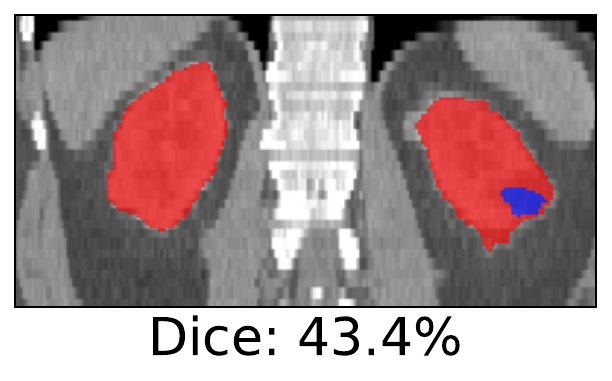}
        \end{minipage}
        \begin{minipage}[t]{0.245\linewidth}
          \includegraphics[width=\linewidth]{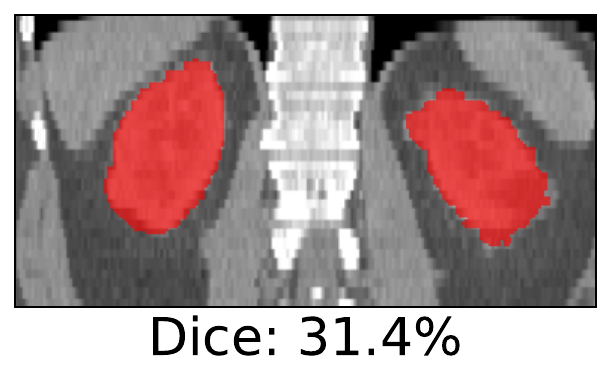}
        \end{minipage}
        \begin{minipage}[t]{0.245\linewidth}
          \includegraphics[width=\linewidth]{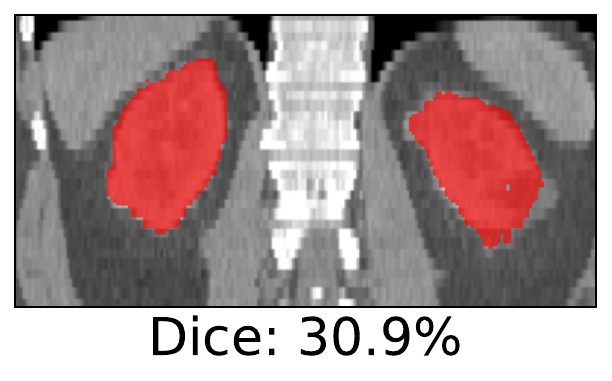}
        \end{minipage}
        \begin{minipage}[t]{0.245\linewidth}
          \includegraphics[width=\linewidth]{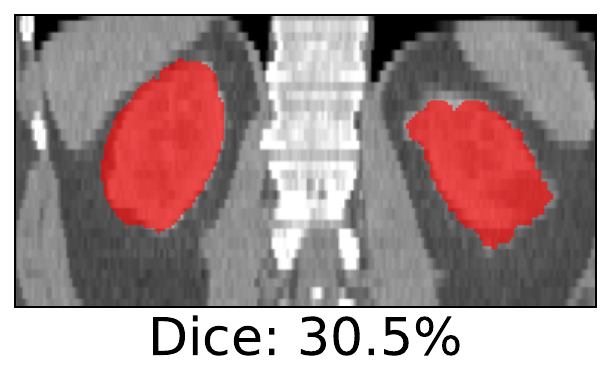}
        \end{minipage}
        \hrule
        \vspace{0.5em}
        \begin{minipage}[t]{\linewidth}
          \centering{FNOSeg}
        \end{minipage}
        \begin{minipage}[t]{0.245\linewidth}
          \includegraphics[width=\linewidth]{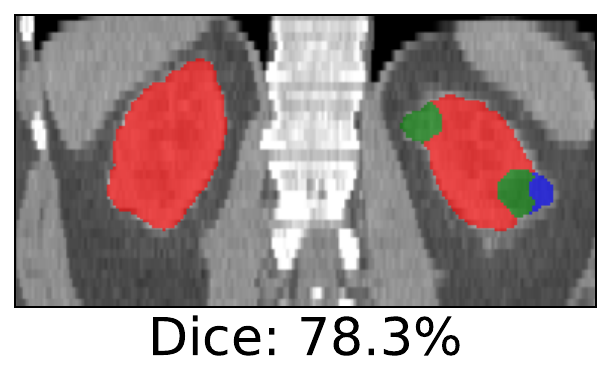}
        \end{minipage}
        \begin{minipage}[t]{0.245\linewidth}
          \includegraphics[width=\linewidth]{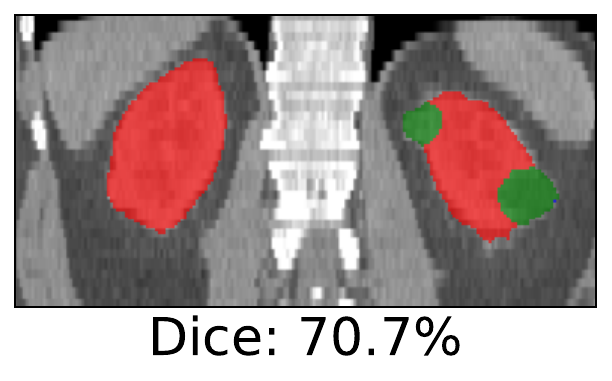}
        \end{minipage}
        \begin{minipage}[t]{0.245\linewidth}
          \includegraphics[width=\linewidth]{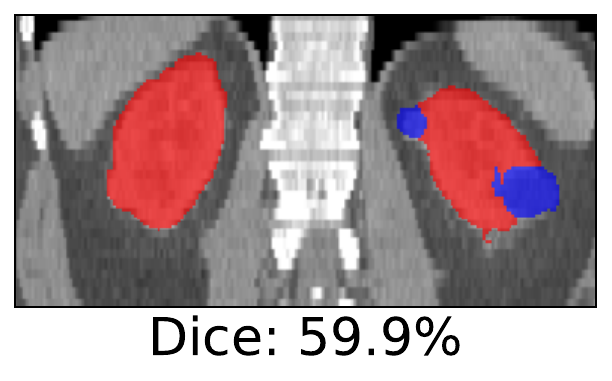}
        \end{minipage}
        \begin{minipage}[t]{0.245\linewidth}
          \includegraphics[width=\linewidth]{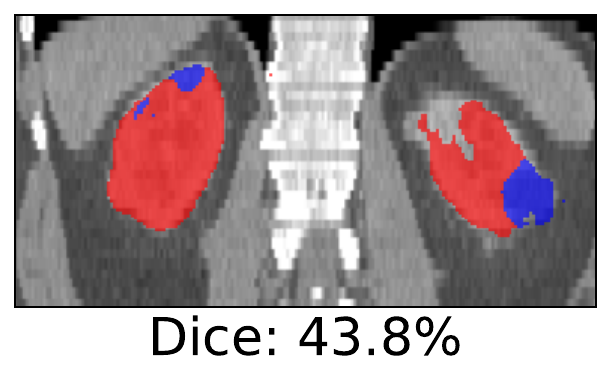}
        \end{minipage}
        \hrule
        \vspace{0.5em}
        \begin{minipage}[t]{\linewidth}
          \centering{HNOSeg}
        \end{minipage}
        \begin{minipage}[t]{0.245\linewidth}
          \includegraphics[width=\linewidth]{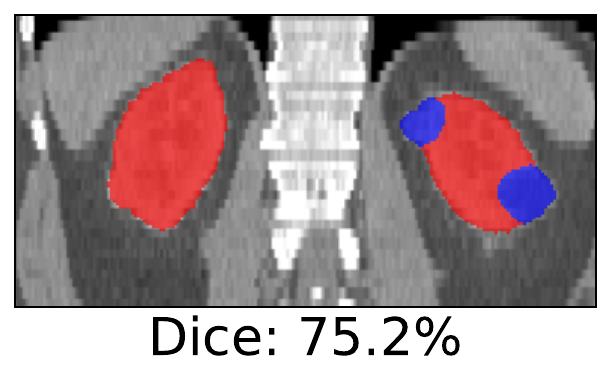}
        \end{minipage}
        \begin{minipage}[t]{0.245\linewidth}
          \includegraphics[width=\linewidth]{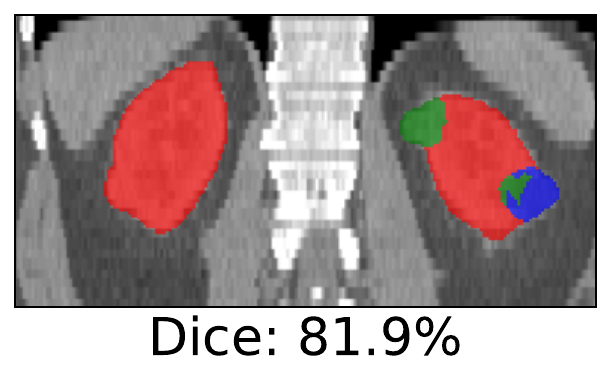}
        \end{minipage}
        \begin{minipage}[t]{0.245\linewidth}
          \includegraphics[width=\linewidth]{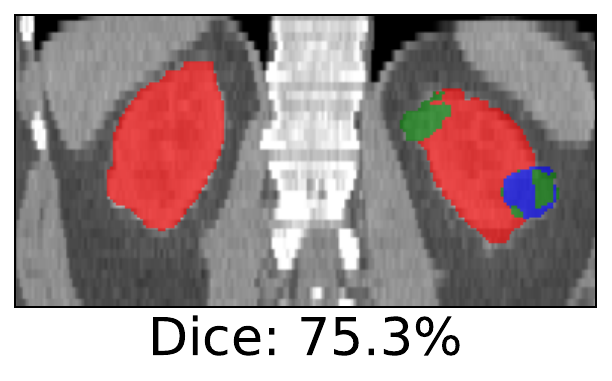}
        \end{minipage}
        \begin{minipage}[t]{0.245\linewidth}
          \includegraphics[width=\linewidth]{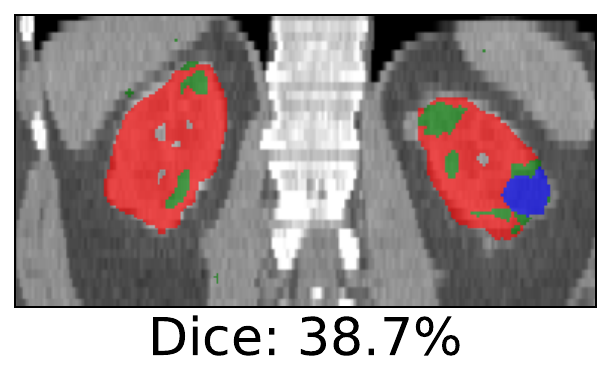}
        \end{minipage}
        \hrule
        \vspace{0.5em}
        \begin{minipage}[t]{\linewidth}
          \centering{HNOSeg-XS}
        \end{minipage}
        \begin{minipage}[t]{0.245\linewidth}
          \includegraphics[width=\linewidth]{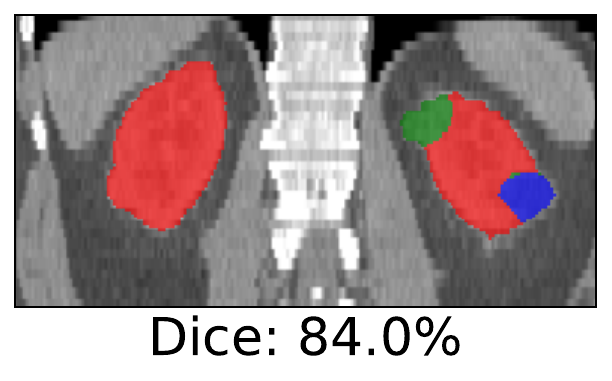}
        \end{minipage}
        \begin{minipage}[t]{0.245\linewidth}
          \includegraphics[width=\linewidth]{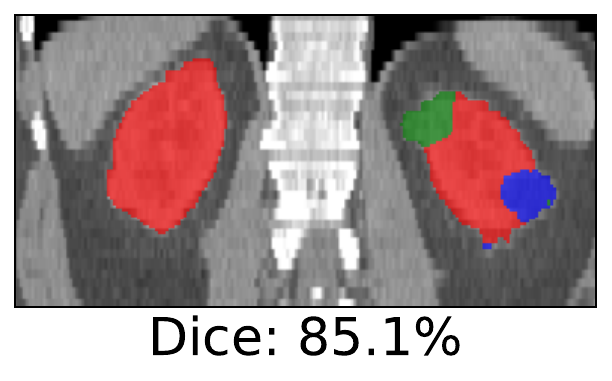}
        \end{minipage}
        \begin{minipage}[t]{0.245\linewidth}
          \includegraphics[width=\linewidth]{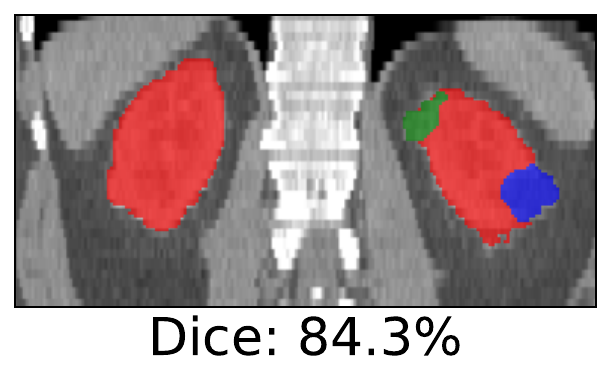}
        \end{minipage}
        \begin{minipage}[t]{0.245\linewidth}
          \includegraphics[width=\linewidth]{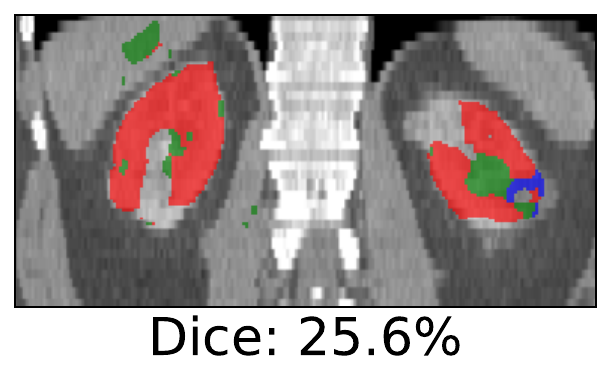}
        \end{minipage}
        \hrule
        \vspace{0.5em}
        \begin{minipage}[t]{0.245\linewidth}
          \centering{256$\times$256$\times$128}
        \end{minipage}
        \begin{minipage}[t]{0.245\linewidth}
          \centering{128$\times$128$\times$64}
        \end{minipage}
        \begin{minipage}[t]{0.245\linewidth}
          \centering{86$\times$86$\times$43}
        \end{minipage}
        \begin{minipage}[t]{0.245\linewidth}
          \centering{64$\times$64$\times$32}
        \end{minipage}
    \end{minipage}
    \caption{Visual comparisons of models trained with different image resolutions (KiTS'23), tested on an unseen sample of size 256$\times$256$\times$128. The Dice coefficients were averaged from the KM, M, and T regions. Note that the 2D visualizations may not reflect the Dice coefficients between 3D volumes.}
    \label{fig:kits_visual}
\end{figure}

\begin{figure}[t]
\fontsize{6}{7}\selectfont
    \centering
    \begin{minipage}[t]{0.19\linewidth}
      \vspace{24.5em}
      \centering{Ground truth} \\
      \includegraphics[width=\linewidth]{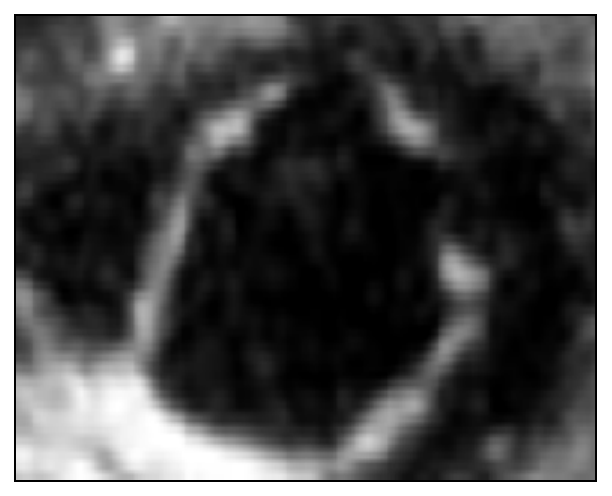}\\
      \includegraphics[width=\linewidth]{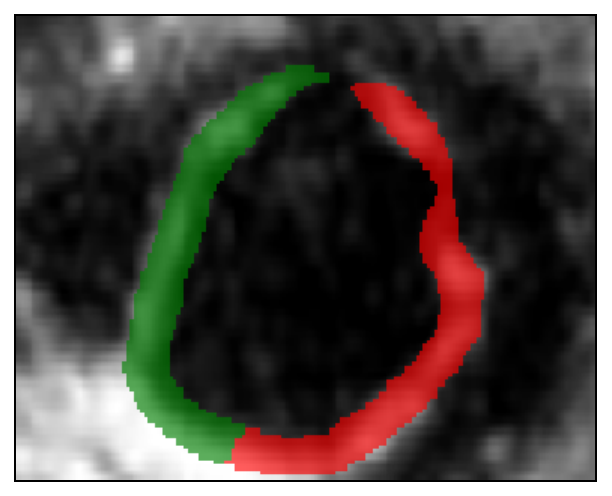}\\
    \end{minipage}
    \begin{minipage}[t]{0.79\linewidth}
        \hrule
        \vspace{0.5em}
        \begin{minipage}[t]{\linewidth}
          \centering{V-Net-DS}
        \end{minipage}
        \begin{minipage}[t]{0.24\linewidth}
          \includegraphics[width=\linewidth]{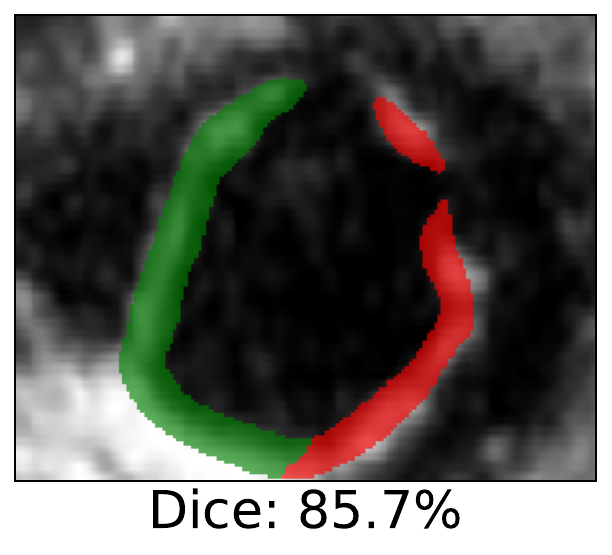}
        \end{minipage}
        \begin{minipage}[t]{0.24\linewidth}
          \includegraphics[width=\linewidth]{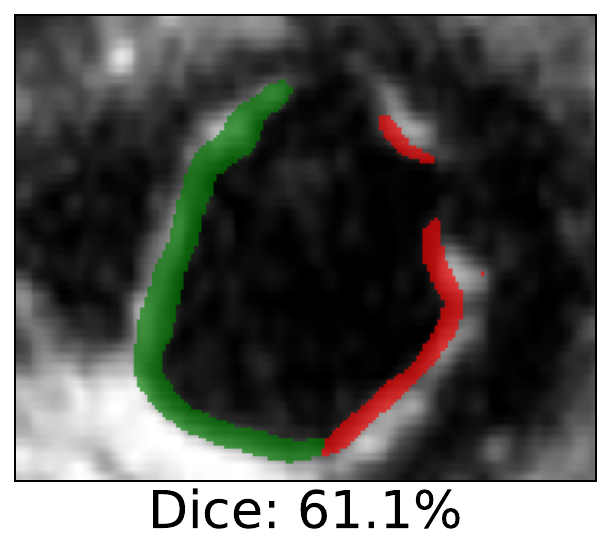}
        \end{minipage}
        \begin{minipage}[t]{0.24\linewidth}
          \includegraphics[width=\linewidth]{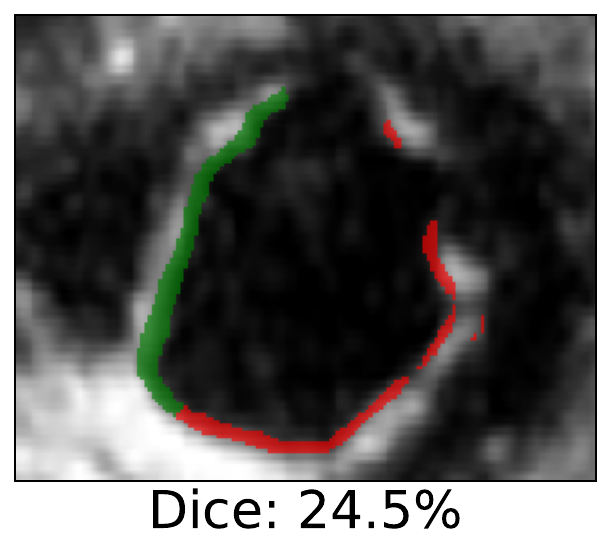}
        \end{minipage}
        \begin{minipage}[t]{0.24\linewidth}
          \includegraphics[width=\linewidth]{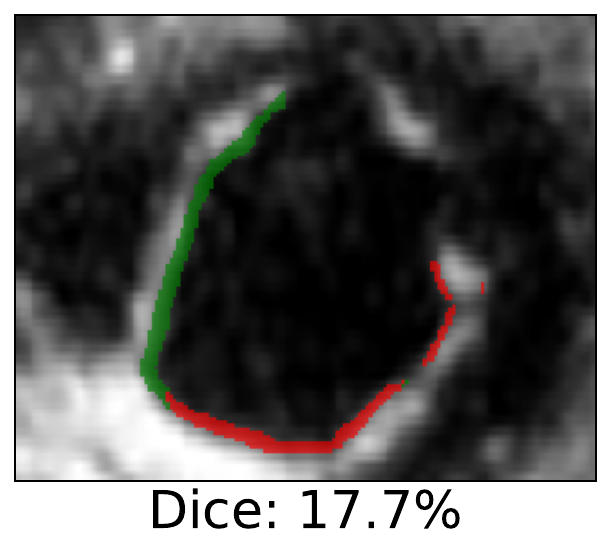}
        \end{minipage}
        \hrule
        \vspace{0.5em}
        \begin{minipage}[t]{\linewidth}
          \centering{UTNet}
        \end{minipage}
        \begin{minipage}[t]{0.24\linewidth}
          \includegraphics[width=\linewidth]{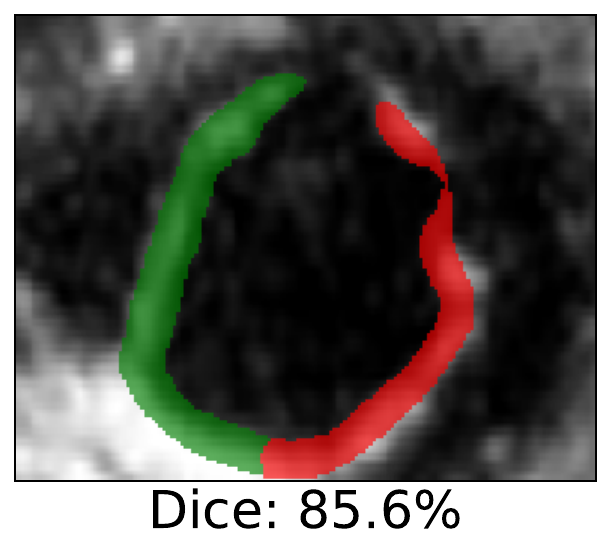}
        \end{minipage}
        \begin{minipage}[t]{0.24\linewidth}
          \includegraphics[width=\linewidth]{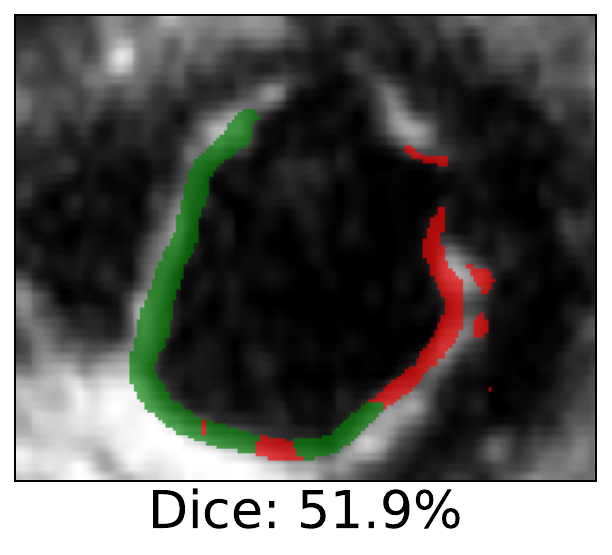}
        \end{minipage}
        \begin{minipage}[t]{0.24\linewidth}
          \includegraphics[width=\linewidth]{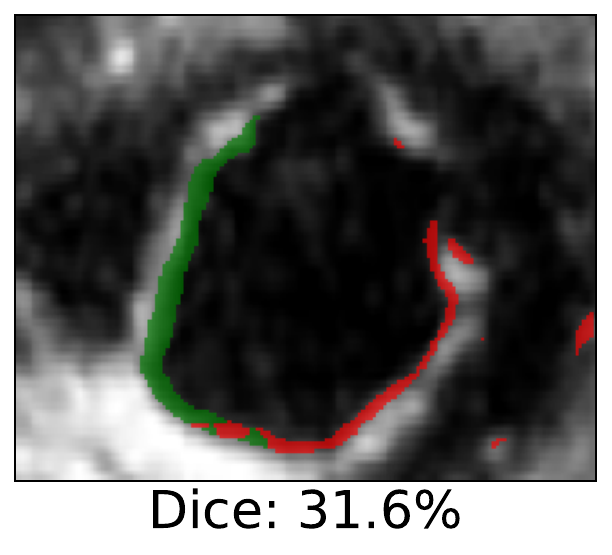}
        \end{minipage}
        \begin{minipage}[t]{0.24\linewidth}
          \includegraphics[width=\linewidth]{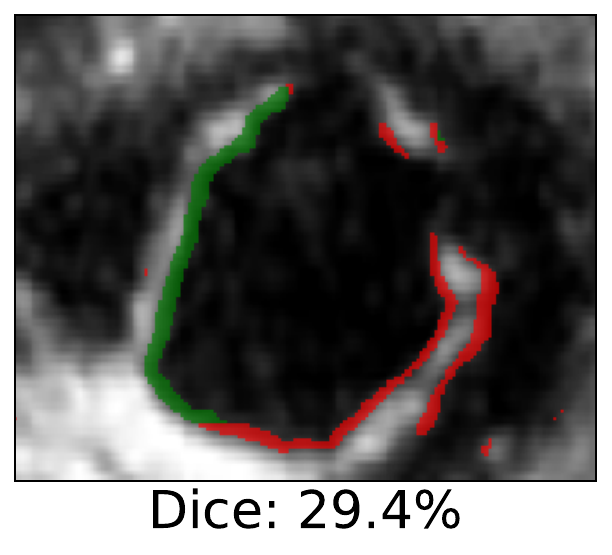}
        \end{minipage}
        \hrule
        \vspace{0.5em}
        \begin{minipage}[t]{\linewidth}
          \centering{TransUNet}
        \end{minipage}
        \begin{minipage}[t]{0.24\linewidth}
          \includegraphics[width=\linewidth]{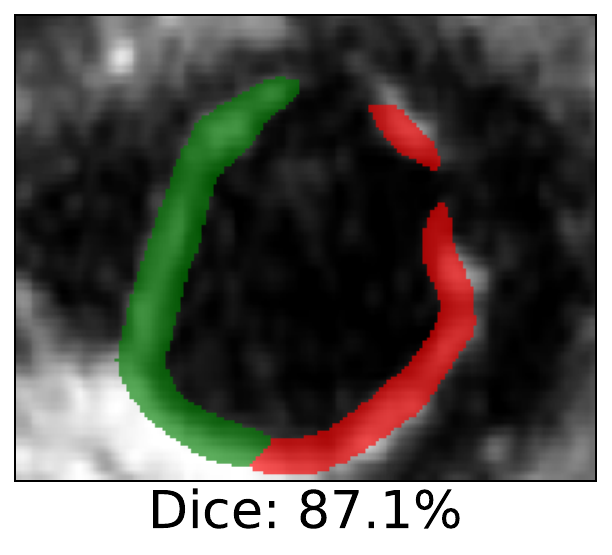}
        \end{minipage}
        \begin{minipage}[t]{0.24\linewidth}
          \includegraphics[width=\linewidth]{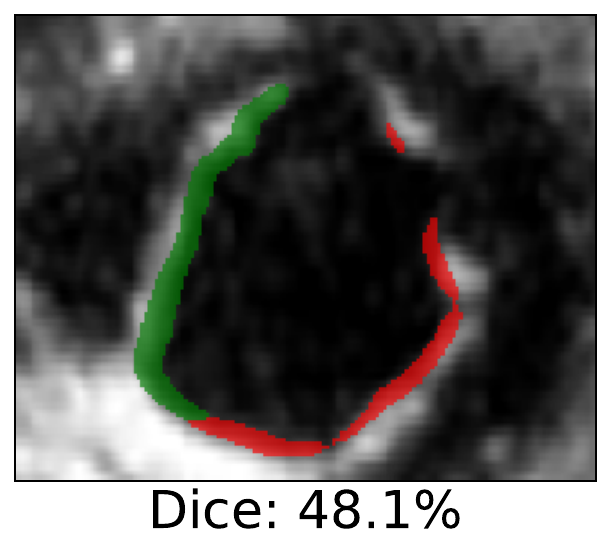}
        \end{minipage}
        \begin{minipage}[t]{0.24\linewidth}
          \includegraphics[width=\linewidth]{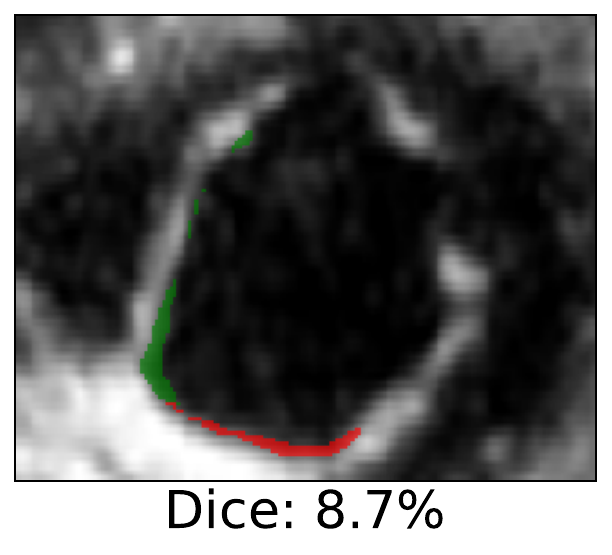}
        \end{minipage}
        \begin{minipage}[t]{0.24\linewidth}
          \includegraphics[width=\linewidth]{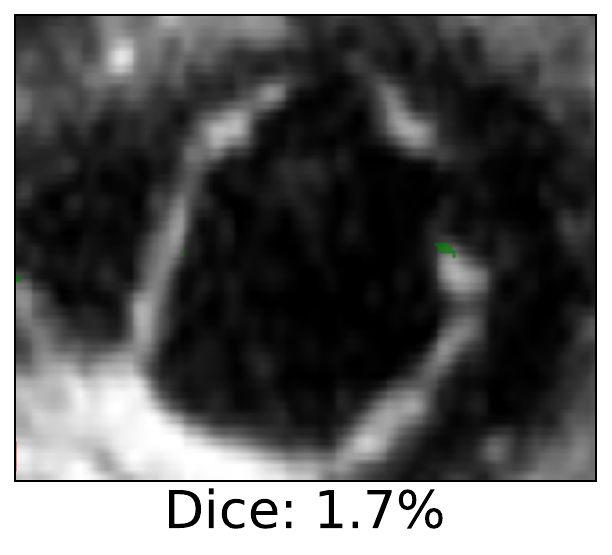}
        \end{minipage}
        \hrule
        \vspace{0.5em}
        \begin{minipage}[t]{\linewidth}
          \centering{nnFormer}
        \end{minipage}
        \begin{minipage}[t]{0.24\linewidth}
          \includegraphics[width=\linewidth]{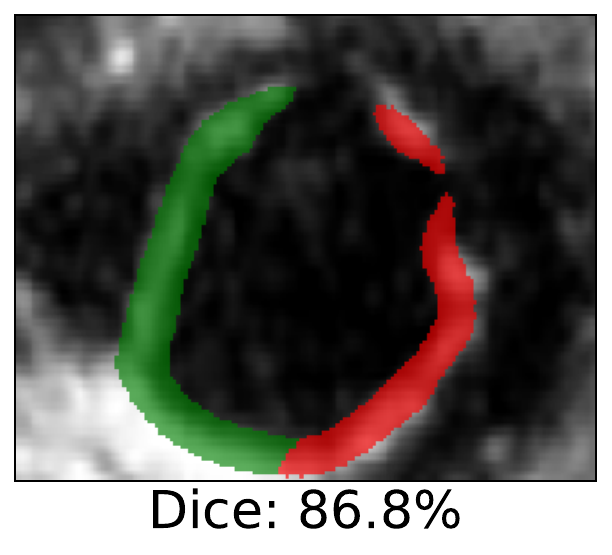}
        \end{minipage}
        \begin{minipage}[t]{0.24\linewidth}
          \includegraphics[width=\linewidth]{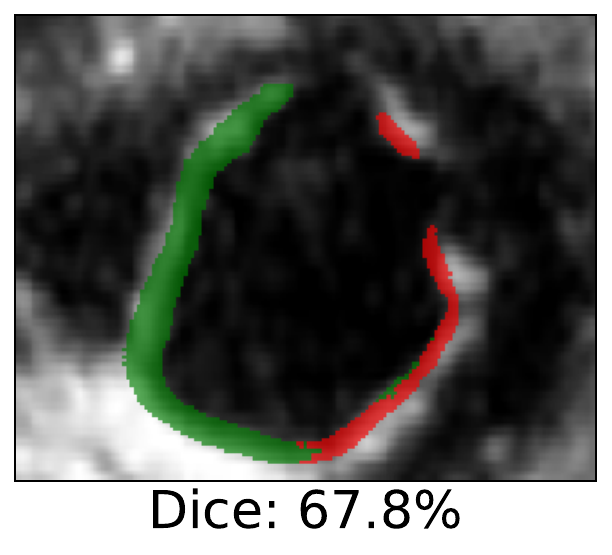}
        \end{minipage}
        \begin{minipage}[t]{0.24\linewidth}
          \includegraphics[width=\linewidth]{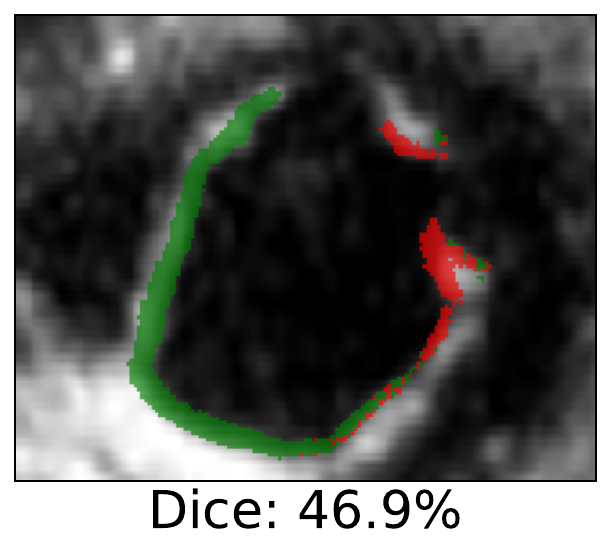}
        \end{minipage}
        \begin{minipage}[t]{0.24\linewidth}
          \includegraphics[width=\linewidth]{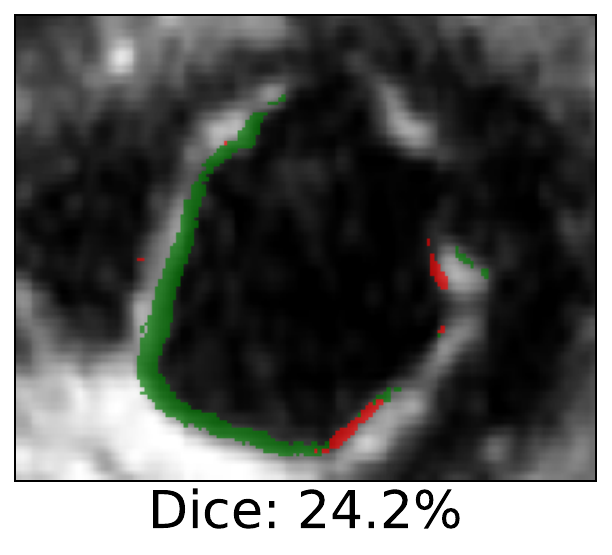}
        \end{minipage}
        \hrule
        \vspace{0.5em}
        \begin{minipage}[t]{\linewidth}
          \centering{FNO}
        \end{minipage}
        \begin{minipage}[t]{0.24\linewidth}
          \includegraphics[width=\linewidth]{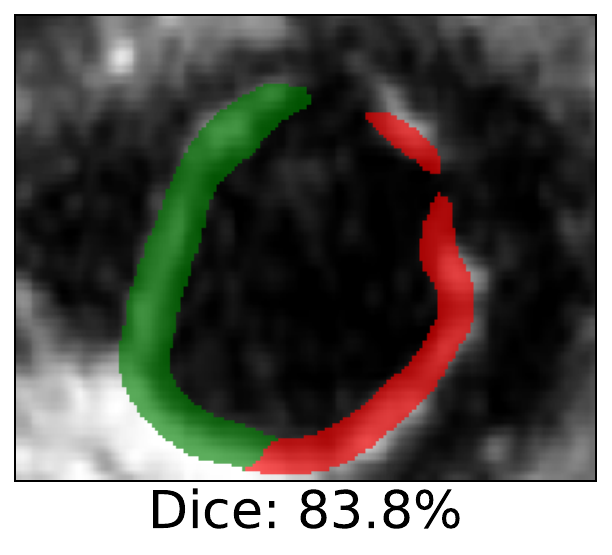}
        \end{minipage}
        \begin{minipage}[t]{0.24\linewidth}
          \includegraphics[width=\linewidth]{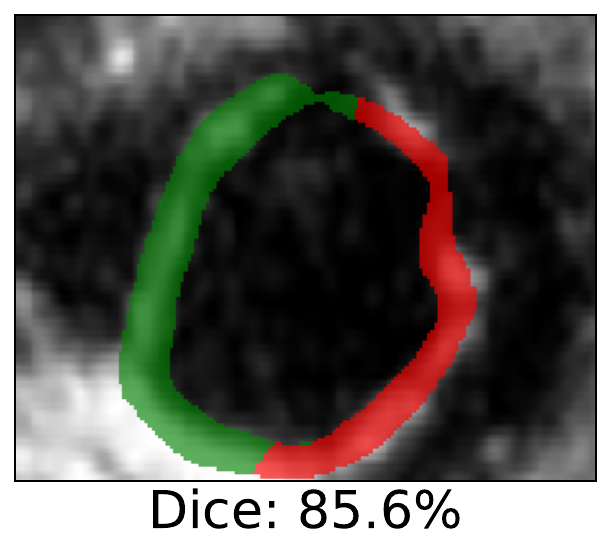}
        \end{minipage}
        \begin{minipage}[t]{0.24\linewidth}
          \includegraphics[width=\linewidth]{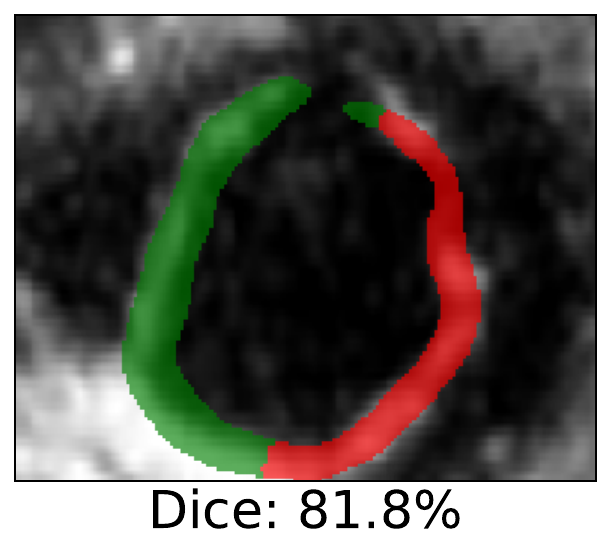}
        \end{minipage}
        \begin{minipage}[t]{0.24\linewidth}
          \includegraphics[width=\linewidth]{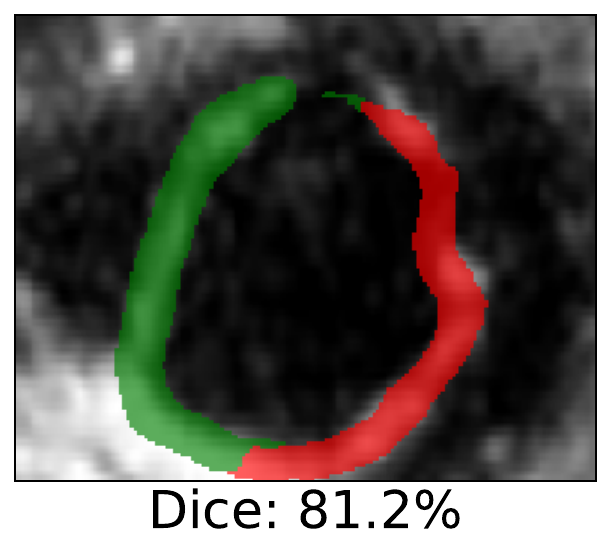}
        \end{minipage}
        \hrule
        \vspace{0.5em}
        \begin{minipage}[t]{\linewidth}
          \centering{FNOSeg}
        \end{minipage}
        \begin{minipage}[t]{0.24\linewidth}
          \includegraphics[width=\linewidth]{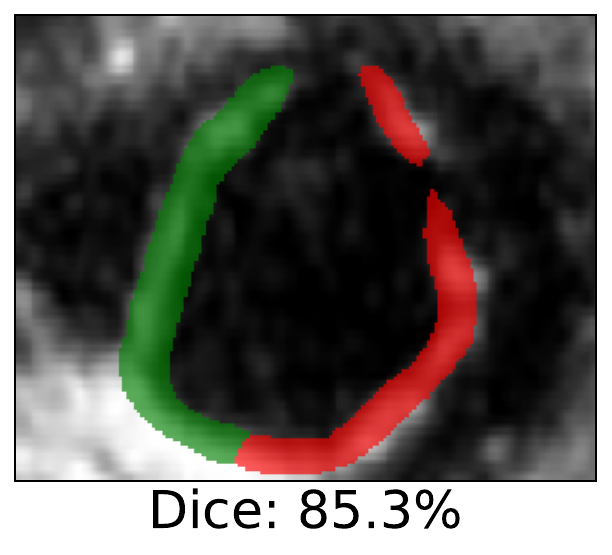}
        \end{minipage}
        \begin{minipage}[t]{0.24\linewidth}
          \includegraphics[width=\linewidth]{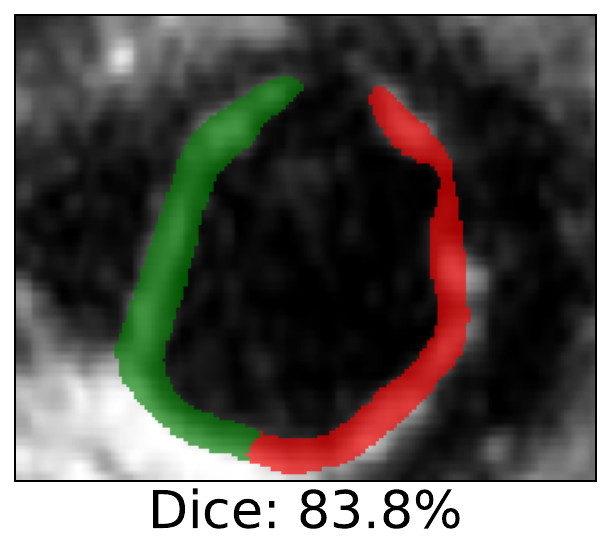}
        \end{minipage}
        \begin{minipage}[t]{0.24\linewidth}
          \includegraphics[width=\linewidth]{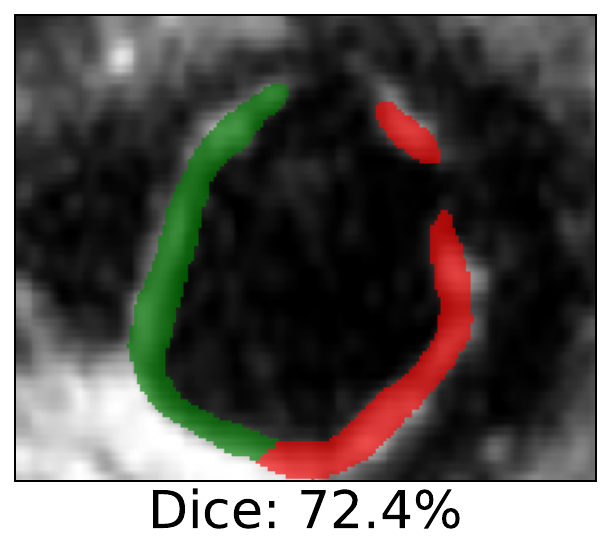}
        \end{minipage}
        \begin{minipage}[t]{0.24\linewidth}
          \includegraphics[width=\linewidth]{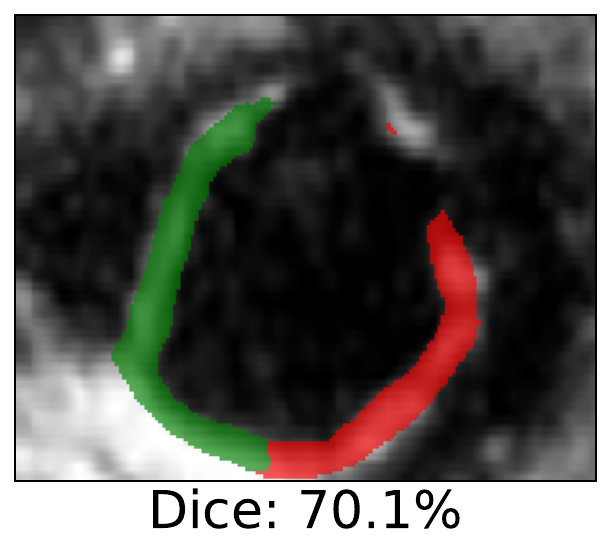}
        \end{minipage}
        \hrule
        \vspace{0.5em}
        \begin{minipage}[t]{\linewidth}
          \centering{HNOSeg}
        \end{minipage}
        \begin{minipage}[t]{0.24\linewidth}
          \includegraphics[width=\linewidth]{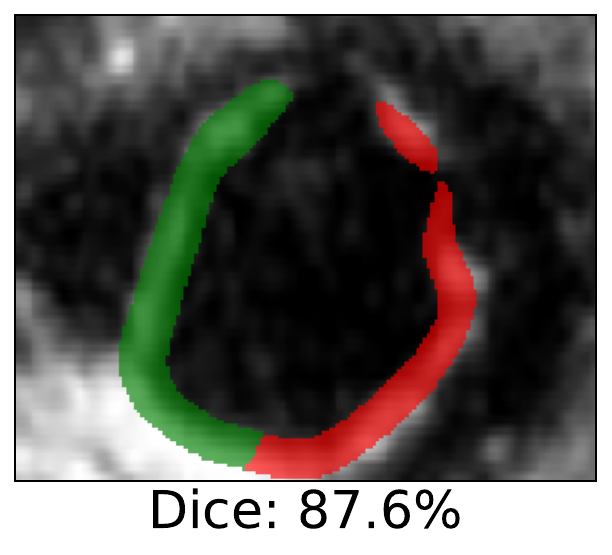}
        \end{minipage}
        \begin{minipage}[t]{0.24\linewidth}
          \includegraphics[width=\linewidth]{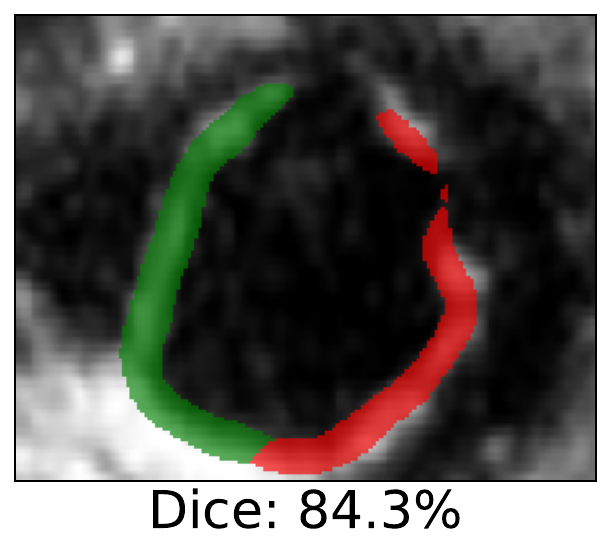}
        \end{minipage}
        \begin{minipage}[t]{0.24\linewidth}
          \includegraphics[width=\linewidth]{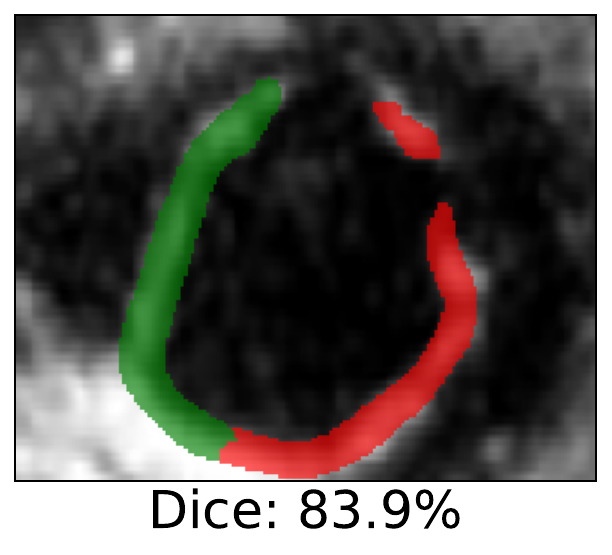}
        \end{minipage}
        \begin{minipage}[t]{0.24\linewidth}
          \includegraphics[width=\linewidth]{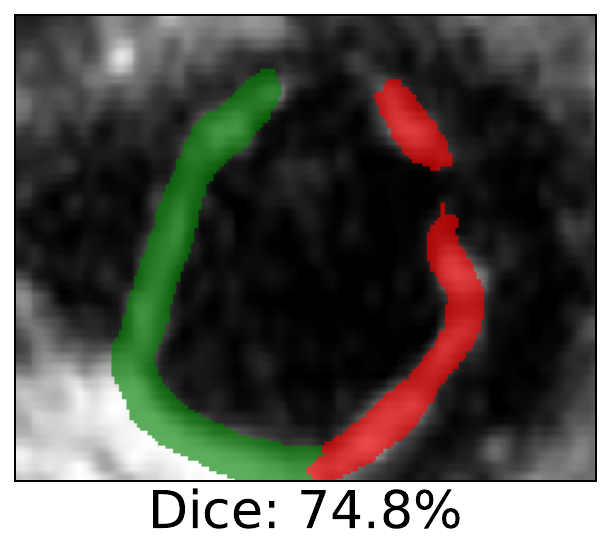}
        \end{minipage}
        \hrule
        \vspace{0.5em}
        \begin{minipage}[t]{\linewidth}
          \centering{HNOSeg-XS}
        \end{minipage}
        \begin{minipage}[t]{0.24\linewidth}
          \includegraphics[width=\linewidth]{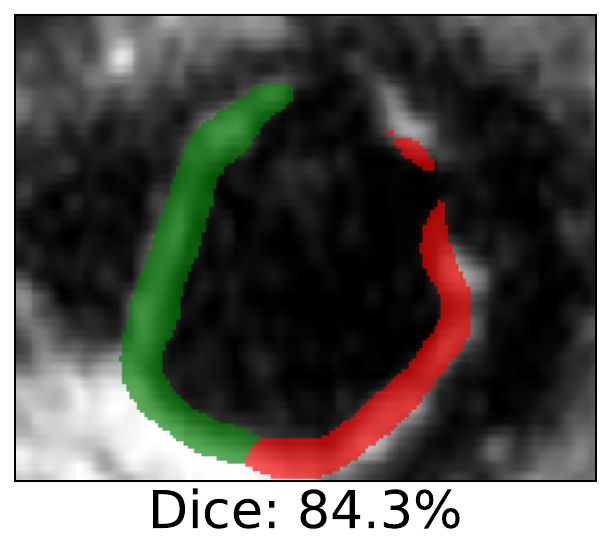}
        \end{minipage}
        \begin{minipage}[t]{0.24\linewidth}
          \includegraphics[width=\linewidth]{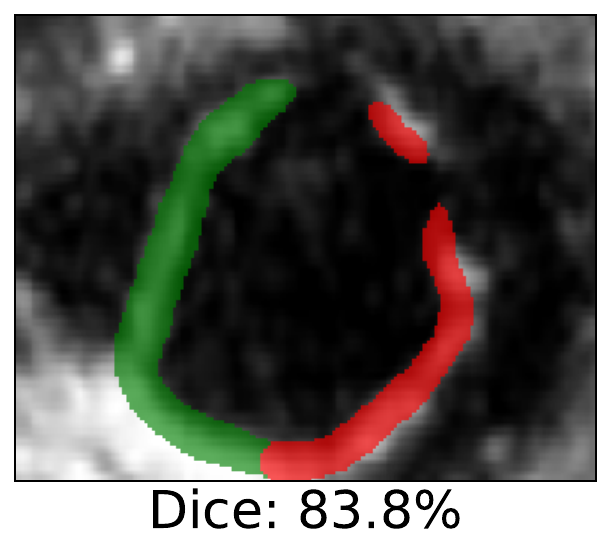}
        \end{minipage}
        \begin{minipage}[t]{0.24\linewidth}
          \includegraphics[width=\linewidth]{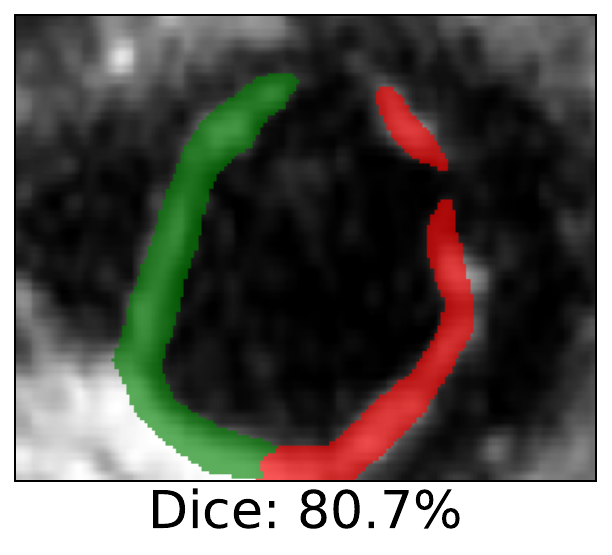}
        \end{minipage}
        \begin{minipage}[t]{0.24\linewidth}
          \includegraphics[width=\linewidth]{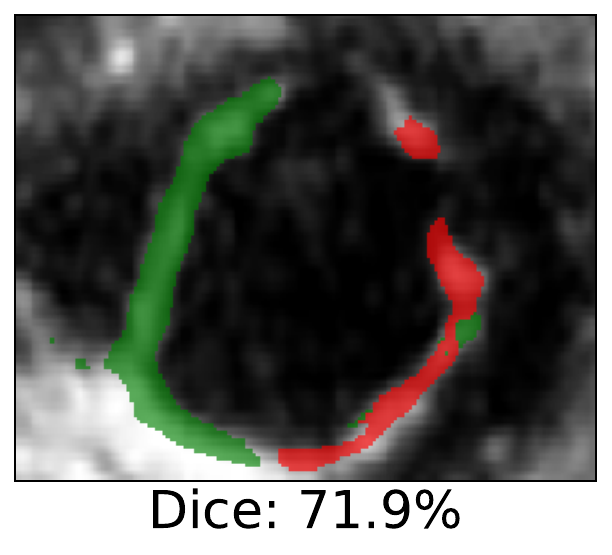}
        \end{minipage}
        \hrule
        \vspace{0.5em}
        \begin{minipage}[t]{0.24\linewidth}
          \centering{224$\times$160$\times$208}
        \end{minipage}
        \begin{minipage}[t]{0.24\linewidth}
          \centering{112$\times$80$\times$104}
        \end{minipage}
        \begin{minipage}[t]{0.24\linewidth}
          \centering{75$\times$54$\times$70}
        \end{minipage}
        \begin{minipage}[t]{0.24\linewidth}
          \centering{56$\times$40$\times$52}
        \end{minipage}
    \end{minipage}
    \caption{Visual comparisons of models trained with different image resolutions (MVSeg'23), tested on an unseen sample of size 224$\times$160$\times$208. The Dice coefficients were averaged from the PL and AL regions. Note that the 2D visualizations may not reflect the Dice coefficients between 3D volumes.}
    \label{fig:mvseg_visual}
\end{figure}

\subsection{Comparisons of Different Models}

To study the robustness to image resolution, models were trained with images downsampled by different factors (1, 2, 3, and 4), and tested on images without downsampling. Fig. \ref{fig:vs_imgsize} and Table \ref{table:results} show that for all datasets, while the non-neural operator (NNO) models obtained decent results at the highest training resolutions, their accuracies decreased quickly with the reduction in training image sizes. In contrast, the neural operator models were more robust to training resolutions. In general, nnFormer had the best performance among the NNO models regardless of the training resolutions. Among the neural operator models, HNOSeg and HNOSeg-XS had similar performance and they outperformed FNOSeg. FNO was the most robust model, but its performance was inconsistent. 

For BraTS'23, all neural operator models were robust to the training image sizes. For example, when the training image size changed from 240$\times$240$\times$155 to 80$\times$80$\times$52, the reductions in Dice coefficients of HNOSeg and HNOSeg-XS were 4.4\% and 3.5\%, respectively. On the contrary, the Dice coefficient of nnFormer dropped by 14.4\%, and this was the smallest drop among the NNO models. For KiTS'23, all models were more sensitive to the training image sizes. When the training resolution changed from 256$\times$256$\times$128 to 128$\times$128$\times$64, the Dice coefficients of HNOSeg and HNOSeg-XS dropped by 4.6\% and 3.9\%, respectively, while that of nnFormer dropped by 23.2\%. At the training resolution of 86$\times$86$\times$43, the difference between the Dice coefficients of HNOSeg-XS and nnFormer was 39.1\%. FNO had the worst accuracy with KiTS'23 as it could not probably segment the masses and tumors. For MVSeg'23, the differences among the models were similar to those of BraTS'23 and KiTS'23. Therefore, HNOSeg and HNOSeg-XS had superior robustness regardless of imaging modalities and organ types.

At the highest training resolutions, i.e., when the training and testing resolutions were the same, nnFormer had the best accuracies in all datasets. For BraTS'23, the average Dice coefficient of nnFormer was 89.7\%. HNOSeg had the second highest score of 89.1\%, and the score of HNOSeg-XS was 88.7\%. For KiTS'23, the Dice coefficient of nnFormer was 81.7\%. HNOSeg and HNOSeg-XS had the second and third highest scores of 79.5\% and 77.6\%, respectively. For MVSeg'23, the Dice coefficient of nnFormer was 84.3\%. V-Net-DS had the second highest score of 82.9\% and the score of HNOSeg-XS was 82.1\%. Hence, except nnFormer, HNOSeg and HNOSeg-XS had similar, and sometimes better, segmentation capability as other NNO models.

Fig. \ref{fig:brats_visual}, \ref{fig:kits_visual}, and \ref{fig:mvseg_visual} show visual examples of segmentation results. Consistent with the numerical results, the neural operator models were superior in resolution robustness. Fig. \ref{fig:brats_visual} shows that there were some high-frequency features in the necrotic (red) region. At the highest training resolution, the NNO models could  capture such features while the neural operator models missed them. Interestingly, such features could be partially captured by the neural operator models when they were trained at lower resolutions. Fig. \ref{fig:mvseg_visual} shows that when the training resolution decreased, the thicknesses of the leaflets segmented by the NNO models decreased simultaneously, which were consistent with the thicknesses of the low-resolution ground truths in training. Similarly, Fig. \ref{fig:kits_visual} shows that even at the training size of 128$\times$128$\times$64, the small mass and tumor could not be properly segmented by the NNO models. These show that the smaller the size, the more difficult an object to be segmented by the NNO models when the resolutions changed. On the other hand, as the neural operator models are resolution agnostic, they could segment properly unless the object features were corrupted by downsampling (e.g., at the training resolution of 64$\times$64$\times$32 for KiTS'23).

\begin{table}[t]
\caption{Number of parameters, inference time per image in seconds, and memory use in GiB with a batch size of 1. The inference time was averaged from images of size 240$\times$240$\times$155 for BraTS'23, 256$\times$256$\times$128 for KiTS'23, and 224$\times$160$\times$208 for MVSeg'23. The best two values in each column are highlighted.}
\label{table:resources}

\scriptsize
\centering

\newcolumntype{R}{r@{\extracolsep{\fill}}}
\newcolumntype{L}{l@{\extracolsep{2pt}}}
\newcommand{\boldblue}[1]{\textcolor{blue}{\textbf{#1}}}

\begin{tabularx}{\linewidth}{LRRRRRRRRR}
\toprule
 & \multicolumn{3}{c}{\textbf{BraTS'23}} & \multicolumn{3}{c}{\textbf{KiTS'23}} & \multicolumn{3}{c}{\textbf{MVSeg'23}}\\
 \cline{2-4} \cline{5-7} \cline{8-10} \noalign{\smallskip}
 & Param & Time & Mem & Param & Time & Mem & Param & Time & Mem \\
\midrule
V-Net-DS
& 22.5M & 0.39 & 7.2
& 22.5M & 0.33 & 6.7
& 22.5M & 0.28 & 5.9
\\
UTNet 
& 29.4M & 0.39 & 2.8
& 29.4M & 0.32 & 1.4
& 29.4M & 0.28 & 2.1
\\
TransUNet 
& 16.7M & 0.28 & 3.3
& 16.7M & \boldblue{0.21} & 3.0
& 16.7M & \boldblue{0.16} & 2.7
\\
nnFormer 
& 138.8M & 0.49 & 2.8
& 138.8M & 0.42 & 2.5
& 138.8M & 0.37 & 2.3
\\
FNO 
& 15.9M & \boldblue{0.25} & \boldblue{1.3}
& 28.3M & 0.25 & \boldblue{1.1}
& 185.8M & 0.20 & 1.6
\\
FNOSeg 
& 71.2k & 0.31 & \boldblue{1.2}
& 70.6k & 0.29 & \boldblue{1.0}
& 70.6k & 0.24 & \boldblue{0.9}
\\
HNOSeg 
& \boldblue{57.4k} & 0.39 & \boldblue{1.2}
& \boldblue{56.8k} & 0.39 & \boldblue{1.1}
& \boldblue{56.8k} & 0.32 & \boldblue{0.9}
\\
HNOSeg-XS 
& \boldblue{28.2k} & \boldblue{0.23} & 1.6
& \boldblue{34.7k} & \boldblue{0.24} & 1.8
& \boldblue{27.6k} & \boldblue{0.14} & \boldblue{1.2}
\\
\bottomrule
\end{tabularx}
\end{table}

\subsection{Computational Requirements} 

Table \ref{table:resources} and Fig. \ref{fig:param_time_mem} show the computational requirements on a single image. Except FNO, all neural operator models had fewer than 72k model parameters. As FNO used different parameters at different frequencies, its numbers of model parameters were $k_{max}$ dependent and ranged from 15.9M to 185.8M. HNOSeg-XS had the fewest model parameters ($<$ 35k), which were fewer than 0.2\% of TransUNet (16.7M) and 0.03\% of nnFormer (138.8M). FNOSeg required parameters for the real and imaginary parts, thus it had more parameters than HNOSeg. For memory use, the neural operator models used less memory ($<$ 1.8 GiB) than the NNO models in general. HNOSeg-XS used more memory than HNOSeg because of the U-Net skip connections. The memory uses of FNOSeg and HNOSeg were almost identical. This is because in implementation, as the FFT of a real signal is Hermitian-symmetric, its negative frequencies in the last dimension can be omitted, and this offsets the real-valued advantage of HNOSeg. For computation time, TransUNet and HNOSeg-XS were the fastest on average ($<$ 0.22 s), with HNOSeg-XS slightly faster. FNOSeg was faster than HNOSeg because PyTorch does not provide implementation of the Hartley transform. We computed the Hartley transform using the standard FFT and $(\mathcal{H}f) = \mathrm{Real}(\mathcal{F}f) - \mathrm{Imag}(\mathcal{F}f)$, which was much slower than the FFT specialized for real-valued inputs.

\section{Discussion}

The experimental results show that HNOSeg and NHOSeg-XS outperform FNO and FNOSeg in accuracy, and they are superior in resolution robustness compared to NNO models. Furthermore, HNOSeg and HNOSeg-XS are more computationally efficient than NNO models. Comparing between HNOSeg and HNOSeg-XS, HNOSeg is more accurate and uses less memory, while HNOSeg-XS is faster with fewer model parameters. As HNOSeg and HNOSeg-XS have their own pros and cons, it is difficult to decide which one is better, and choosing which to use depends on the computational and accuracy requirements of the application.

Given the relatively simple structures of the HNO and HNO-XS blocks, they can be flexibly incorporated into different architectures. The use of U-Net skip connections in HNOSeg-XS is a good example, and more complicated operations such as providing features to more sophisticated architectures (e.g., \cite{Conference:Xie:MICCAI2021:cotr,Journal:Chen:MIA2024:transunet}) are also possible. Furthermore, although we focus on training with image downsampling in this paper, our models can also be used with patch-wise training to provide better robustness to patch sizes and computational efficiency. 

Apart from model training, the resolution robustness of our models can also benefit real-world applications. Depending on factors such as patient size, scanner model, and acquisition technique (e.g., pre-operative vs. intraoperative), image resolutions can be very different even within the same facility \cite{Journal:Smith:JMI2021:variability}. In this aspect, the use of resolution-robust models can reduce the number of models required, the complexity of the supporting system, and also the chance of segmentation failures caused by mismatched resolutions. As a result, the costs of deployment and maintenance can be reduced.

\section{Conclusion}

In this paper, we propose HNOSeg-XS to be an alternative image segmentation model. By replacing the Fourier transform by the Hartley transform which produces real numbers in the frequency domain, more sophisticated deep learning operations in the frequency domain can be performed for better accuracy and efficiency. The use of shared parameters in the frequency domain largely reduces the number of model parameters, and the adoption of SNN and SELU further reduces computation time and memory use. Experimental results show that HNOSeg-XS is robust to training image resolution, computationally efficient, and extremely parameter efficient. Therefore, HNOSeg-XS can be a promising alternative to existing CNN and transformer models especially when computational resources are limited.

\bibliographystyle{IEEEtran}
\bibliography{Ref}

\end{document}